\documentclass[10pt,twocolumn,letterpaper]{article}

\usepackage[pagenumbers]{wacv} 

\usepackage{graphicx}
\usepackage{amsmath}
\usepackage{amssymb}
\usepackage{booktabs}

\usepackage{subcaption}
\usepackage{colortbl}
\usepackage{xcolor}
\usepackage{placeins}
\usepackage[accsupp]{axessibility}

\newcommand{\subheading}[1]{\textbf{#1}.}
\newcolumntype{a}{>{\columncolor{verylightgray}}c}

\definecolor{verylightgray}{HTML}{E0E0E0}

\renewcommand{\eqref}[1]{Eq.~(\ref{#1})}

\newcommand{\papertitle}{Neural Texture Puppeteer\xspace}
\newcommand{\paperabr}{NeTePu\xspace}
\newcommand{\dataabr}{NePuMoo\xspace}

\usepackage[pagebackref,breaklinks,colorlinks]{hyperref}
\usepackage{balance}

\usepackage[capitalize]{cleveref}
\crefname{section}{Sec.}{Secs.}
\Crefname{section}{Section}{Sections}
\Crefname{table}{Table}{Tables}
\crefname{table}{Tab.}{Tabs.}

\def\wacvPaperID{9} 
\def\confName{WACV}
\def\confYear{2024}

\begin{document}

\title{\papertitle: A Framework for Neural Geometry and Texture Rendering of Articulated Shapes, Enabling Re-Identification at Interactive Speed}

\author{Urs Waldmann \and Ole Johannsen \\
\\
University of Konstanz, Germany \\
{\tt\small \{firstname.lastname\}@uni-konstanz.de}
\and Bastian Goldluecke 
}
\maketitle

\begin{abstract}
   In this paper, we present a neural rendering pipeline for textured articulated shapes that we call Neural Texture Puppeteer.
   Our method separates geometry and texture encoding.
   The geometry pipeline learns to capture spatial relationships on the surface of the articulated shape from ground truth data that provides this geometric information.
   A texture auto-encoder makes use of this information to encode textured images into a
   global latent code.
   This global texture embedding can be efficiently trained separately from the 
   geometry, and used in a downstream task to identify individuals.
   The neural texture rendering and the identification of individuals run at interactive speeds.
   %
   To the best of our knowledge, we are the first to offer a promising alternative to CNN- or transformer-based approaches for re-identification of articulated individuals based on neural rendering.
   Realistic looking novel view and pose synthesis for different synthetic cow textures further demonstrate the quality of our method.
   Restricted by the availability of ground truth data for the articulated shape's geometry, the quality for real-world data synthesis is reduced.
   We further demonstrate the flexibility of our model for real-world data by applying a synthetic to real-world texture domain shift where we reconstruct the texture from a real-world 2D RGB image.
   Thus, our method can be applied to endangered species where data is limited.
   Our novel synthetic texture dataset~\dataabr is publicly available to inspire further development in the field of neural rendering-based re-identification.
\end{abstract}

\section{Introduction}
%
%
Recent developments in neural rendering brought a big boost to many vision applications~\cite{neural_fields, tewari2021advances, tewari2020state}.
%
%
One of them is novel view and novel pose synthesis.
With the success of NeRF~\cite{NeRF} that can render rigid objects in 3D from multiple input images, many other methods for novel view synthesis of static content were developed~\cite{lin2020sdf,niemeyer2020differentiable}.
%
%
Next, approaches that additionally handle articulated shapes leveraging the SMPL mesh model~\cite{SMPL} were developed~\cite{NeuralBody,AniNeRF}.
For example,~\cite{wang2019re} generates an UV texture map that is combined with a rendering tensor generated with OpenDR~\cite{loper2014opendr} from the SMPL~\cite{SMPL} model.
A drawback of all methods that leverage the SMPL model~\cite{SMPL} is that it limits the methods to a pre-defined class of shapes.
Articulated shapes can also be handled with methods that leverage implicit neural representations~\cite{A-NeRF} and do not rely on the SMPL model.

However, NeRF-based approaches use volumetric rendering which is time and memory intensive.
In contrast, there are methods for static content~\cite{LFN} and articulated shapes~\cite{NePu} that render in 2D.
Their advantage compared to volumetric rendering of NeRF-based approaches is that they require only a single neural network evaluation per ray.
This results in significantly faster inference speed~\cite{LFN,NePu}.
%

%
Like neural rendering, tracking is a vast field of research and its applications range from self-driving cars~\cite{liu2021robust} to the study of collective behaviour~\cite{Kays2015te}.
Therefore reliable and accurate tracking of objects like humans~\cite{chen2020cross,rafi2020self,cui2022joint,rajasegaran2022tracking} and animals~\cite{maDLC,SLEAP,TRex,I-MuPPET,3d-muppet,unlabprop} is required.
An essential extension for many tracking applications is the re-identification of individuals when leaving and re-entering the scene.
Most frameworks use CNNs or vision transformers to extract image features for object re-identifiaction~\cite{maDLC, he2021transreid, zheng2017discriminatively, liu2017end}.
%

Surprisingly, only one study that we are aware of employs neural rendering to identify objects in a tracking scenario,
namely~\cite{yuan2021star}, which tracks single rigid objects in motion in a self-supervised manner.
%
%
It however has limitations in practical applications because it handles only rigid objects and has a slow inference speed as it is NeRF-based~\cite{NeRF} and uses time- and memory-intensive volumetric rendering.
%

In particular for the study of collective behaviour, but also for other research in biology directed towards animals, it is crucial to have high-performance tracking~\cite{SORT} and re-identification of individuals~\cite{ferreira2020deep}.
The aim of our work is therefore to leverage the recent advances in neural rendering and develop a pipeline that can handle more than one textured articulated shape in a single model, thus enabling re-identification in tracking scenarios.
%

\subheading{Contributions}
We present a flexible neural rendering pipeline for textured articulated shapes that we call~\papertitle (\paperabr).
Our method separates geometry and texture encoding.
The geometry pipeline learns to capture spatial relationships on the surface of the articulated shape from ground truth data that provides this geometric information.
For this purpose we extend the idea in~\cite{NOCS} to articulated shapes, cf.~\cref{sec:positional_encoding}.
We show that our method encodes a distinct global texture embedding (cf.~\cref{fig:reid:netepu}), which is computed from this geometric information and 2D color information.
This global texture embedding can be used in a downstream task to identify individuals.
Our neural rendering-based re-identification of individuals runs at interactive speeds (cf.~\cref{sec:experiments:reID}) and thus offers a promising alternative to CNN- or transformer-based approaches in tasks such as the re-identification of individuals in tracking scenarios.
To the best of our knowledge, we are the first to provide a framework for re-identification of articulated individuals based on neural rendering.
We also demonstrate the quality of our method with realistic looking novel view and pose synthesis for different synthetic cow textures, cf.~\cref{fig:novel-pose-novel-views}.
We further demonstrate the flexibility of our model by applying a synthetic to real-world texture domain shift where we reconstruct the texture from a real-world 2D RGB image with a model trained on synthetic data only.
This makes our model useful for real-world applications with endangered animal species.

Our novel synthetic texture dataset~\dataabr (cf.~\cref{sec:dataset}) 
together with the code to reproduce the results of this paper
are publicly available at~\url{https://github.com/urs-waldmann/NeTePu} to inspire further development in the field of neural rendering-based re-identification.
%

\section{Related Work}
We give an overview on re-identification, the texture prediction for articulated objects and datasets for these tasks.
For an overview on neural rendering we refer to~\cite{neural_fields, tewari2021advances, tewari2020state}.

\subheading{Texture Prediction of Articulated Objects}
There exist a wide range of methods to predict the texture of articulated human shapes.
Good results can be achieved with NeRF-based~\cite{NeRF} approaches~\cite{A-NeRF, AniNeRF},
approaches based on implicit neural representations~\cite{NeuralBody, PIFu}
or approximate differentiable rendering~\cite{wang2019re}.
%
%
In contrast to our method, all these approaches make some strong simplifications.
While~\cite{AniNeRF, NeuralBody, wang2019re} leverage the SMPL~\cite{SMPL} model, \cite{A-NeRF, AniNeRF} learn only one model per texture.
Furthermore,~\cite{PIFu, A-NeRF, AniNeRF} render in 3D, while our method shifts rendering to 2D
by employing techniques from~\cite{NePu},
which makes our method much faster.
%

For animals, we are aware of four works~\cite{SMALR, SMALST, CMR, li2020self} which learn UV maps which they project on a mesh model.
%
%
In contrast to our neural rendering approach, they all use morphable models on which they map the predicted texture.
%

Only recently
Wu~\etal published~\cite{wu2023magicpony}.
This framework reconstructs textured articulated objects from single images.
In contrast to this work, we extract NNOPCS maps (similar to NOCS maps~\cite{NOCS}, cf.~\cref{sec:positional_encoding}) instead of a prior mesh for the object category.
Further, instead of a collection of single-view images for an object category, we leverage a multi-view setup to assure that the individual's texture has been observed completely during training. This guarantees a meaningful global texture latent code for each individual from any viewpoint for re-identification in tracking tasks.

\subheading{Re-identification of Individuals}
Tracking is a vast field under constant development.
%
%
For a summary, we refer
to~\cite{yilmaz2006object, ciaparrone2020deep, dendorfer2021motchallenge}.
%
%
Re-identification of individuals is an important task in scenarios where individuals exit and re-enter the scene.
Popular pipelines for object re-identification train CNN backbones or vision transformers to extract image features~\cite{maDLC, he2021transreid, zheng2017discriminatively, liu2017end},
and often~\cite{zheng2017discriminatively, liu2017end, he2021transreid} need positive and negative pairs of the object in their training phase.
Notably, \cite{maDLC, he2021transreid, zheng2017discriminatively, liu2017end} base re-indentification on selected image features and do not make sure to encode the whole texture of the individual.
In contrast, our encoded global latent code is used to reconstruct the complete texture from arbitrary viewpoints
within our neural texture rendering pipeline.
From a single input image,~\cite{wang2019re} generates an UV texture map that is combined with a rendering tensor generated with OpenDR~\cite{loper2014opendr} using the SMPL~\cite{SMPL} mesh model.
In the end, they use a pre-trained person re-identification model to extract features of the rendered and the input image on which they calculate a re-identification loss.
In contrast to our work, the authors use a re-identification network to generate a texture, while
we use the texture for re-identification, which is a fundamental technical difference.
Due to this, we can not perform an experimental comparison with~\cite{wang2019re}.
%

The only other work that we aware of that uses neural rendering for tracking is~\cite{yuan2021star}.
While~\cite{yuan2021star} reconstructs single rigid objects in motion from multi-view RGB videos in a self-supervised manner with a NeRF-based~\cite{NeRF} neural rendering method,
we can reconstruct articulated objects, which makes our method applicable to humans and animals.
Furthermore our method shifts rendering from 3D to 2D, as it is based on the neural rendering framework~\cite{NePu}.
This fundamental difference makes our method much faster compared to NeRF-based~\cite{NeRF} approaches like~\cite{yuan2021star}.
%
%

\subheading{Datasets}
We aim to reconstruct textures of articulated shapes with a keypoint-based neural rendering pipeline that leverages our NNOPCS maps.
%
For this task we need datasets that contain articulated shapes like humans or animals together with keypoints, segmentation masks, RGB images and NNOPCS maps.
For humans the most popular real-world dataset is H36M~\cite{H36M}.
For animals we are aware of four datasets containing more than one individual~\cite{marshall2021pair,3D-POP,OpenMonkeyStudio,han2023social}.
All these datasets lack ground truth NNOPCS maps for the articulated shapes they contain.
That is why we create a novel synthetic dataset of textured articulated shapes that provides NNOPCS map ground truth.
For a proof of concept, we choose cows as an example species because their textures can be clearly distinguished.
%
%

\section{Architecture}
\label{sec:architecture}
\begin{figure}[t]
    \centerline{
        \includegraphics[width=\linewidth,trim={8cm 5cm 9cm 4cm},clip]{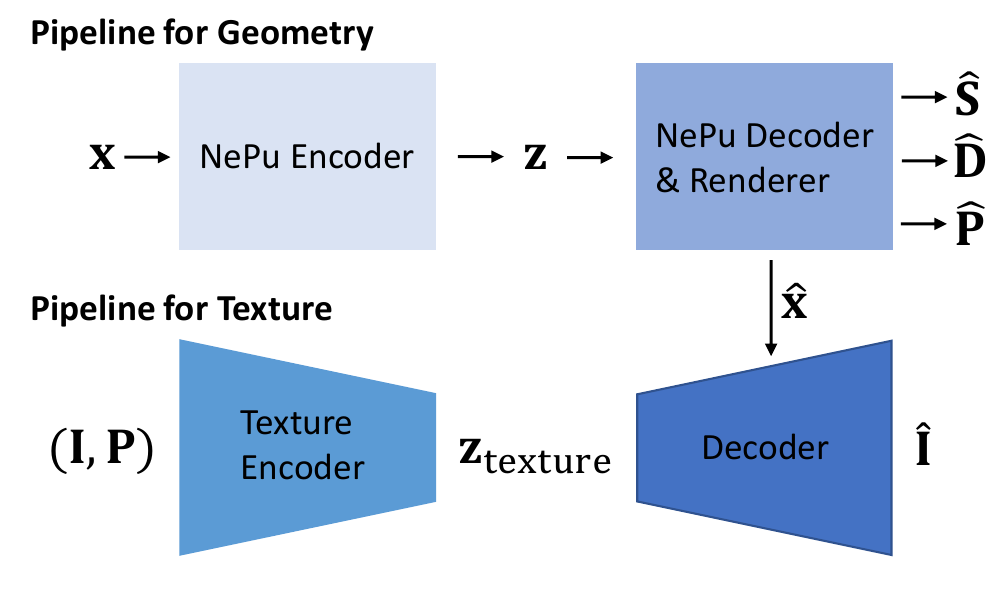}
    }
    \caption{
        \textbf{Complete Pipeline.}
        Our key idea is to disentangle geometry and texture information of articulated shapes.
        We modify the original NePu~\cite{NePu} pipeline to predict a NNOPCS map $\mathbf{\hat{P}}$ (top).
        We also predict the silhouette $\mathbf{\hat{S}}$ and depth map $\mathbf{\hat{D}}$ like the original NePu pipeline.
        The heart of our framework is an encoder-decoder network for the texture of an articulated shape (bottom) that reconstructs an image $\mathbf{I}$.
        %
        %
        Details and the model's loss are in~\cref{sec:architecture}.
    }
    \label{fig:architecture_overview}
\end{figure}
The heart of our framework is an encoder-decoder network that reconstructs the texture of an articulated shape, cf.~\cref{fig:architecture_overview}.
As a basis, we make use of Neural Puppeteer (NePu)~\cite{NePu}
for neural rendering.

\textbf{NePu} is a neural rendering pipeline designed specifically for articulated shapes, which takes as input
a set of~$K$ 3D keypoints~${\mathbf x}\in\mathbb{R}^{K \times 3}$ and a camera position, and maps them to a RGB image of the shape (similar to~\cref{fig:architecture_overview}, top: Pipeline for Geometry).
Notably, NePu renders in the 2D domain, which makes it much faster than volumetric rendering: via a transformer network, local features are computed for each keypoint (NePu Decoder in~\cref{fig:architecture_overview}, top), which are then projected onto the image plane by the camera. Another attention-based network, the actual renderer, then generates a complete 2D image from the projected features at the keypoint locations (NePu Renderer in~\cref{fig:architecture_overview}, top).

Our key idea in this work is to disentangle the pipelines for geometry and texture information of the shape that we call \textbf{\papertitle (\paperabr)}.
Thus~\paperabr, unlike NePu and~\cite{A-NeRF, AniNeRF}, can handle more than one texture in a single model.
The original NePu pipeline is therefore modified to produce what we call a NNOPCS map (cf.~\cref{sec:positional_encoding}) instead of a final image, cf.~\cref{fig:architecture_overview} (top).
This NNOPCS map essentially encodes within the image plane the complete geometric information about the shape relevant for rendering this particular view.
Local features~$\mathbf f$ (we refer to~\cite{NePu} and~\cref{sec:positional_encoding} for details) and global latent code~$\mathbf z$ (cf.~\cref{fig:architecture_overview}, top) for the geometry are augmented with 
separate local features~$\mathbf f_\text{texture}$ and a latent code~$\mathbf{z}_{texture}$ for the texture map.
An encoder network generates~$\mathbf{z}_{texture}$ for a texture from a view of the shape,
while the decoder is a second NePu-based renderer to generate the final image (cf.~\cref{fig:architecture_overview}, bottom).
Putting together the encoder and decoder for~$\mathbf{z}_{texture}$, we essentially obtain a texture auto-encoder which, given a fixed geometry, can be efficiently trained separately from the geometry network.

In the following subsections, we will give the details for these different modules.

\subsection{Pipeline for Geometry}
\label{sec:positional_encoding}
In our framework, the geometry and pose of an articulated shape is defined by the~$K$ different keypoint locations
and a single global latent code~${\mathbf z}\in\mathbb{R}^{d_z}$ for the shape, which is decoded into a $d_f$-dimensional
local feature vector~$\mathbf f$ for each keypoint as input to the renderer. By varying the keypoint locations
and keeping the features fixed, one can generate renderings of arbitrary poses of the same geometry~\cite{NePu}.
We now use the rendering pipeline in~\cite{NePu}, but do not interpret the output as a rendered color image.
Instead, at each point in the image plane, the 3D output defines a 3D coordinate
of a point on the shape in a normalized coordinate space (NOCS~\cite{NOCS}), and additionally in a normalized neutral pose, cf.~\cref{fig:nnopcs_map_neutral_pose}, which is why we denote it \textbf{N}ormalized \textbf{N}eutral \textbf{O}bject \textbf{P}ose and \textbf{C}oordinate \textbf{S}pace (NNOPCS). Thus, the same point on the shape always has the same encoding,
no matter the pose of the object and the camera view. 
The NNOPCS map can thus be viewed as a generalization of a NOCS map~\cite{NOCS}.

\subheading{NNOPCS Maps for Articulated Shapes}
Similar to~\cite{NePu}, a coordinate on the object is defined via a 3D vector for each pixel of the target view with width~$W$ and height~$H$, thus the NNOPCS map is a function of the form
\begin{equation}
\begin{aligned}
    \mathcal{P}_{\mathbf{E}, \mathbf{K}}: \mathbb{R}^{K \times 3} \times \mathbb{R}^{K \times d_f} \times \mathbb{R}^{d_z} &\rightarrow [0,1]^{H \times W \times 3}.
\end{aligned}
\label{eq:nnopcs_map_function}
\end{equation}
Here, $\mathbf{E}, \mathbf{K}$ are the given camera extrinsics and intrinsics, respectively.
The architecture remains the same, so we refer to~\cite{NePu} for details on how the function is implemented as an
attention-based neural network.
Similar to~\cite{NOCS}, the interpretation of the 3D output of the function in~\cref{eq:nnopcs_map_function} is a dense pixel-NNOPCS correspondence.
Ground truth data to learn the NNOPCS maps for articulated shapes end-to-end is rendered via Blender (www.blender.org), see section~\ref{sec:dataset}.
%
%
%

\subsection{Pipeline for Texture}
\label{sec:texture_pipeline}
\subheading{Texture Encoder}
The texture of a shape is described by a~$d_z$-dimensional global texture latent code~$\mathbf{z}_{texture}$ (same dimension as~$\mathbf{z}$), which augments the global geometry latent code~$\mathbf{z}$, cf.~\cref{fig:architecture_overview}. An encoder network~$enc$ computes~$\mathbf{z}_{texture}$ from
a masked input image~$\mathbf{I}$ together with the NNOPCS map~$\mathbf{P}$ (cf.~\cref{sec:positional_encoding}) 
for a certain camera position,
\begin{equation}
\begin{gathered}
    enc: \mathbb{R}^{H\times W\times 3} \times \mathbb{R}^{H\times W\times 3} \rightarrow \mathbb{R}^{d_z},\\
    (\mathbf{I}, \mathbf{P}) \mapsto \mathbf{z}_{texture}
    \label{eq:texture-encoder}
\end{gathered}
\end{equation}
%
%
The texture encoder is implemented as a convolutional neural network with~$23$ convolution and two dense layers.
Between the layers we use batch normalization~\cite{BatchNorm} and ReLu activations.
Similar to a variational autoencoder~\cite{Kingma2014}, the texture encoder outputs a mean~$\mathbf{\mu} \in \mathbb{R}^{d_z}$
and standard deviation $\mathbf{\sigma}^2 \in \mathbb{R}^{d_z}$
from which we sample the global texture latent code $\mathbf{z}_{texture}$.
%
%

During training, we use the available ground truth NNOPCS maps (cf.~\cref{sec:positional_encoding}) and silhouette masks. 
During inference, the silhouette mask and the NNOPCS map is estimated with the network trained above, cf.~\cref{fig:architecture_overview}. Both the images as well as the NNOPCS maps are set to zero outside the
shape's silhouette.

\subheading{Neural Texture Rendering}
%
Similar to how we render the NNOPCS maps (cf.~\cref{sec:positional_encoding}), we first decode the global texture embedding $\mathbf{z}_{texture}$
from~\eqref{eq:texture-encoder} into local texture features with a decoder network
\begin{equation}
    dec: \mathbb{R}^{d_z} \rightarrow \mathbb{R}^{K\times d_f},~ \mathbf{z}_{texture} \mapsto \mathbf{f}_{texture},
\end{equation}
where~$\mathbf{f}_{texture}$ gives the $d_f$-dimensional local texture features for each keypoint.
Note that the dimension of~$\mathbf{f}_{texture}$ is the same as the local geometry features~$\mathbf{f}$, so the final renderer
for a 2D image seen from camera view~$c$ conditioned on~$\mathbf{z}_{texture}$ and keypoints~$\mathbf{x}$ is again (compare with~\cref{eq:nnopcs_map_function}) a function of the form
\begin{equation}
    \mathcal{C}_{\mathbf{E}, \mathbf{K}}: \mathbb{R}^{K \times 3} \times \mathbb{R}^{K \times d_f} \times \mathbb{R}^{d_z} \rightarrow [0,1]^{H \times W \times 3}.
    \label{eq:color-rendering-function}
\end{equation}

%

\section{Datasets and Training}
\label{sec:dataset}
\subheading{\dataabr Dataset}
For this proof of concept, we choose cows as an example shape because their textures can be clearly distinguished. We extend~\cite{NePu_dataset} with its Holstein cow texture by adding eleven additional cow textures.
For each texture we provide the same $910$ poses, each captured from the same $24$ perspectives, placed in three evenly sampled rings at different heights around the model.
This results in $21840$ views per texture and a total of $262080$ views.
Each view consists of the ground truth 3D and 2D keypoints, the rendered RGB image, a silhouette of the cow, a depth map and a NNOPCS map, cf.~\cref{sec:positional_encoding}.
Each view has a resolution of $1024 \times 768$ px.
All images were rendered using Blender (www.blender.org) and the Cycles rendering engine.
We provide the same $25$ keypoints distributed among the joints of the cow as~\cite{NePu_dataset}.

%
Our synthetic texture dataset also contains instances of occlusions.
We generated $50$ instances captured from two of the $24$ camera views.
In these samples the Holstein (texture $0$) partially occludes the Limousine (texture $11$) cow.
For these cases we provide the occluded and complete silhouettes.
%
%

%
%

\subheading{H36M Dataset}
This human real-world dataset contains keypoint annotations, segmentations masks and RGB images of eleven actors in 17 scenarios, cf.~\cite{H36M}.
We use the first $910$ frames of the ``Posing'' scenario from $7$ subjects (S1, S5, S6, S7, S8, S9, S11) for this study, similar to~\cite{AniNeRF}, and all four camera views for training.
%

\subheading{Training the Pipeline for Geometry}
To train the NNOPCS maps of articulated shapes, cf.~\cref{sec:positional_encoding}, we use the same regimen as in~\cite{NePu}.
However, instead of color images we train end-to-end with the ground truth NNOPCS map observations
\begin{equation}
\mathbf{P}_{m,c},\; m\in\{1,\dots,M\},\; c\in\{1,\dots,C\}
\end{equation}
over~$M$ distinct poses captured by~$C$ different cameras.
All model parameters are trained jointly. In the composite loss from~\cite{NePu}, instead of the color loss~$\mathcal{L}_{\text{col}}$ we define
\begin{equation}
    \mathcal{L}_{\text{NNOPCS}} = \sum_{m=1}^M\sum_{c=1}^C \Vert \mathcal{P}_{\mathbf{E}_{c}, \mathbf{K}_{c}} - \mathbf{P}_{m,c}\Vert_2^2,
\end{equation}
which is the squared pixel-wise difference over all three channels.
For the details of this equation and the composite loss, we refer to~\cite{NePu}.
%

\subheading{Training the Pipeline for Texture}
While training our neural texture encoder, decoder and renderer (cf.~\cref{fig:architecture_overview}, bottom), we keep the weights of the pipeline for geometry (cf.~\cref{fig:architecture_overview}, top)
fixed and use the ground truth NNOPCS as the fixed geometry.
We thus learn only the texture relevant weights of the pipeline in this step.

Assuming the dataset has $T$~textures with images~$\mathbf{I}_{t,m,c}$ provided for
each texture and the same poses and cameras as the NNOPCS maps, we define 
the color rendering loss
\begin{equation}
    \mathcal{L}_{\text{col}} = \sum_{t=1}^T\sum_{m=1}^M\sum_{c=1}^C \Vert \mathcal{C}_{\mathbf{E}_{c}, \mathbf{K}_{c}} - \mathbf{I}_{m,c,t}\Vert_2^2
    \label{eq:loss_col}
\end{equation}
as the squared pixel-wise difference over all color channels, where $\mathcal{C}_{\mathbf{E}, \mathbf{K}}$
is the color rendering function from~\cite{NePu} defined in~\eqref{eq:color-rendering-function}.
To regularize the texture latent space, we employ the
Kullback–Leibler divergence~\cite{kullback1951information}~$\mathcal{L}_{\text{KLD}}$
between the predicted multi-dimensional normal distribution and the standard normal distribution.
We enforce this loss with the same sampling scheme and reparametrization trick as in the training
of a variational autoencoder~\cite{Kingma2014}.
The total loss is thus
\begin{equation}
    \mathcal{L} =  \lambda_{\text{col}}\mathcal{L}_{\text{col}} +
                  \lambda_{\text{KLD}}\mathcal{L}_{\text{KLD}},
    \label{eq:training_loss_texture}
\end{equation}
where the different positive numbers~$\lambda$ are hyperparameters to balance the influence of the two losses.

\section{Experiments}
In~\cref{sec:experiments:implementation_details} we provide implementation details of our architecture from~\cref{sec:architecture}.
We then evaluate the learned NNOPCS maps (cf.~\cref{sec:positional_encoding}) in~\cref{sec:experiments:positional_encoding}.
In~\cref{sec:experiments:neural_texture_rendering} we show novel pose synthesis on our novel~\dataabr and the H36M~\cite{H36M} dataset and provide texture reconstructions of a real-world example with a model trained on synthetic data only.
In~\cref{sec:experiments:reID} you find our kernel density estimation (KDE)~\cite{rosenblatt1956remarks,parzen1962on} of the t-SNE~\cite{JMLR:v9:vandermaaten08a} of the global texture embedding that we compare to the embedding of~\cite{he2021transreid}.

\subsection{Implementation Details}
\label{sec:experiments:implementation_details}
To train \paperabr, we extend the training procedure and parameters from~\cite{NePu}.
We do not render complete NNOPCS (cf.~\cref{sec:positional_encoding}) and texture maps to compute $\mathcal{L}_{\text{NNOPCS}}$ and $\mathcal{L}_{\text{col}}$ during training, as this would be too costly.
Instead, we choose~$500$ points in each iteration that we sample uniformly within the ground truth mask.
For the cow shape we increase the number of nearest neighbours in the vector cross-attention (VCA) module from~$12$ to~$20$
as we observe better performance (cf.~\cref{tab:quantitative-pos-enc-depth}).

\subheading{Pipeline for Geometry}
While training the NNOPCS maps, we also learn to reconstruct masks and depth, just as in~\cite{NePu}.
However, in contrast to~\cite{NePu}, we choose to use the 2-norm (vs. MSE in~\cite{NePu}) as depth loss during training
which leads to better results in terms of MAE, cf.~\cref{tab:quantitative-pos-enc-depth}.

\begin{table}[t]
    \centering
    \small
    \begin{tabular}{c|c}
        \toprule
        NNOPCS Maps MSE & Depth MAE [mm] \\
        
        \begin{tabular}{c}
        Ours \\
        \midrule
        $1.6\cdot 10^{-4}$ \\
        \end{tabular} &
        
        \begin{tabular}{cac}
        NePu~\cite{NePu} & Ours & $\Delta$ \\
        \midrule
        $22.3$ & $\mathbf{17.8}$ & $-4.5$ \\
        \end{tabular} \\
        
        \bottomrule
    \end{tabular}
    \caption{{\em Quantitative results for the NNOPCS and depth maps on the~\dataabr test set}.
         We report MSE for the reconstructed NNOPCS maps (cf.~\cref{sec:positional_encoding}) and MAE [mm] for the reconstructed depth maps.
         Comparison of the reconstructed depth map between NePu~\cite{NePu} and our method.
         See~\cref{sec:experiments:positional_encoding} for a discussion of the results.
    }
    \label{tab:quantitative-pos-enc-depth}
\end{table}
%
\subheading{Pipeline for Texture}
We train the neural texture rendering pipeline
using an initial learning rate of $5e^{-4}$, which is decayed
with a factor of $0.2$ on our novel~\dataabr dataset and $0.9$ on the H36M dataset~\cite{H36M} every $500$ epochs.
We choose final weights based on the minimum validation loss, i.e. epoch $535$ for our~\dataabr and $2080$ for the H36M dataset.
We weight the training loss in \eqref{eq:training_loss_texture} with $\lambda_{\text{col}}=5$ and $\lambda_{\text{KLD}}=1e^{-8}$.

\subsection{NNOPCS Maps}
\label{sec:experiments:positional_encoding}
\begin{figure}[t]
    \centering
    \begin{subfigure}[b]{\linewidth}
        \centering
        \includegraphics[width=0.49\textwidth]{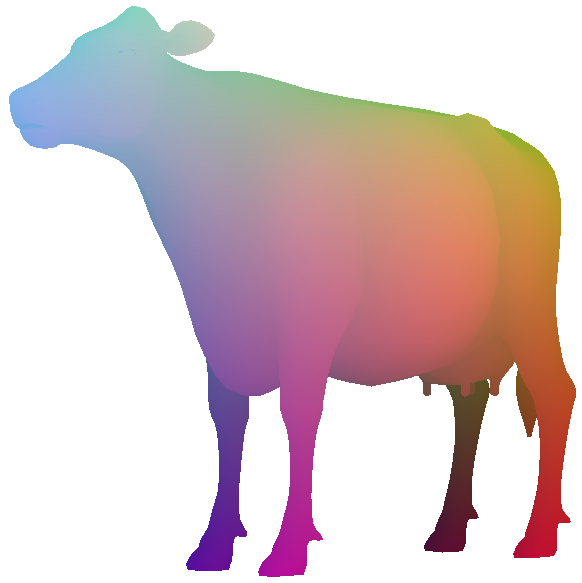}
        \includegraphics[width=0.49\textwidth]{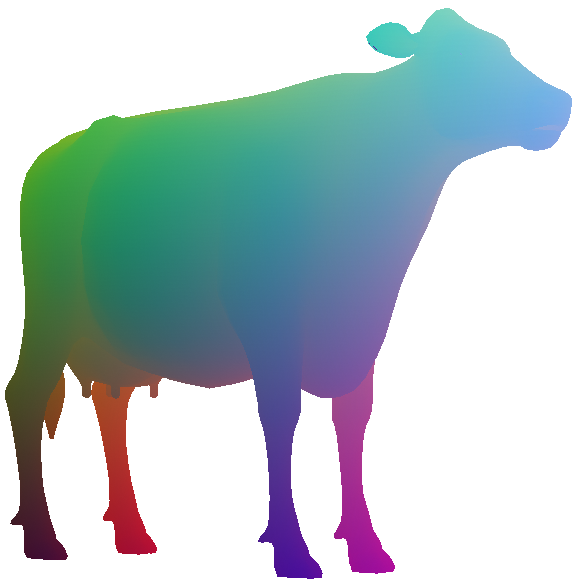}
        \caption{
        \textbf{Neutral Pose.}
        %
        %
        The color at each point on the mesh depends on its position in $x$-, $y$-, and $z$-direction in the neutral pose. Thus, for other poses the color at a specific position on the mesh remains the same, uniquely describing this point.}
        \label{fig:nnopcs_map_neutral_pose}
    \end{subfigure}
    \hfill
    \begin{subfigure}[b]{\linewidth}
        \centering
        \includegraphics[width=0.49\textwidth]{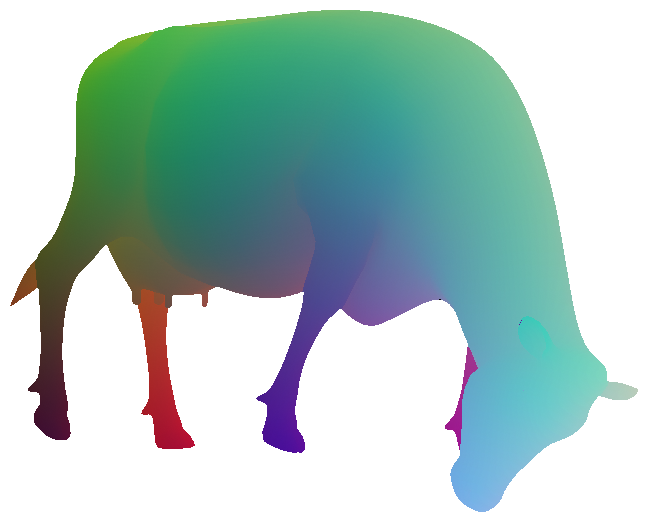}
        \includegraphics[width=0.49\textwidth]{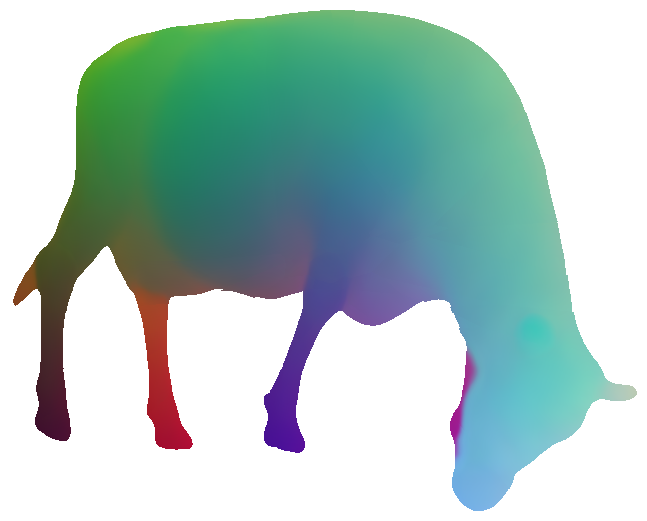}
        \caption{
        \textbf{Novel Pose Synthesis.}
        One example of ground truth (left) and reconstruction (right) from our~\dataabr test set.
        Our reconstruction (right) reflects the overall high quality of our quantitative results in~\cref{tab:quantitative-pos-enc-depth}.
        }
        \label{fig:nnopcs_map_example}
    \end{subfigure}
    \caption{
        \textbf{NNOPCS Maps for Articulated Shapes.}
        We show the neutral pose of the NNOPCS (cf.~\cref{sec:positional_encoding}) projected on two image planes in (a) and one example for a novel pose from our~\dataabr test set in (b).
        }
\end{figure}
Quantitative results for the NNOPCS map (cf.~\cref{sec:positional_encoding}) estimation are shown in~\cref{tab:quantitative-pos-enc-depth}.
Over all cow test samples,
we achieve a mean squared error (MSE) of $1.6\cdot 10^{-4}$.
Since the coordinates of the NNOPCS are normalized to~$[0,1]$ this means that our learned NNOPCS maps have
an error of $1\%$ on average.
This overall high quality is also reflected in our qualitative example
in~\cref{fig:nnopcs_map_example},
and is a necessary step in order to achieve accurate texture renderings.
%

In~\cref{tab:quantitative-pos-enc-depth}, we also compare depth map reconstructions of~\paperabr to the original~NePu~\cite{NePu}
in terms of the mean absolute error (MAE) in $mm$ over all cow test samples.
We achieve a lower MAE by~$4.5mm$, compared to the length and height of the cow of~$2.2m$ and $1.65m$, respectively.

\subsection{Neural Texture Rendering}
\label{sec:experiments:neural_texture_rendering}
\begin{figure*}[t]
    \centering
    \begin{subfigure}[b]{0.24\linewidth}
        \centering
        \includegraphics[width=0.49\textwidth,trim={9cm 3cm 9cm 3cm},clip]{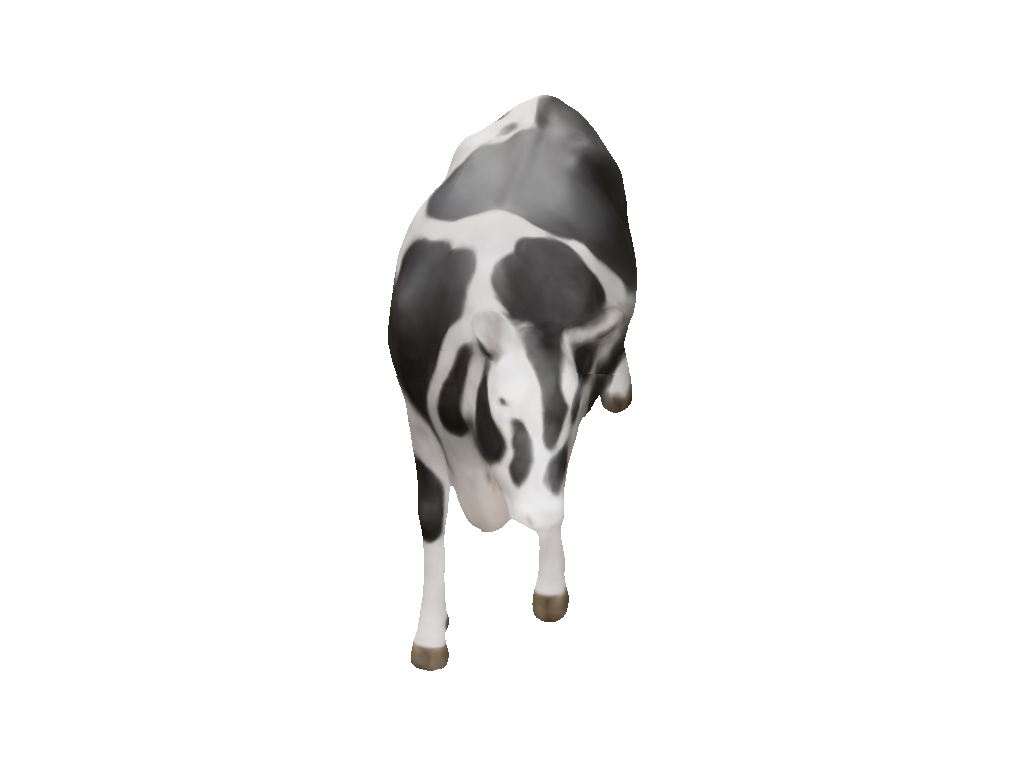}
        \includegraphics[width=0.49\textwidth,trim={8cm 3cm 9cm 0.5cm},clip]{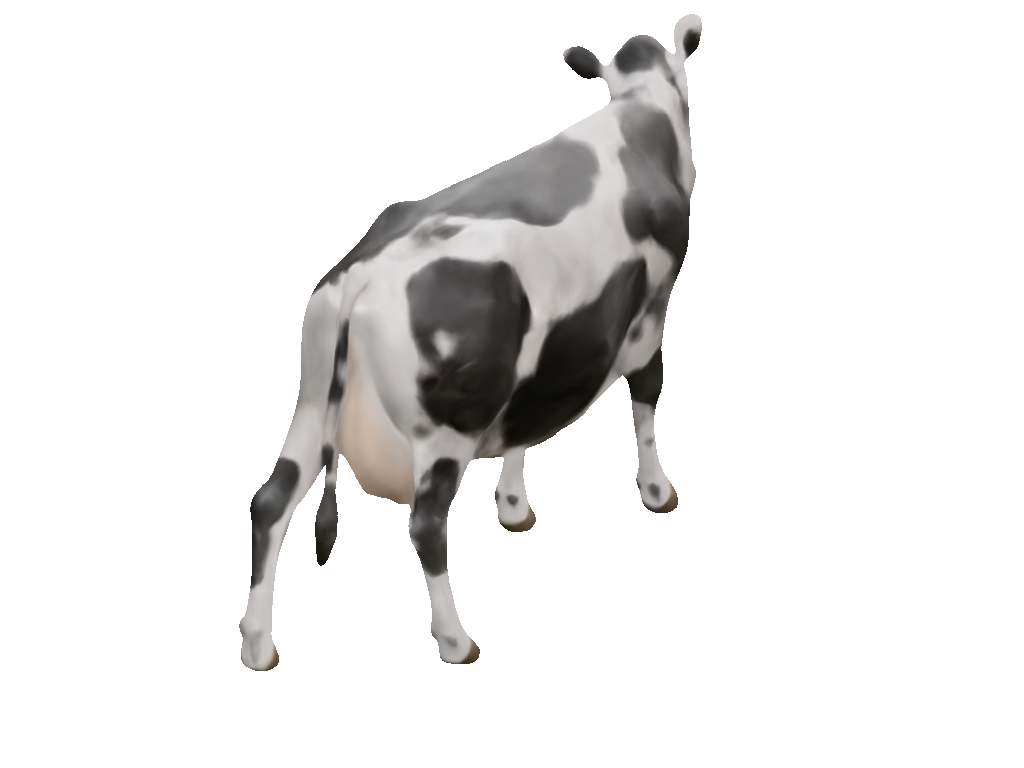}
        \caption{Texture $0$}
        \label{fig:npnv:0}
    \end{subfigure}
    \hfill
    \begin{subfigure}[b]{0.24\linewidth}
        \centering
        \includegraphics[width=0.49\textwidth,trim={9cm 3cm 9cm 3cm},clip]{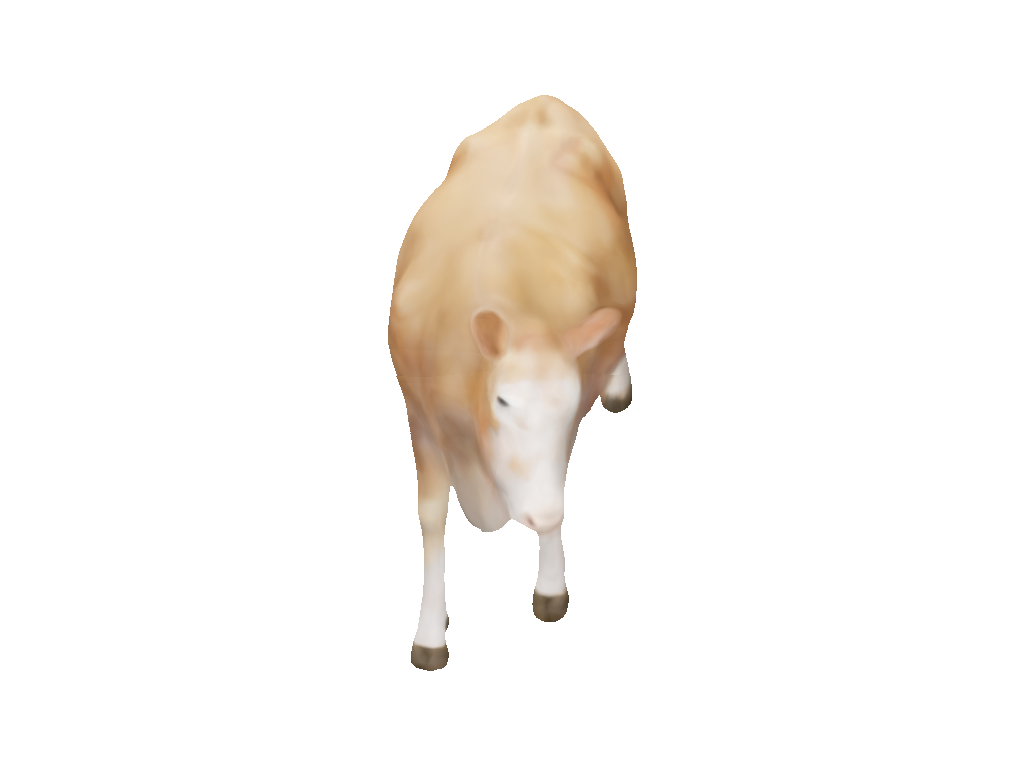}
        \includegraphics[width=0.49\textwidth,trim={8cm 3cm 9cm 0.5cm},clip]{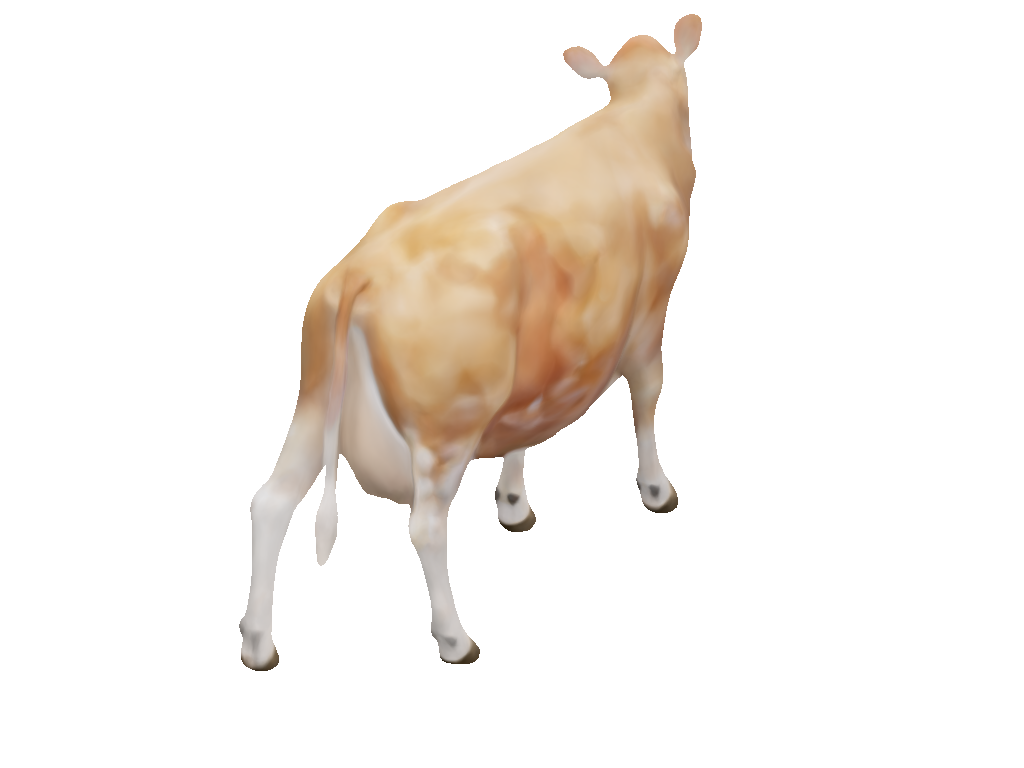}
        \caption{Texture $1$}
        \label{fig:npnv:1}
    \end{subfigure}
    \hfill
    \begin{subfigure}[b]{0.24\linewidth}
        \centering
        \includegraphics[width=0.49\textwidth,trim={9cm 3cm 9cm 3cm},clip]{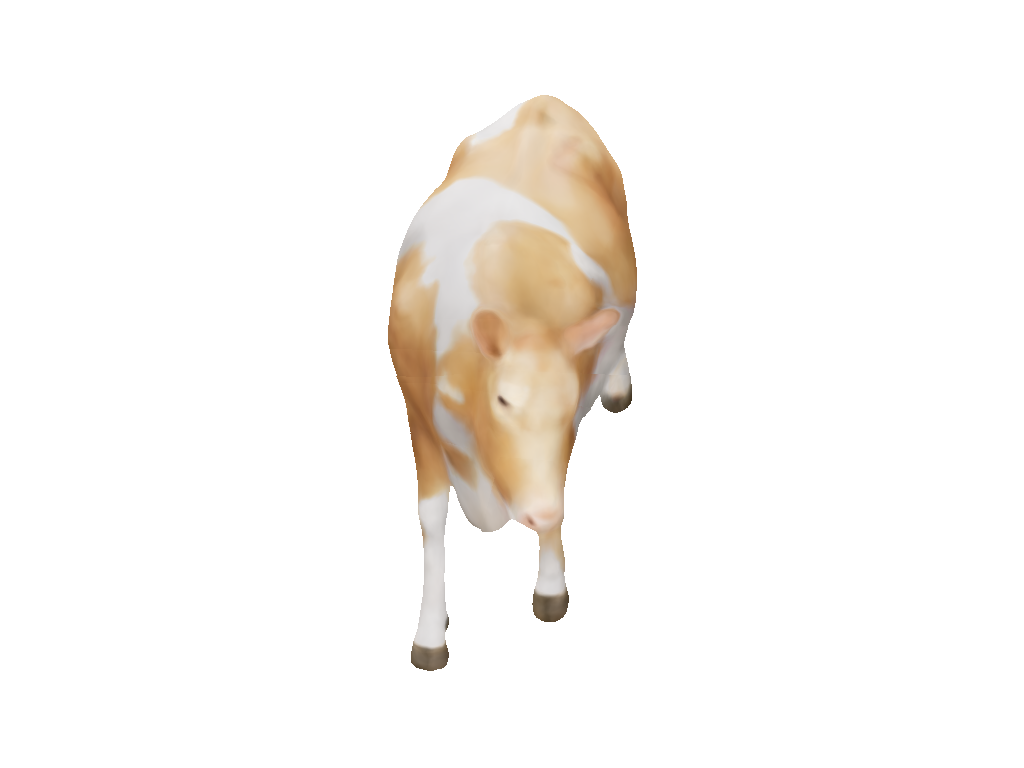}
        \includegraphics[width=0.49\textwidth,trim={8cm 3cm 9cm 0.5cm},clip]{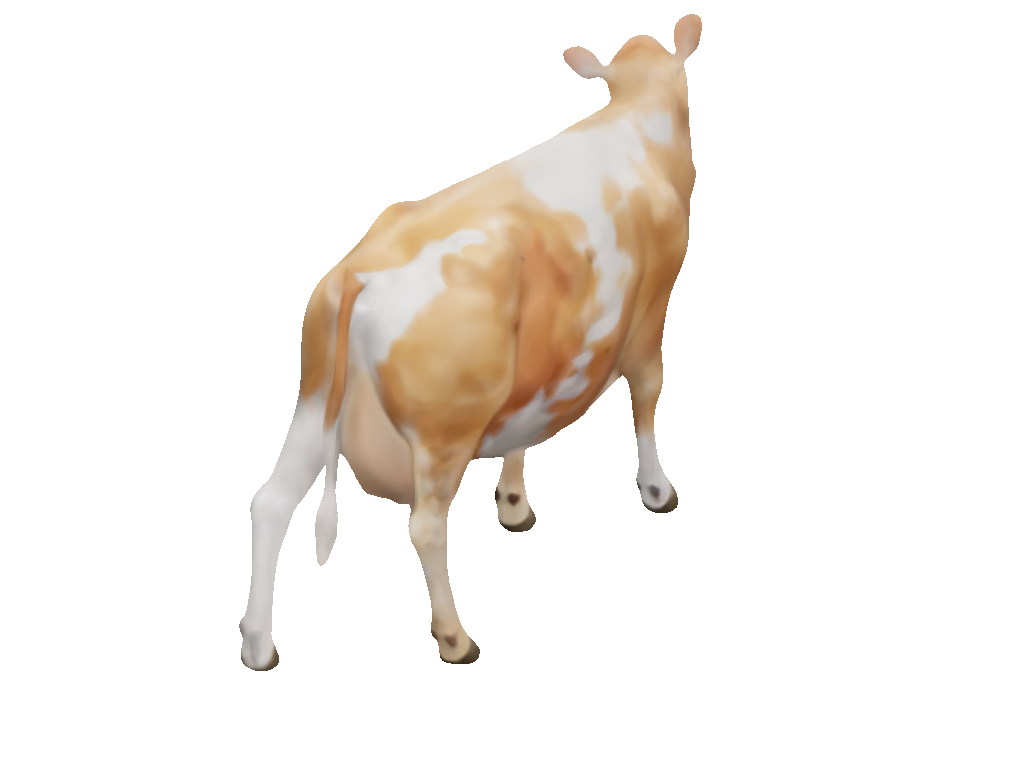}
        \caption{Texture $2$}
    \end{subfigure}
    \hfill
    \begin{subfigure}[b]{0.24\linewidth}
        \centering
        \includegraphics[width=0.49\textwidth,trim={9cm 3cm 9cm 3cm},clip]{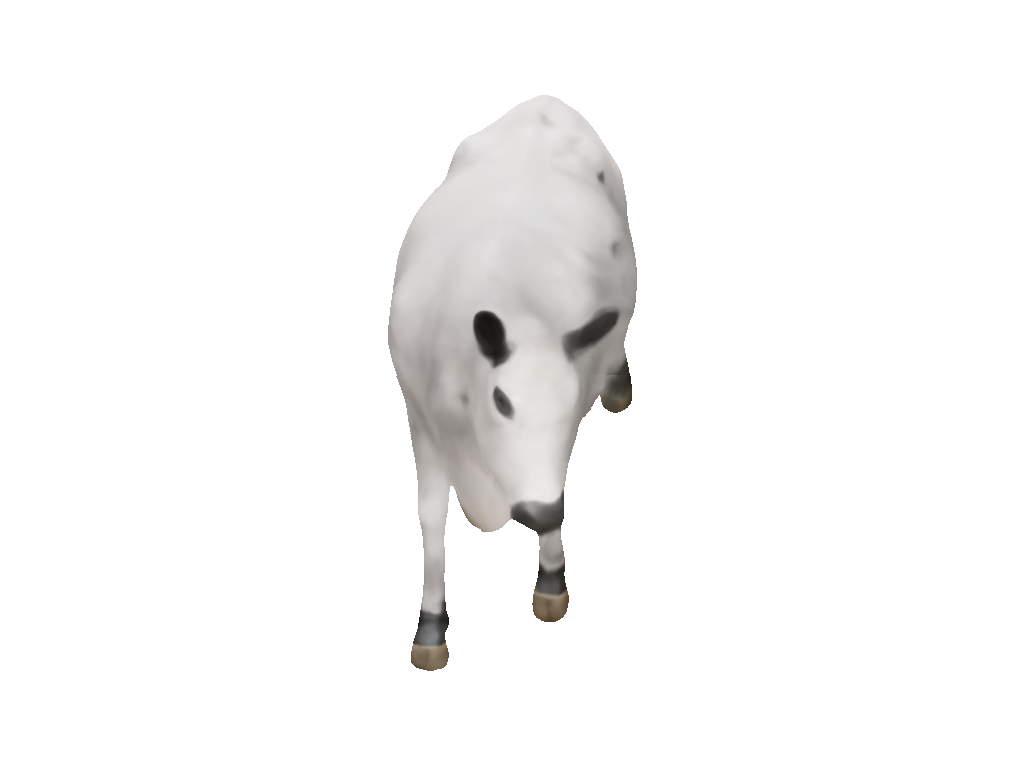}
        \includegraphics[width=0.49\textwidth,trim={8cm 3cm 9cm 0.5cm},clip]{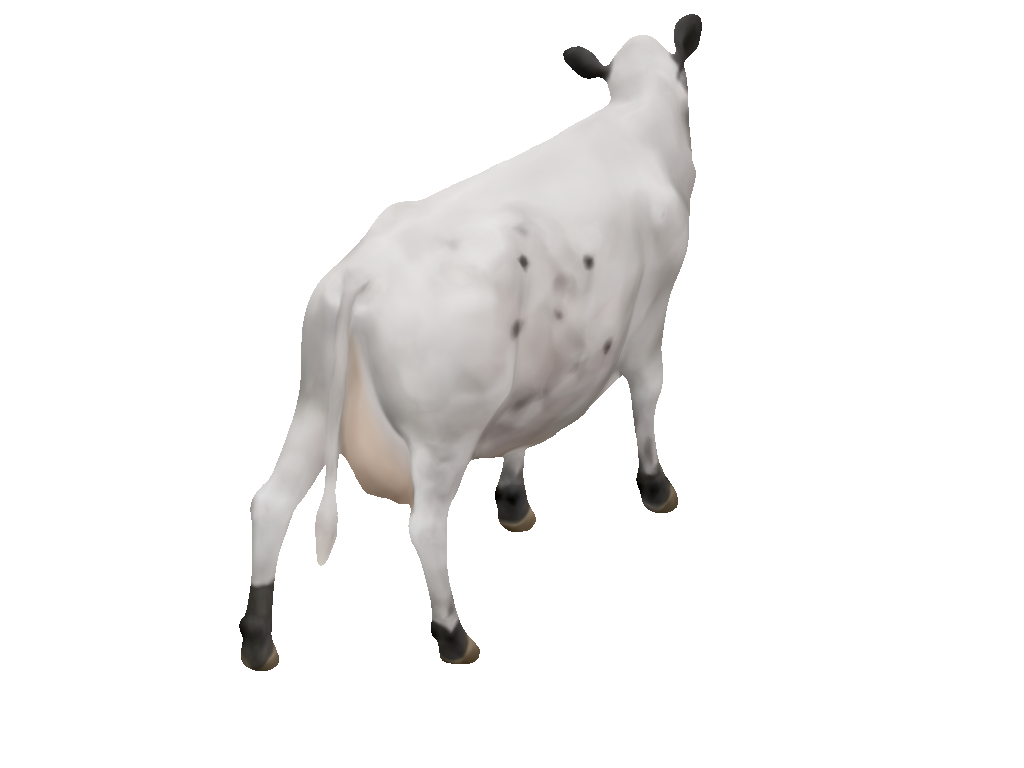}
        \caption{Texture $3$}
        \label{fig:npnv:3}
    \end{subfigure}
    \hfill
    \begin{subfigure}[b]{0.24\linewidth}
        \centering
        \includegraphics[width=0.49\textwidth,trim={9cm 3cm 9cm 3cm},clip]{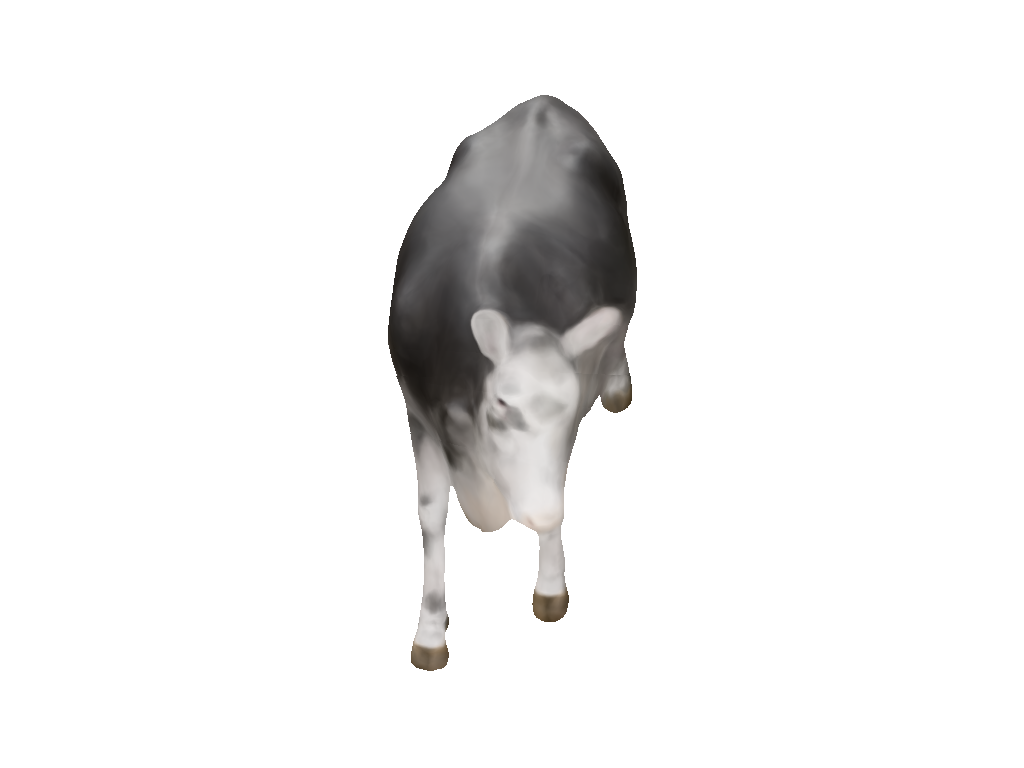}
        \includegraphics[width=0.49\textwidth,trim={8cm 3cm 9cm 0.5cm},clip]{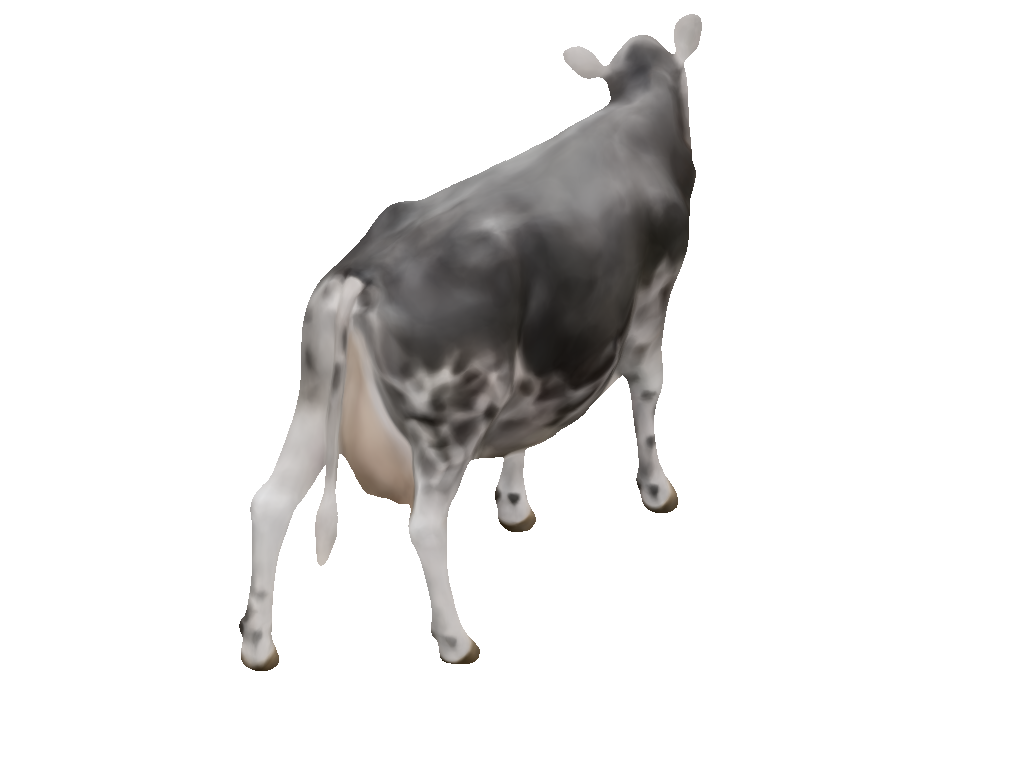}
        \caption{Texture $4$}
    \end{subfigure}
    \hfill
    \begin{subfigure}[b]{0.24\linewidth}
        \centering
        \includegraphics[width=0.49\textwidth,trim={9cm 3cm 9cm 3cm},clip]{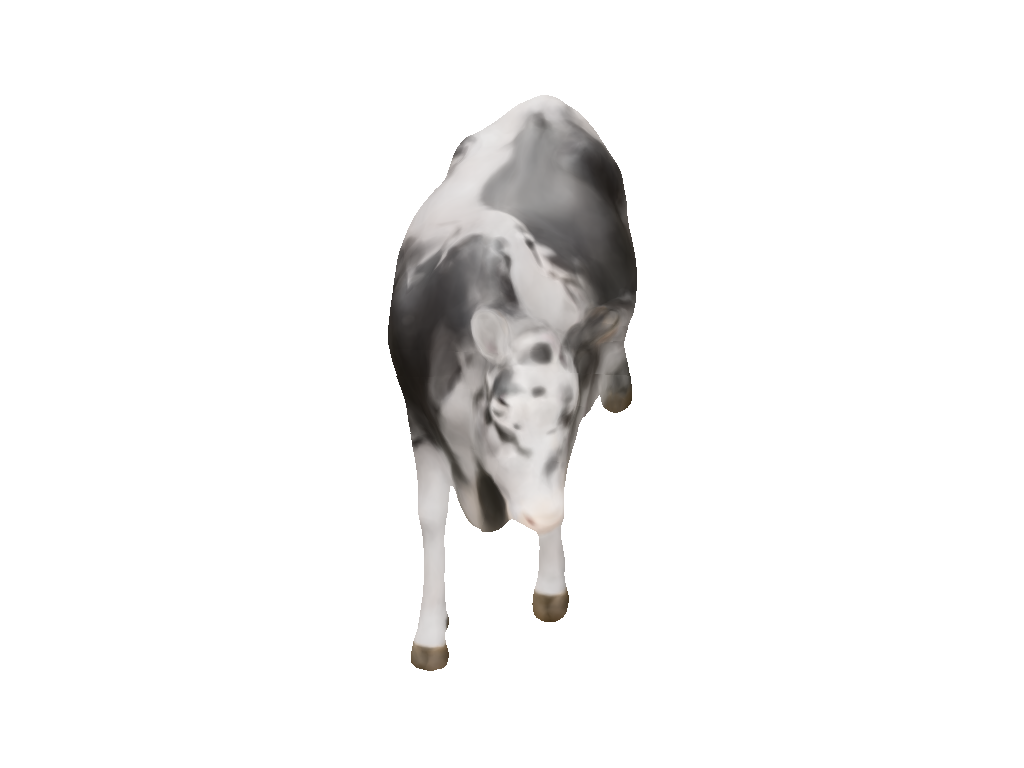}
        \includegraphics[width=0.49\textwidth,trim={8cm 3cm 9cm 0.5cm},clip]{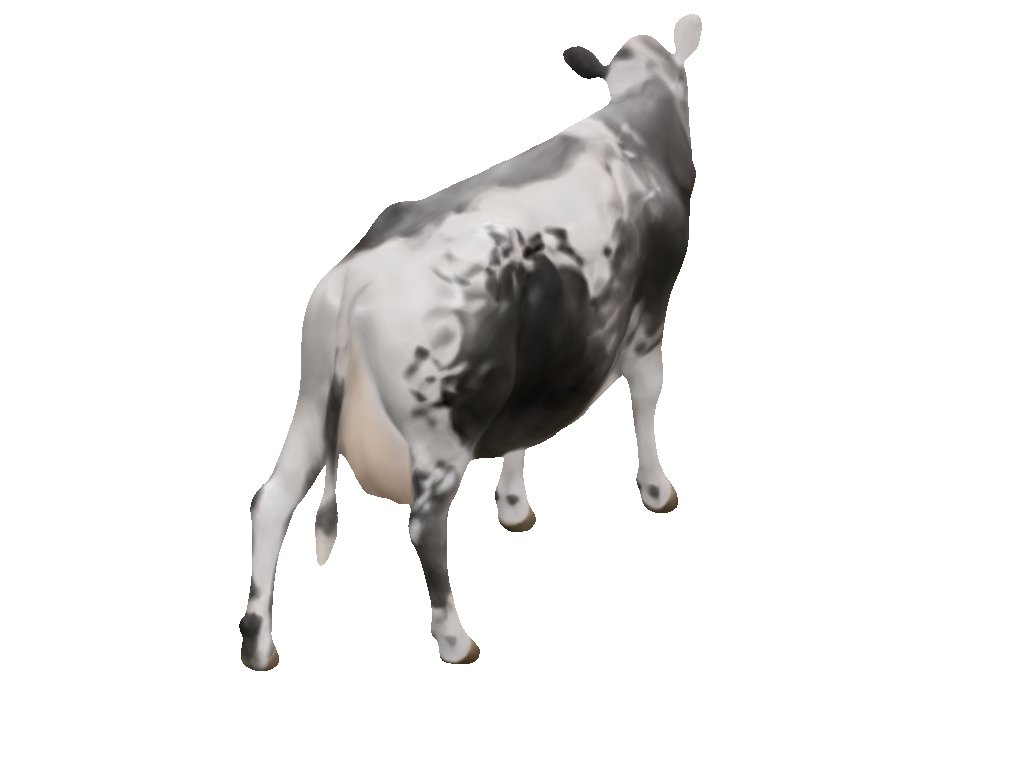}
        \caption{Texture $5$}
        \label{fig:npnv:5}
    \end{subfigure}
    \hfill
    \begin{subfigure}[b]{0.24\linewidth}
        \centering
        \includegraphics[width=0.49\textwidth,trim={9cm 3cm 9cm 3cm},clip]{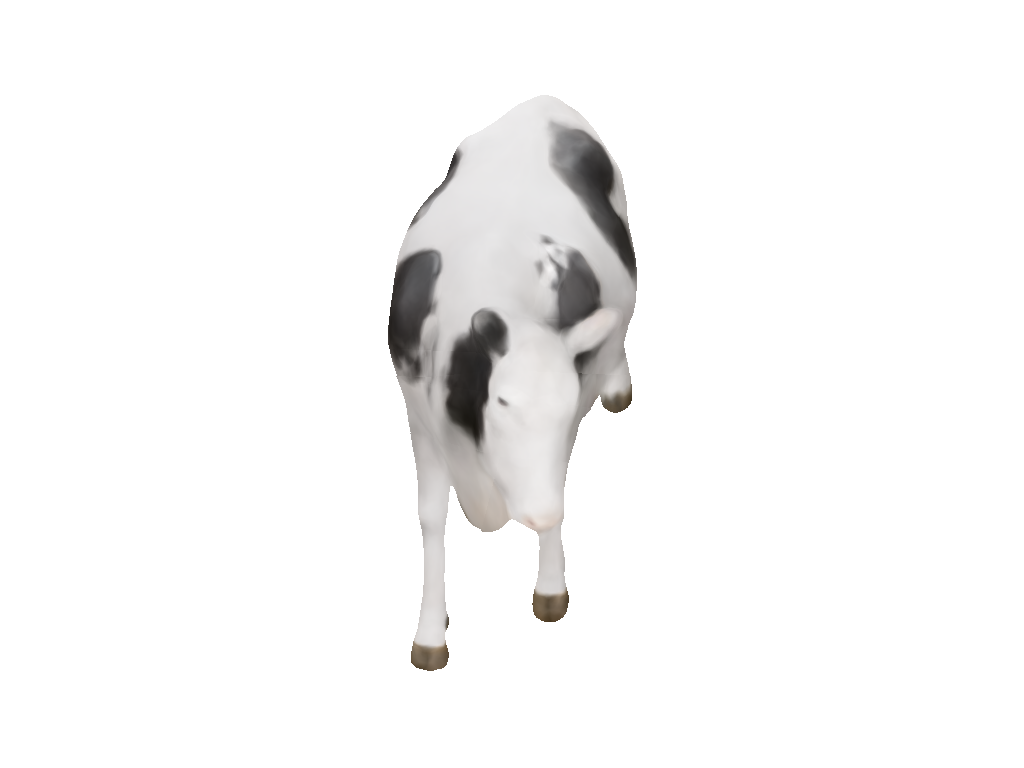}
        \includegraphics[width=0.49\textwidth,trim={8cm 3cm 9cm 0.5cm},clip]{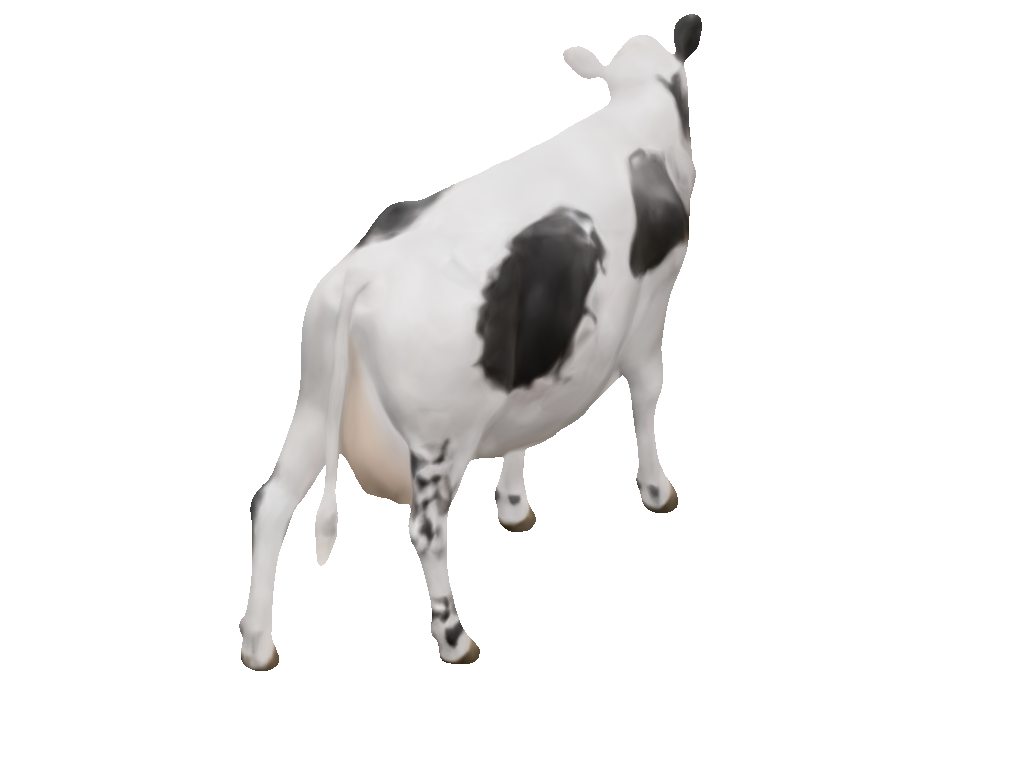}
        \caption{Texture $6$}
    \end{subfigure}
    \hfill
    \begin{subfigure}[b]{0.24\linewidth}
    \centering
        \includegraphics[width=0.49\textwidth,trim={9cm 3cm 9cm 3cm},clip]{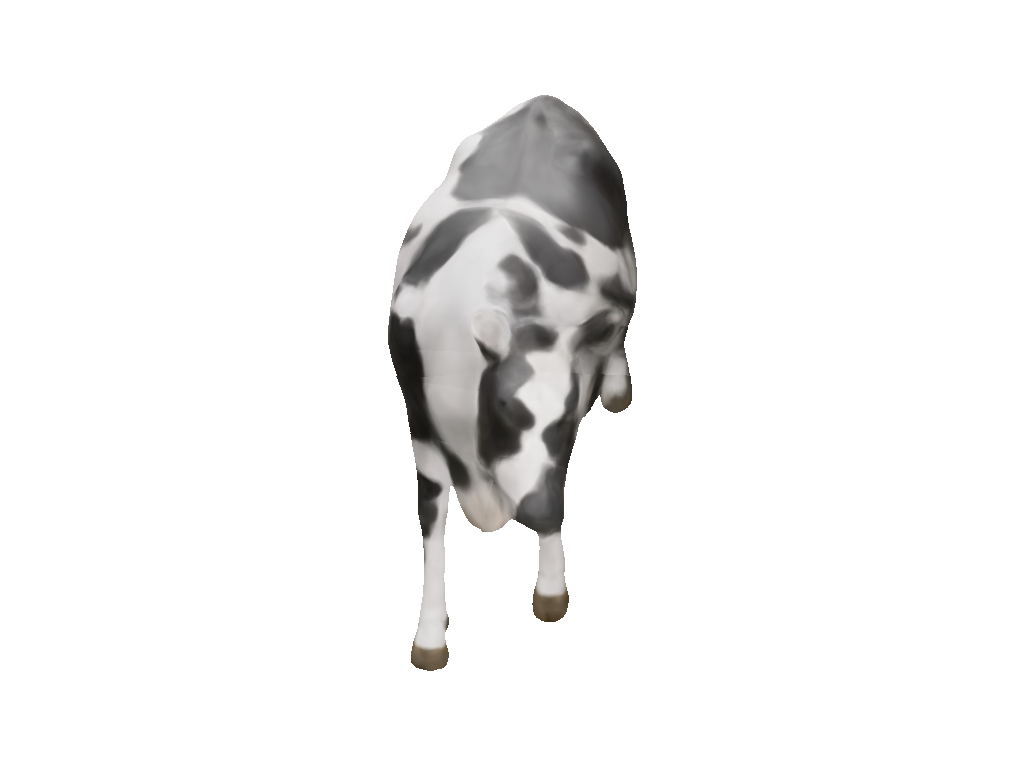}
        \includegraphics[width=0.49\textwidth,trim={8cm 3cm 9cm 0.5cm},clip]{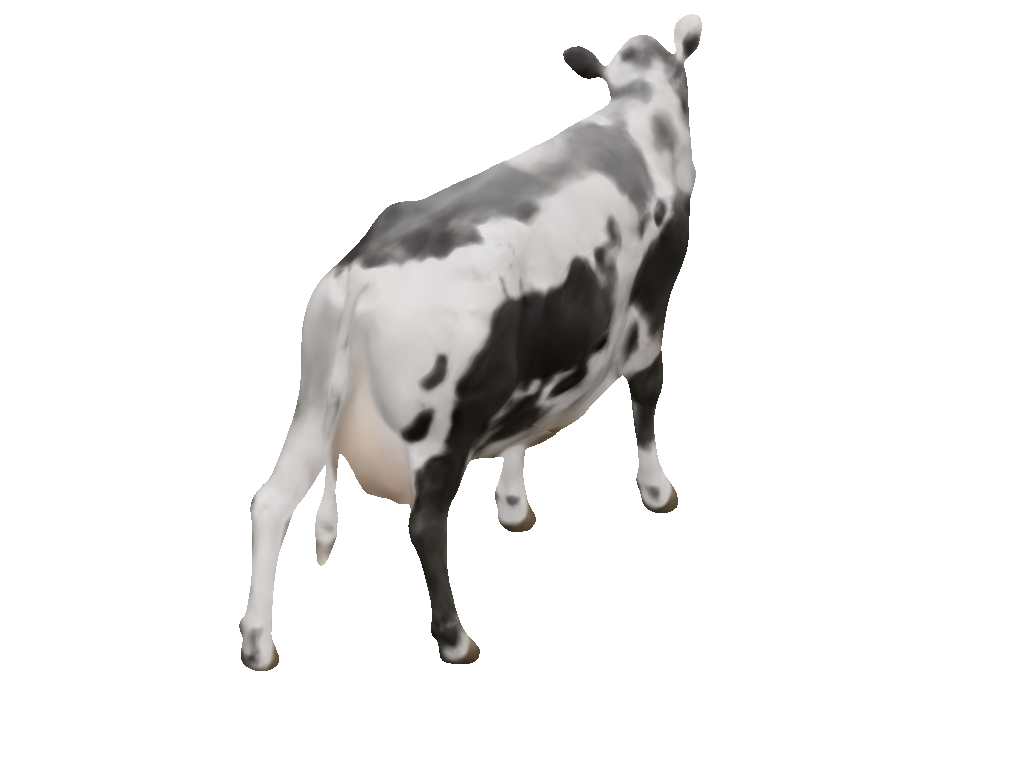}
        \caption{Texture $7$}
    \end{subfigure}
    \hfill
        \begin{subfigure}[b]{0.24\linewidth}
        \centering
        \includegraphics[width=0.49\textwidth,trim={9cm 3cm 9cm 3cm},clip]{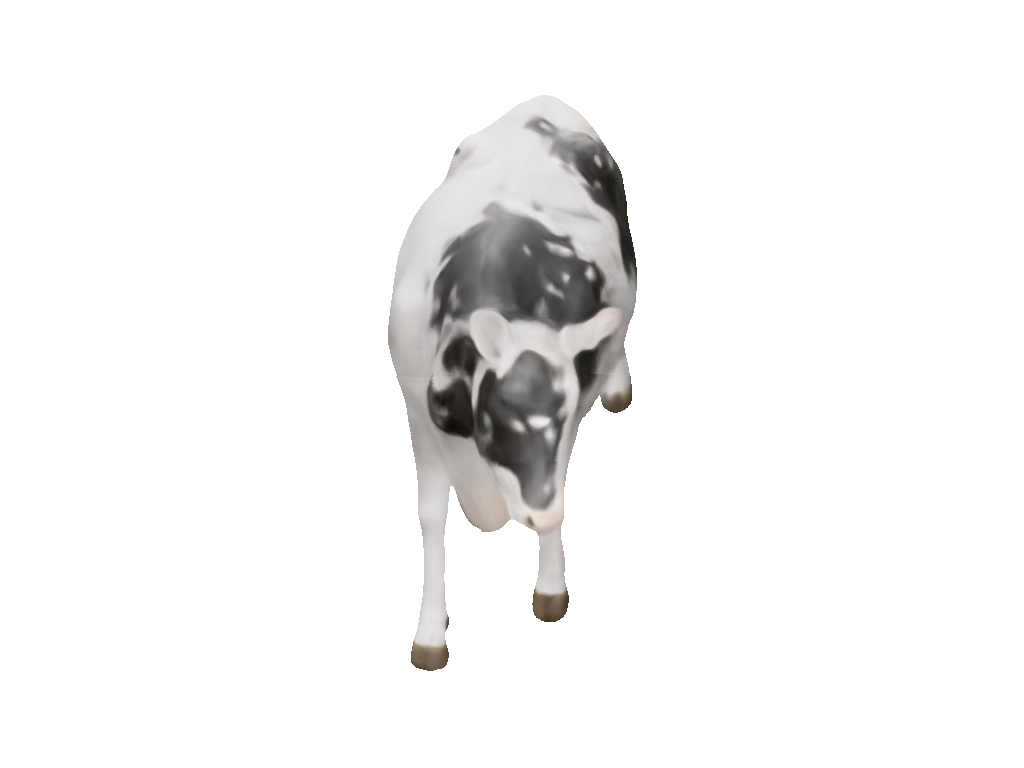}
        \includegraphics[width=0.49\textwidth,trim={8cm 3cm 9cm 0.5cm},clip]{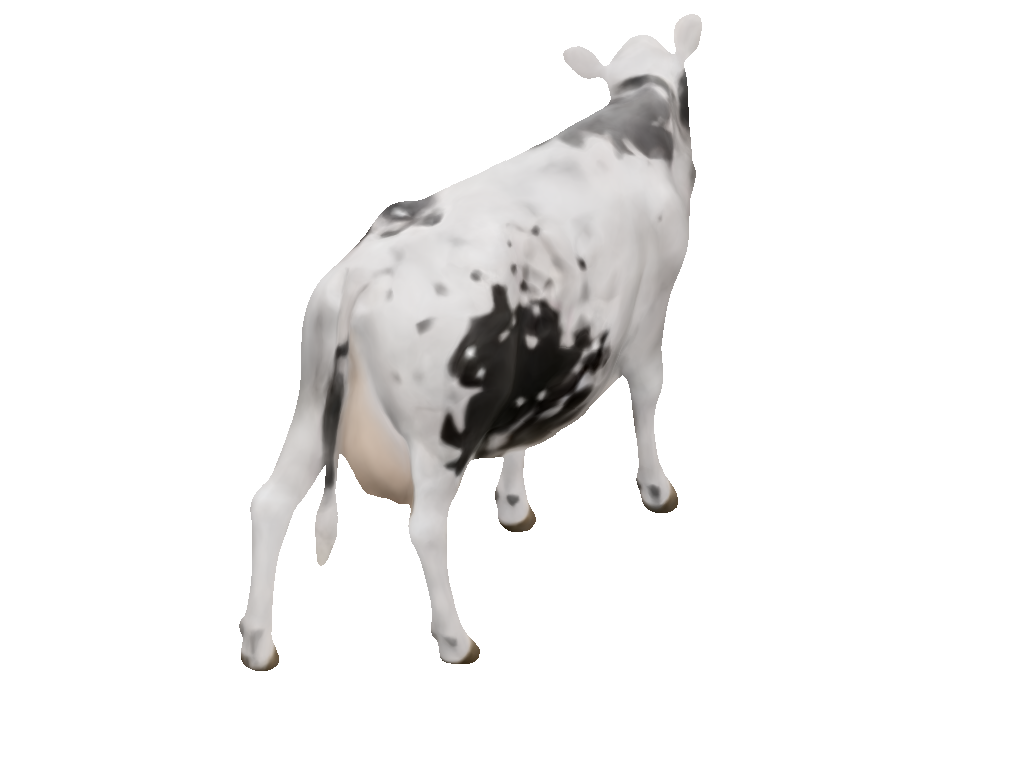}
        \caption{Texture $8$}
    \end{subfigure}
    \hfill
        \begin{subfigure}[b]{0.24\linewidth}
        \centering
        \includegraphics[width=0.49\textwidth,trim={9cm 3cm 9cm 3cm},clip]{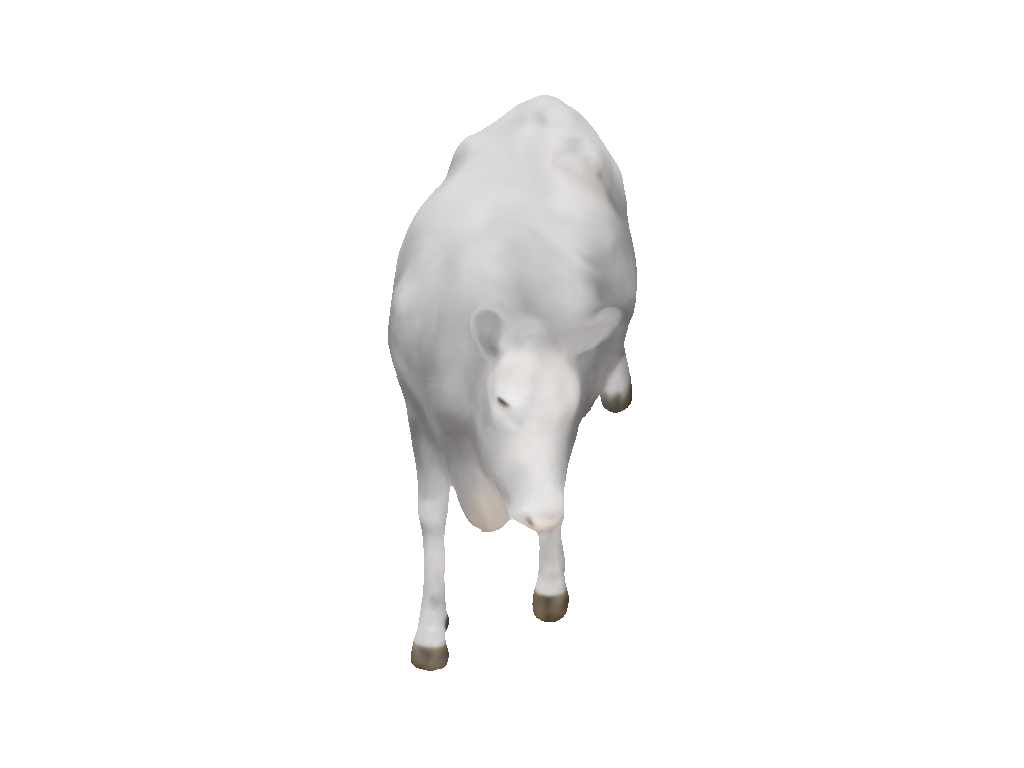}
        \includegraphics[width=0.49\textwidth,trim={8cm 3cm 9cm 0.5cm},clip]{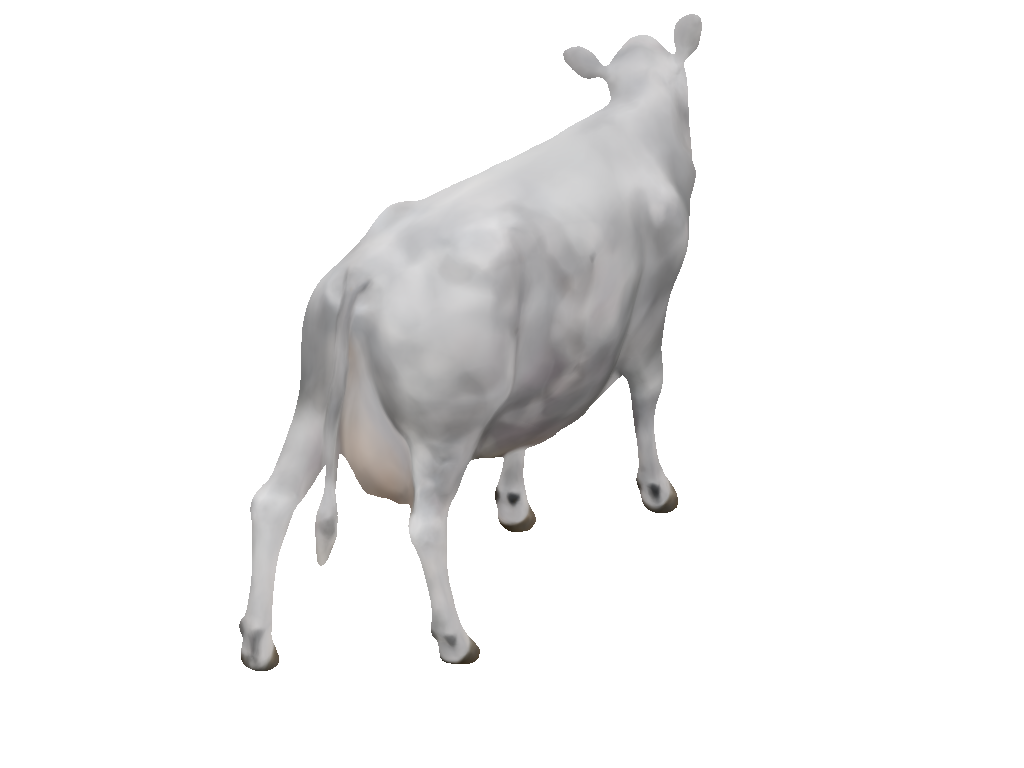}
        \caption{Texture $9$}
        \label{fig:npnv:9}
    \end{subfigure}
    \hfill
        \begin{subfigure}[b]{0.24\linewidth}
        \centering
        \includegraphics[width=0.49\textwidth,trim={9cm 3cm 9cm 3cm},clip]{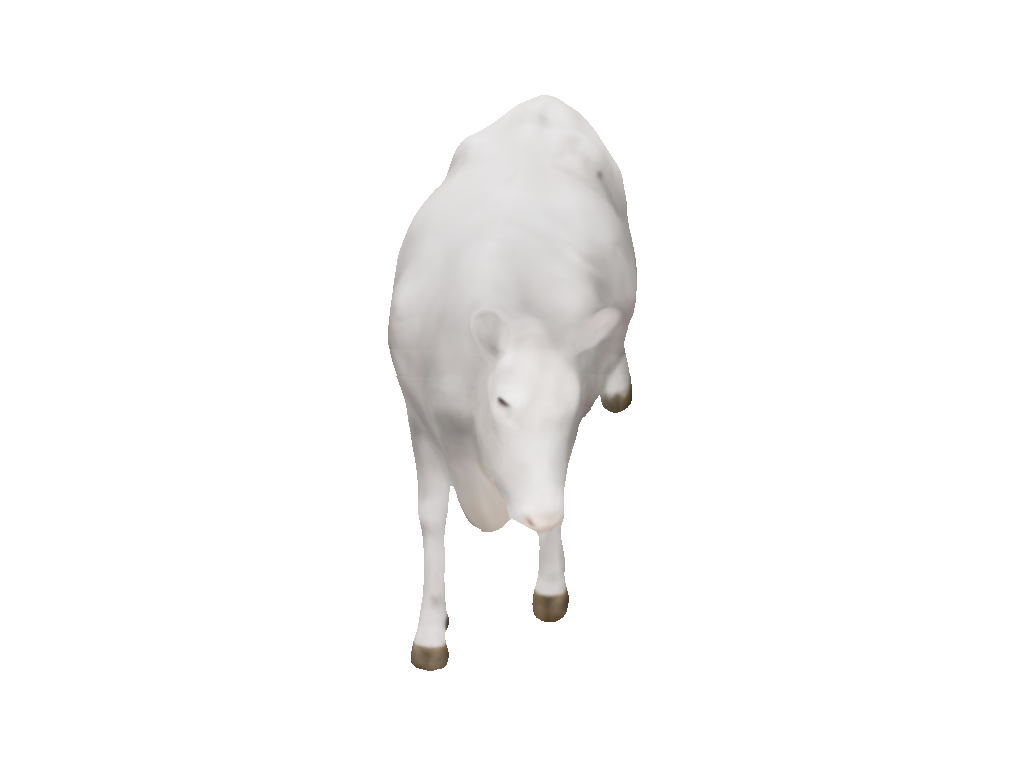}
        \includegraphics[width=0.49\textwidth,trim={8cm 3cm 9cm 0.5cm},clip]{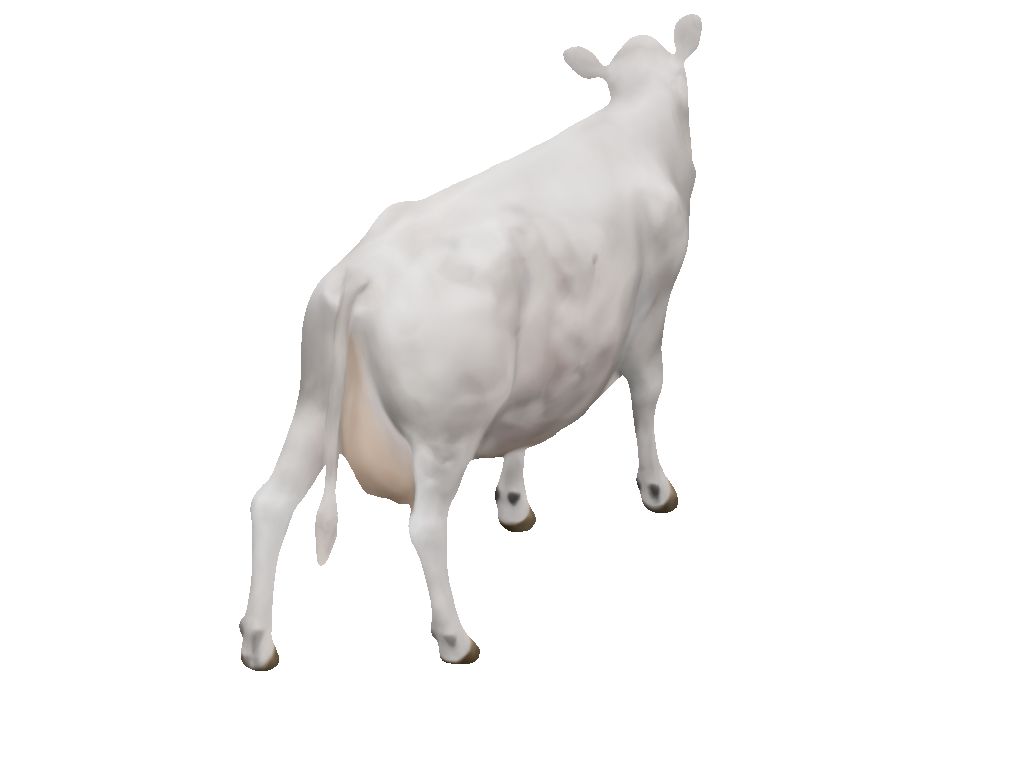}
        \caption{Texture $10$}
        \label{fig:npnv:10}
    \end{subfigure}
    \hfill
        \begin{subfigure}[b]{0.24\linewidth}
        \centering
        \includegraphics[width=0.49\textwidth,trim={9cm 3cm 9cm 3cm},clip]{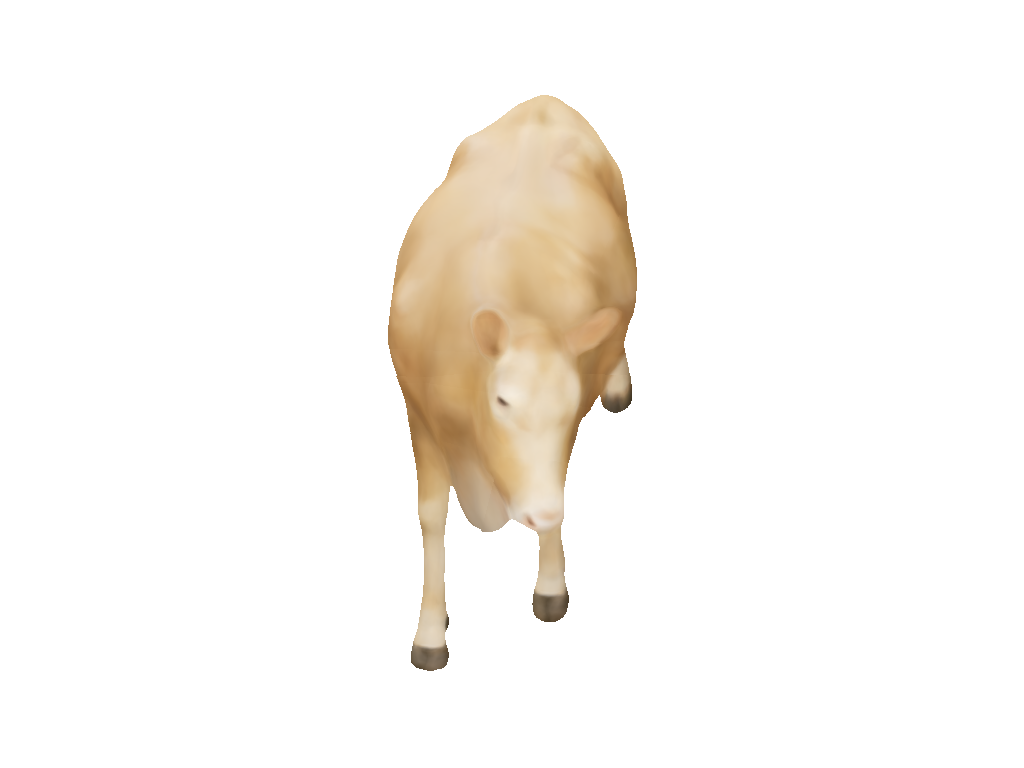}
        \includegraphics[width=0.49\textwidth,trim={8cm 3cm 9cm 0.5cm},clip]{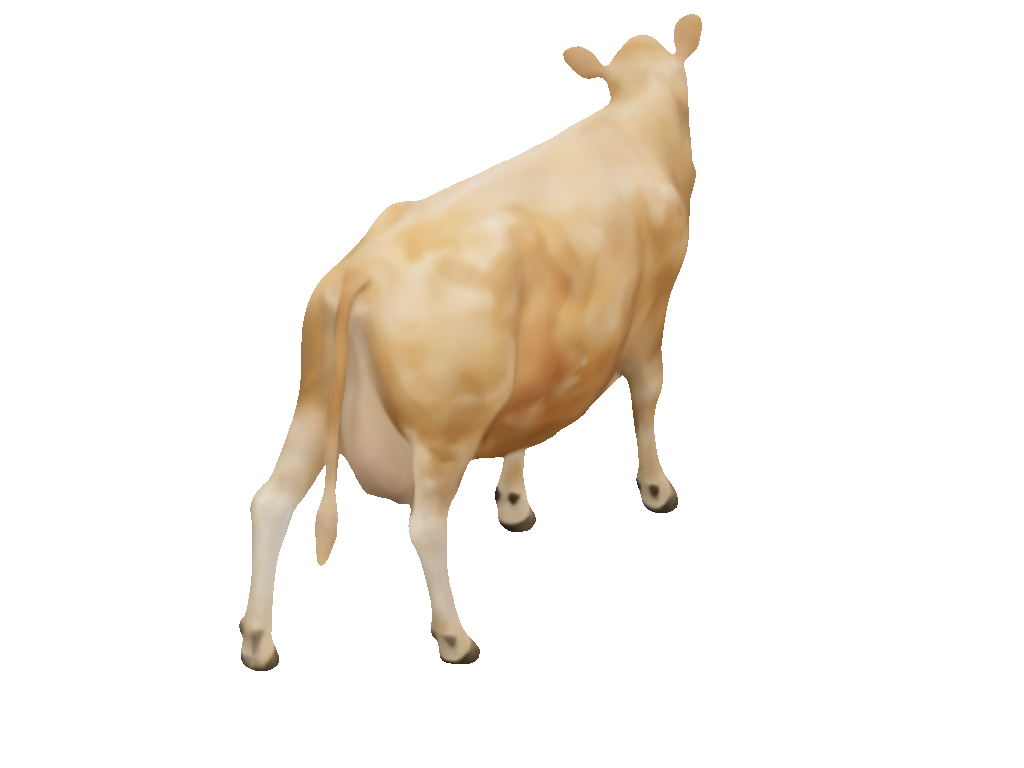}
        \caption{Texture $11$}
        \label{fig:npnv:11}
    \end{subfigure}
    \caption{
        \textbf{Novel Pose and Novel View Synthesis.}
        We reconstruct images for novel views and novel poses from our~\dataabr test set, which have not been seen during training.
        We show two examples for each distinct texture in our dataset.
        Our reconstructions reflect the overall good quality of our quantitative results in~\cref{tab:quantitative-color}.
    }
    \label{fig:novel-pose-novel-views}
\end{figure*}
\begin{table}[t]
    \centering
    \small
    \begin{tabular}{c|c}
        \toprule
        Dataset / Metric & Color PSNR [dB] \\
        \midrule
        H36M~\cite{H36M} & $15.98$ \\
        \dataabr & $19.35$ \\
        \bottomrule
    \end{tabular}
    \caption{{\em Quantitative results for novel pose synthesis}.
         We report the PSNR [dB] for the reconstructed RGB images on the real-world H36M~\cite{H36M} and our~\dataabr test set, cf.~\cref{sec:dataset}.
         See~\cref{sec:experiments:neural_texture_rendering} for a discussion of the results.
    }
    \label{tab:quantitative-color}
\end{table}
%
Quantitative results for novel pose synthesis are shown in~\cref{tab:quantitative-color}.
We report the PSNR [dB] over all test samples for our reconstructed RGB images on the real-world H36M~\cite{H36M} and our~\dataabr test set.
%

We achieve an overall PSNR of $19.35dB$ on our novel~\dataabr data.
This is slightly better than the PSNR of $19.17dB$ reported in~\cite{NePu} although in contrast to~\cite{NePu}, we render twelve textures instead of one with a single model.
%

\begin{figure*}[t]
    \centerline{
        \includegraphics[width=0.24\linewidth,trim={3cm 4cm 3cm 5cm},clip]{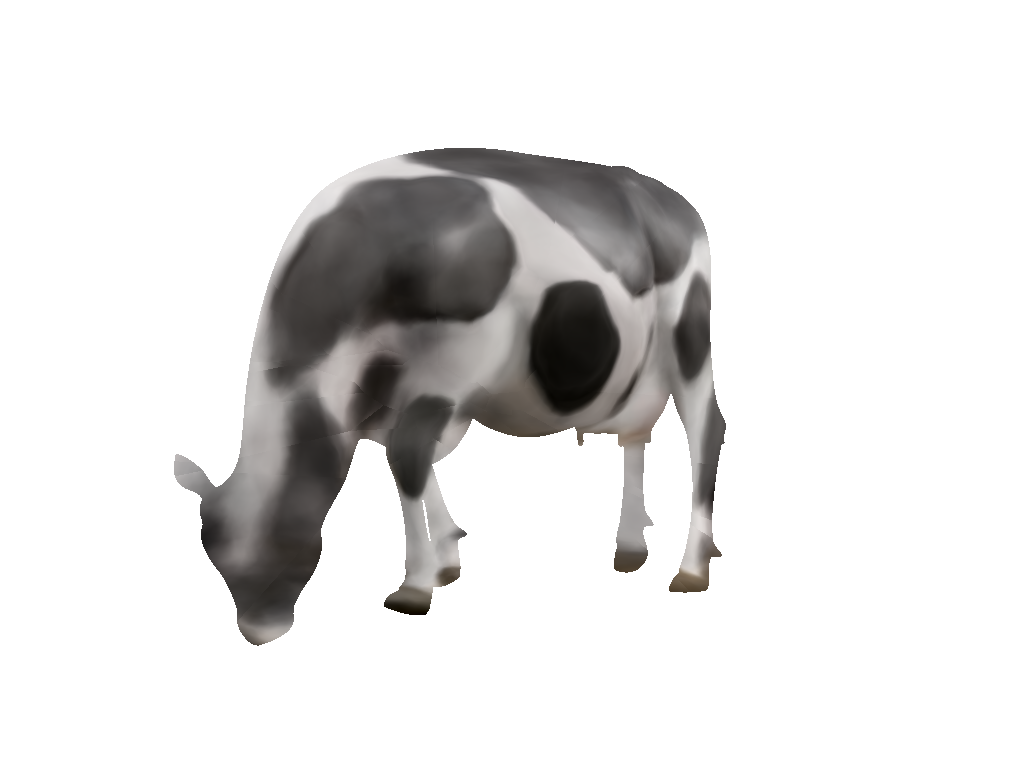}
        \includegraphics[width=0.24\linewidth,trim={3cm 4cm 3cm 5cm},clip]{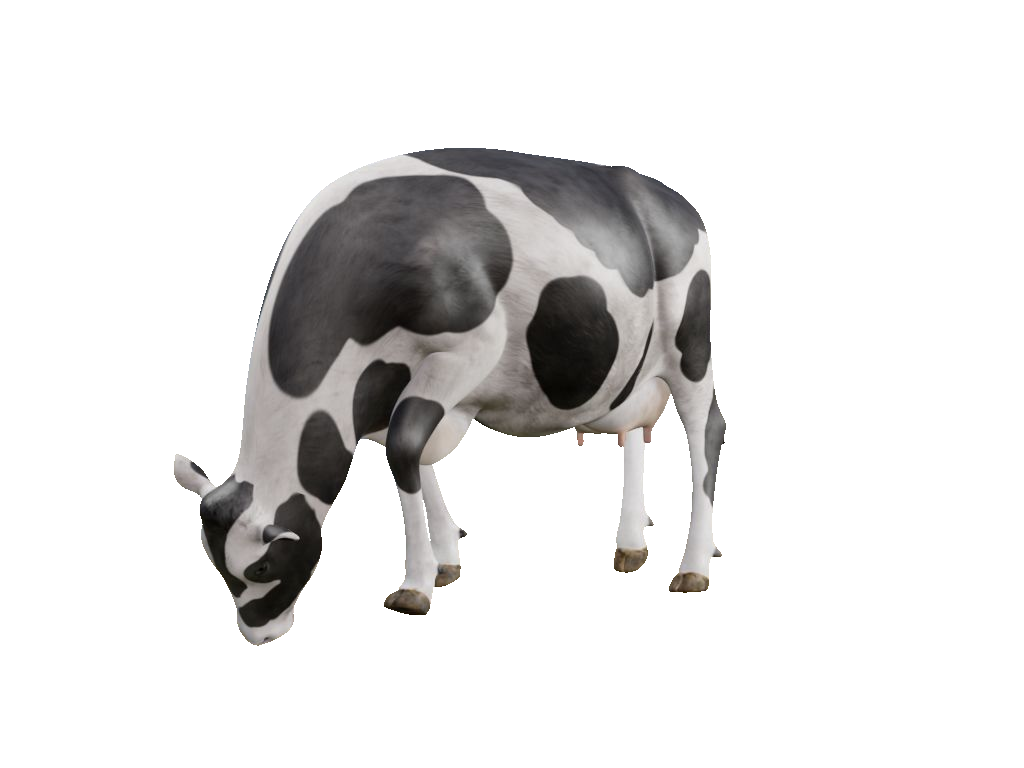}
        \includegraphics[width=0.24\linewidth,trim={6cm 11cm 6cm 9cm},clip]{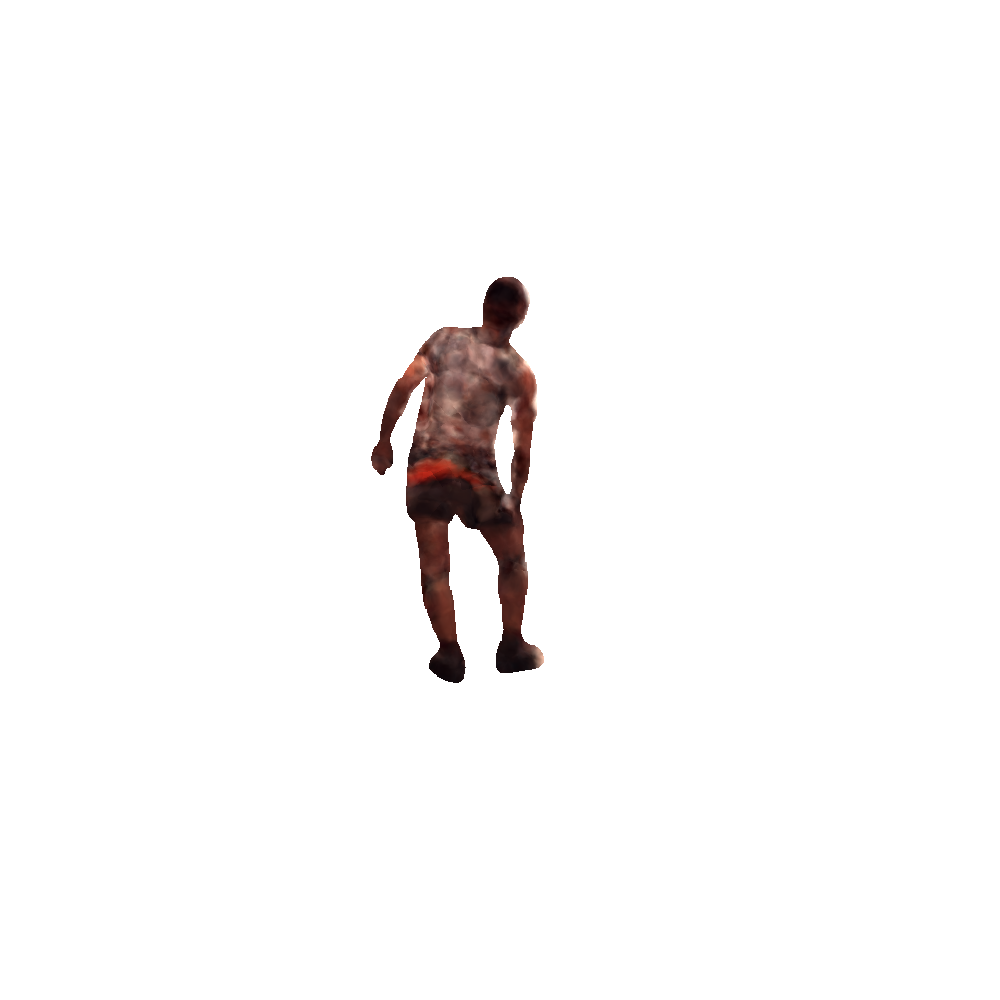}
        \includegraphics[width=0.24\linewidth,trim={6cm 11cm 6cm 9cm},clip]{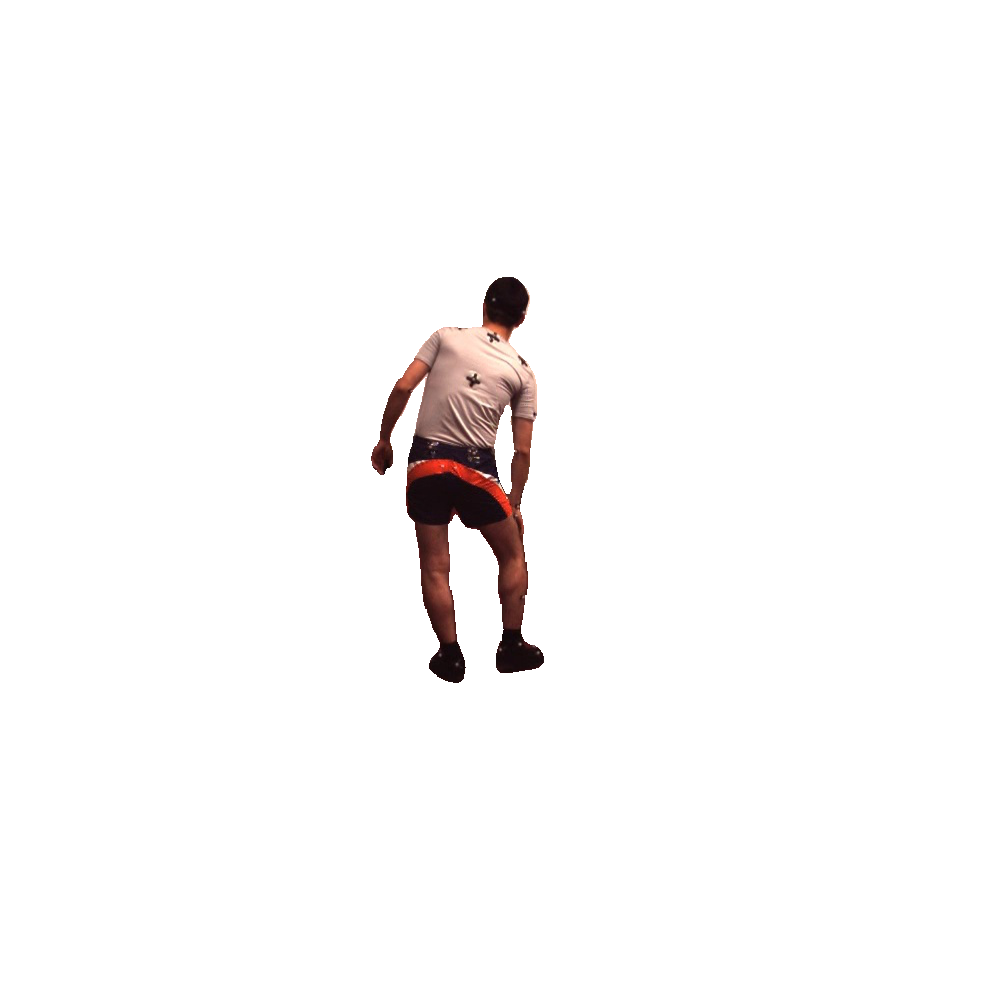}
    }
    \caption{
        \textbf{Novel Pose Synthesis Without NNOPCS Maps.}
        Color reconstruction and ground truth for a novel pose, where the pipeline does not use NNOPCS maps (cf.~\cref{sec:positional_encoding}).
        \textit{Left Pair}: Holstein cow (texture $0$) from our~\dataabr test set, cf.~\cref{sec:dataset}.
        \textit{Right Pair}: Subject $8$ from the H36M dataset~\cite{H36M}.
        See~\cref{sec:experiments:neural_texture_rendering,sec:limitations_future_work} for a discussion of the results.
    }
    \label{fig:ablation_study}
\end{figure*}

On the H36M data~\cite{H36M} we achieve an overall PSNR of $15.98dB$.
See~\cref{fig:ablation_study} (right pair) for a qualitative example.
Please note that the H36M data does not provide NNOPCS maps (cf.~\cref{sec:positional_encoding}) and we thus train~\paperabr without them.
In this case~\paperabr can not leverage the geometric shape information provided by the NNOPCS maps.
This holds for any data that does not provide NNOPCS maps.
For a deeper insight on the influence and the importance of the NNOPCS maps, we refer to our ablation studies.
%

We reconstruct images for novel views and novel poses from our~\dataabr test set, which have not been seen during training.
In~\cref{fig:novel-pose-novel-views} we show two examples for each distinct texture in our dataset.
The reconstructions look realistic and contain details like the black freckles of texture $3-6$ and $8$ (cf. eg.~\cref{fig:npnv:5})
or the black legs, ears and nose tip of texture $3$ (cf.~\cref{fig:npnv:3}).
The dataset contains two challenging texture pairs, which are difficult to distinguish by eye.
We note that even these challenging texture pairs $9$, $10$ and $1$, $11$ are reconstructed correctly (cf.~\cref{fig:npnv:9,fig:npnv:10} and~\cref{fig:npnv:1,fig:npnv:11} respectively).
%
%
\begin{figure*}[t]
    \centering
    \begin{subfigure}[b]{0.24\linewidth}
        \centering
        \includegraphics[width=\linewidth,trim={1cm 1.2cm 1cm 0.5cm},clip]{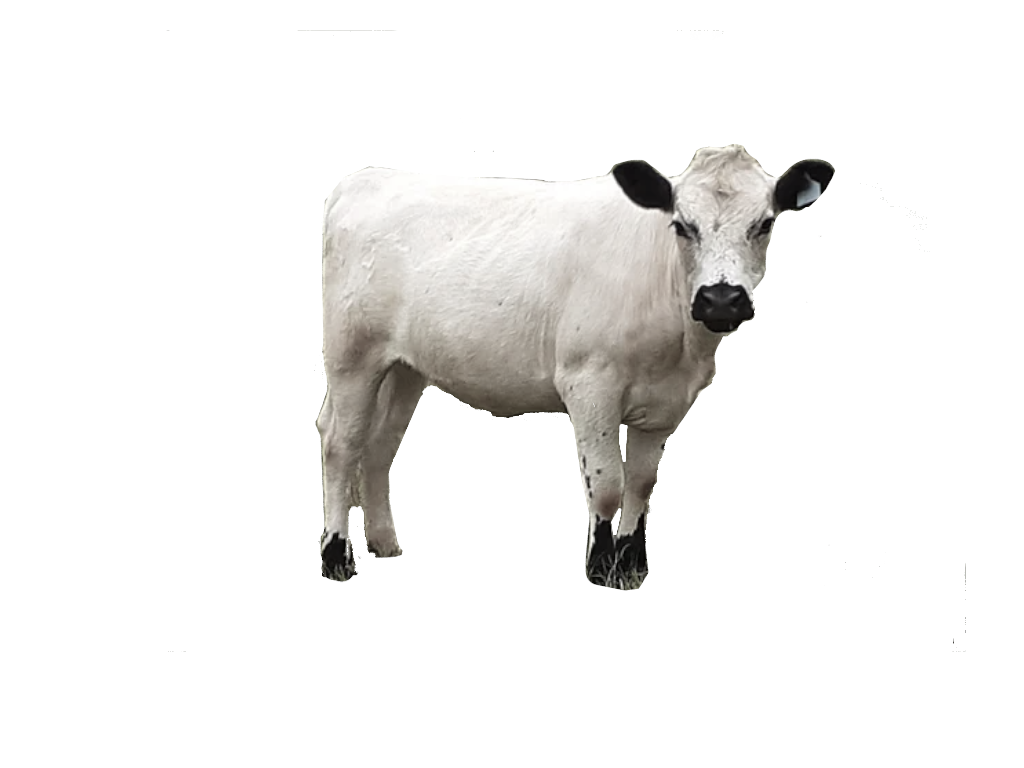}
        \caption{Masked Real-World Image}
        \label{fig:real_world:masked_image}
    \end{subfigure}
    \hfill
    \begin{subfigure}[b]{0.74\linewidth}
        \centering
        \includegraphics[width=0.32\linewidth,trim={5cm 5cm 1cm 3cm},clip]{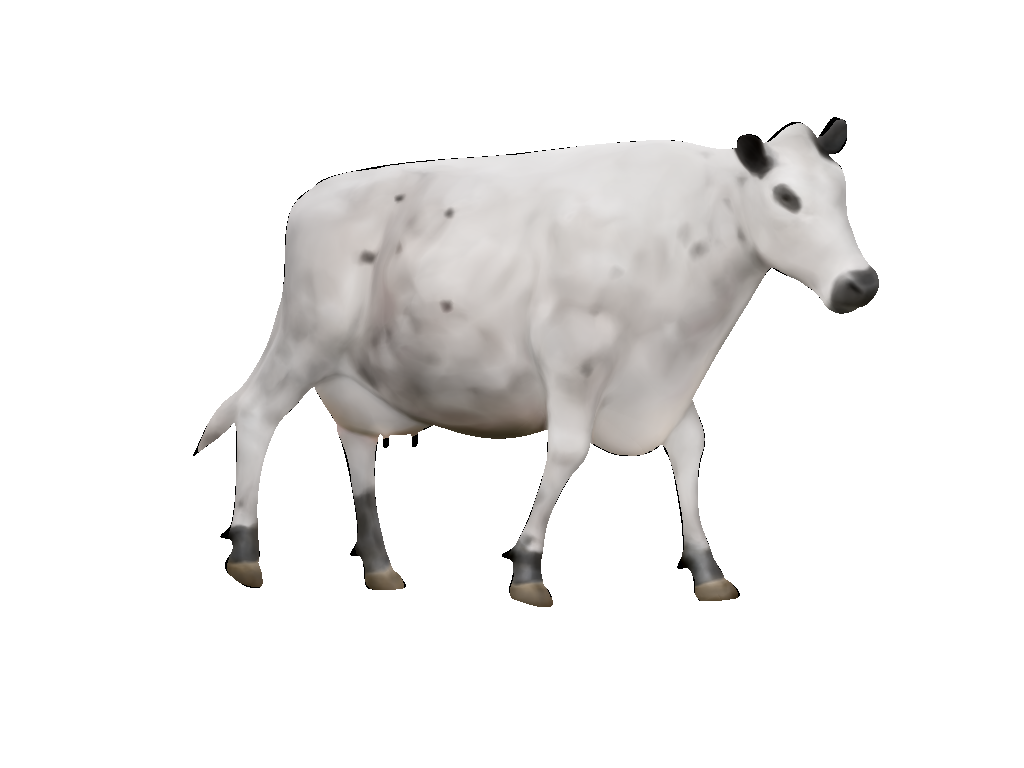}
        \hfill
        \includegraphics[width=0.32\linewidth,trim={5cm 5cm 1cm 3cm},clip]{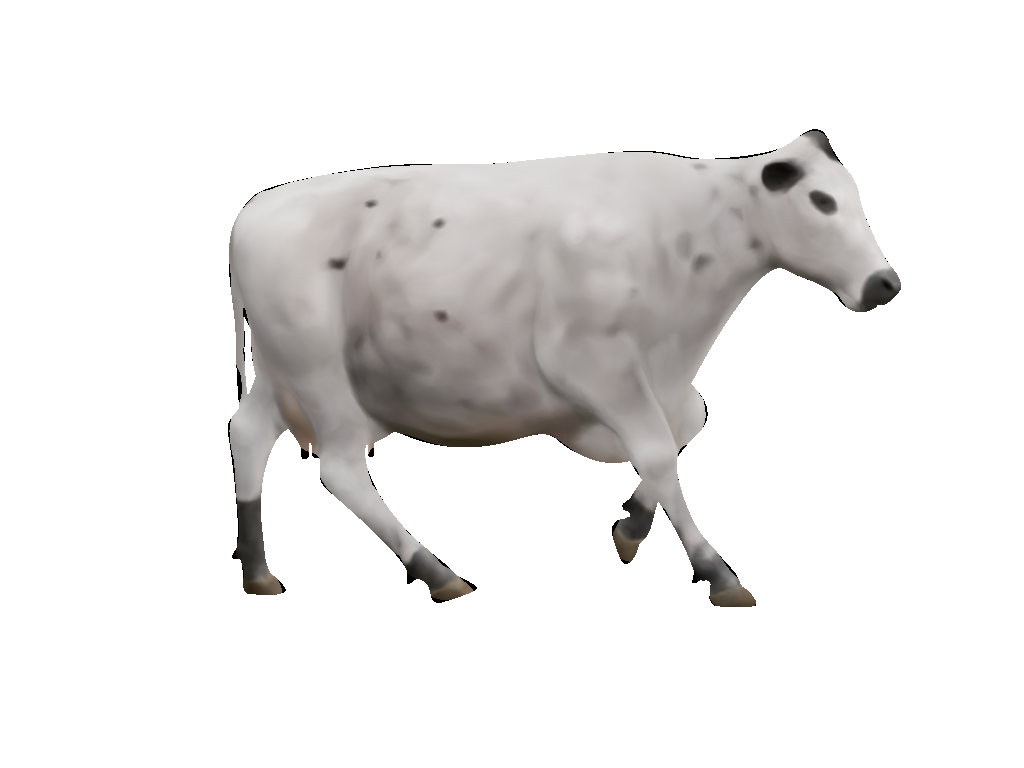}
        \hfill
        \includegraphics[width=0.32\linewidth,trim={5cm 5cm 1cm 3cm},clip]{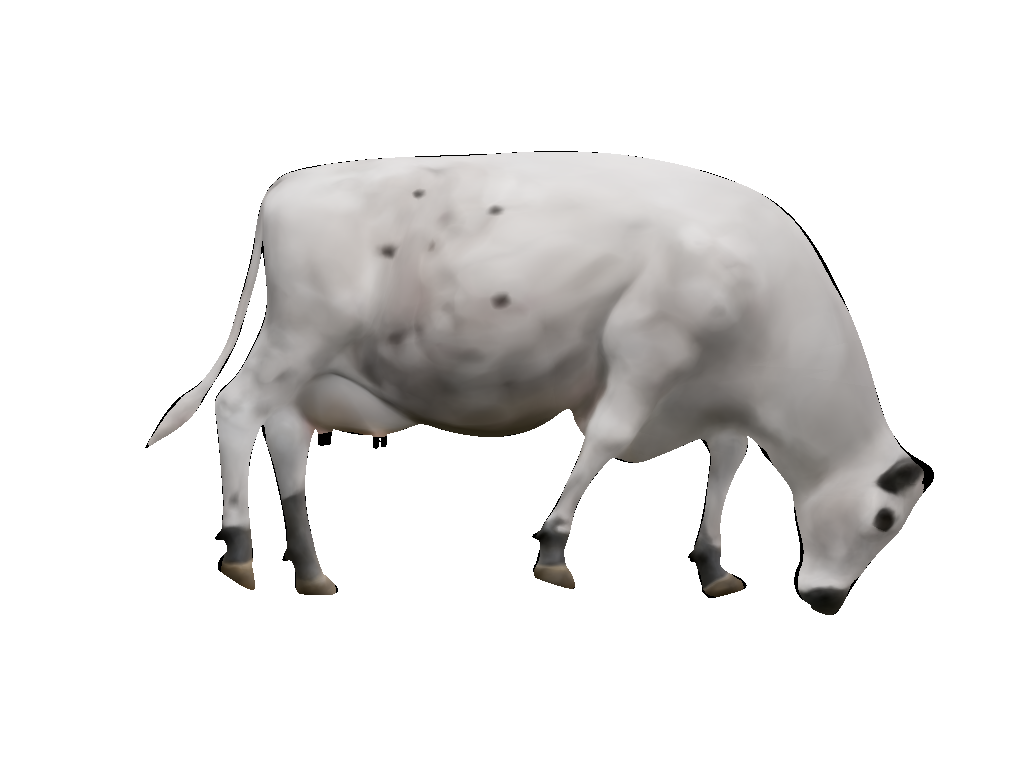}
        \caption{Our Reconstructions}
        \label{fig:real_world:rec}
    \end{subfigure}
    \caption{
        \textbf{Synthetic to Real-World Texture Domain Shift.}
        We show a zero-shot synthetic to real-world example.
        \paperabr reconstructs the texture from a real-world RGB image.
        We show the reconstructed texture in three different poses.
    }
    \label{fig:real_world}
\end{figure*}
We also perform a texture domain shift from synthetic to real.
For this, we use the model that we train on our novel~\dataabr texture dataset and infer a zero-shot synthetic to real-world example.
Our reconstructions are shown in~\cref{fig:real_world}.
We reconstruct the texture of the real-world image for three different poses (cf.~\cref{fig:real_world:rec}).
Even though only trained with synthetic data, this shows that \paperabr can be used for real-world applications that do not provide NNOPCS maps, cf.~\cref{sec:positional_encoding}.
This makes our model useful for real-world applications with endangered animal species where available real-world data is limited.

\subheading{Ablation Studies}
\begin{table}[t]
    \centering
    \small
    \begin{tabular}{c|c}
        \toprule
         & Color PSNR [dB] \\
        \midrule
        w/o NNOPCS Maps & $13.63$ \\
        w/ NNOPCS Maps & $\mathbf{19.35}$ \\
        \bottomrule
    \end{tabular}
    \caption{{\em Ablation study on the influence of the NNOPCS maps}.
         We report the PSNR [dB] for the reconstructed RGB images on our~\dataabr test set, cf.~\cref{sec:dataset}.
         We report results for an architecture that leverages the NNOPCS maps (cf.~\cref{sec:positional_encoding}) and one that does not.
         Best result is bold.
    }
    \label{tab:ablation-study-nnopcs-maps}
\end{table}
To evaluate the influence of the NNOPCS maps (cf.~\cref{sec:positional_encoding}) we run an experiment on our novel~\dataabr dataset with the same parameters as specified in~\cref{sec:experiments:implementation_details}.
The only difference is that we do not use NNOPCS maps in our pipeline, cf.~\cref{fig:architecture_overview}.
We report the color PSNR over all test samples in~\cref{tab:ablation-study-nnopcs-maps}.
If our model can leverage geometric information of the shape that is provided by the NNOPCS maps, \paperabr achieves a better color PSNR on average by $5.72dB$.
This quantitative difference is visible in the qualitative results, cf.~\cref{fig:ablation_study} (left pair).
This also explains the quality of our results on the H36M dataset~\cite{H36M}, cf.~\cref{tab:quantitative-color,fig:ablation_study} (right pair).

%
We perform another ablation study on the influence of the local texture features $\mathbf{f}_{\text{texture}} \in \mathbb{R}^{d_f}$.
We run an experiment on our novel~\dataabr data where we do not augment the local features for geometry $\mathbf{f}$ with local features for texture $\mathbf{f}_{\text{texture}}$.
We see that augmenting the local features for geometry $\mathbf{f}$ with $\mathbf{f}_{\text{texture}}$ achieves a better color PSNR on average by $4.61$ dB ($19.35$ dB in~\cref{tab:quantitative-color} vs. $14.74$ dB).

\subsection{Re-identification}
\label{sec:experiments:reID}
%
Our proposed neural rendering pipeline learns a global texture embedding which can be used to identify individuals.
This is a valuable task for applications that need re-identification of individuals like tracking scenarios where the tracked object leaves and re-enters the scene.
For this task, we use~\eqref{eq:texture-encoder} to encode a global latent
vector~$\mathbf{z}_{texture}^{m,c,t}$ that describes the texture of individual~$t$ observed from camera
view~$c$ in pose~$m$.
In this way, we can generate a global texture embedding $\mathbf{Z}_{texture}$ (capital Z) for all individuals under consideration from our
learned NNOPCS maps (cf.~\cref{sec:positional_encoding})~$\mathbf{\hat{P}}_{m,c}$ and masked 2D color observations~$\mathbf{\tilde{I}}_{m,c,t}$,
\begin{equation}
\begin{gathered}
    \mathbf{Z}_{texture} = \{ \mathbf{z}_{texture}^{m,c,t} \}, \\
    m\in\{1,\dots,M\},\; c\in\{1,\dots,C\},\; t\in\{1,\dots,T\}.
    \label{eq:global-texture-embedding-for-reid}
\end{gathered}
\end{equation}

\subheading{Baseline}
For a comparison, we train TransReID~\cite{he2021transreid} on our novel~\dataabr data, cf.~\cref{sec:dataset}.
\cite{he2021transreid} is a transformer-based framework for object re-identification that is optimized with ID~\cite{zheng2017discriminatively} and triplet loss~\cite{liu2017end}.
During training~\cite{he2021transreid} needs positive and negative pairs of the object.
We resize the input to $128 \times 256$ px and use a batch size of $40$.
For the other parameters we use their standard.
After training, we generate a global feature embedding $\mathbf{Z}$ with their global features for a comparison.
%

%
\begin{figure}[t]
    \centering
        \begin{subfigure}[b]{\linewidth}
            \centering
            \includegraphics[width=\linewidth]{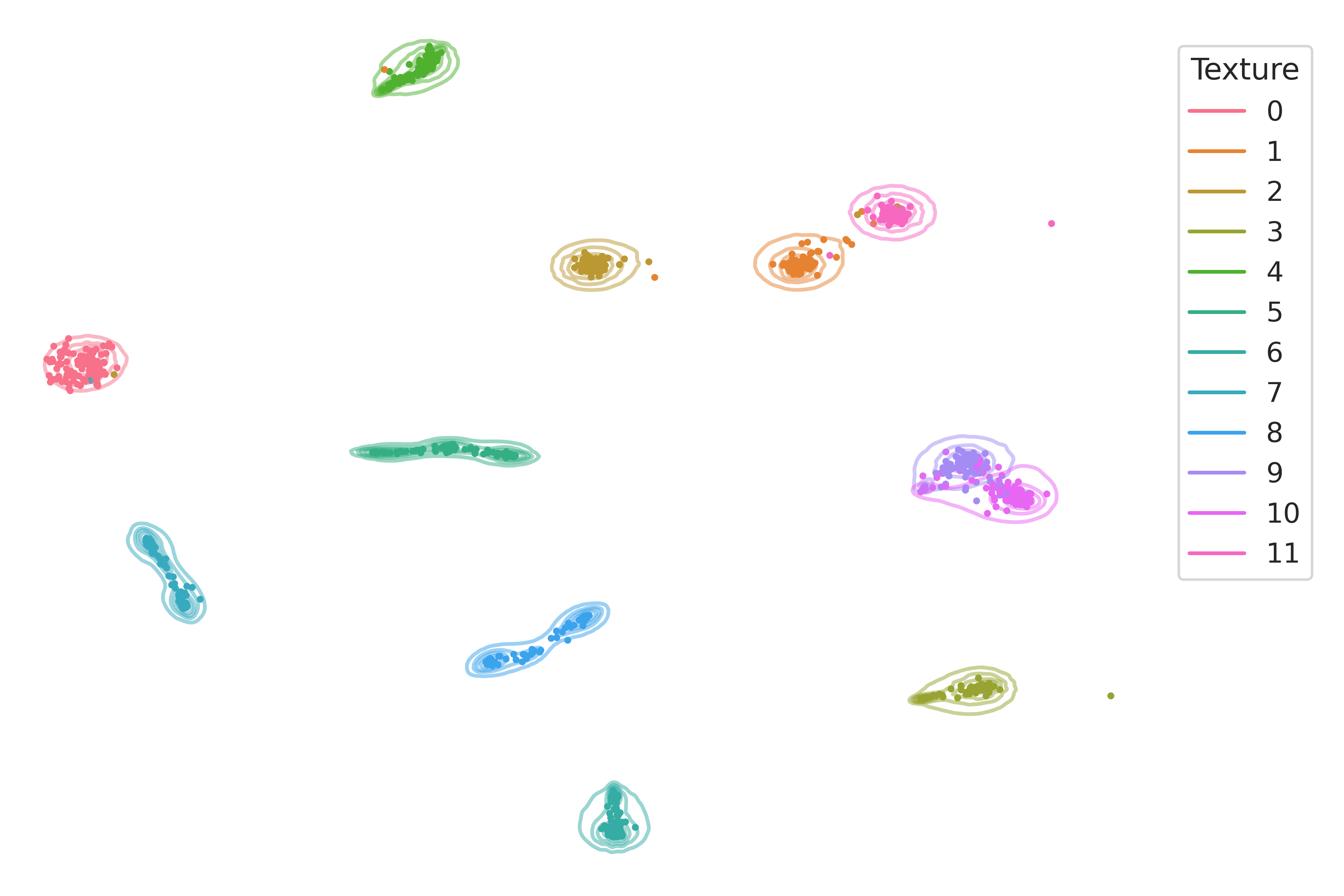}
            \caption{\paperabr:
            In addition to the distribution of latent codes for known camera views from the training set (contour lines), the dots show novel views and fall into the clusters of the known ones.
            }
            \label{fig:reid:netepu}
        \end{subfigure}
        \begin{subfigure}[b]{\linewidth}
            \centering
            \includegraphics[width=\linewidth]{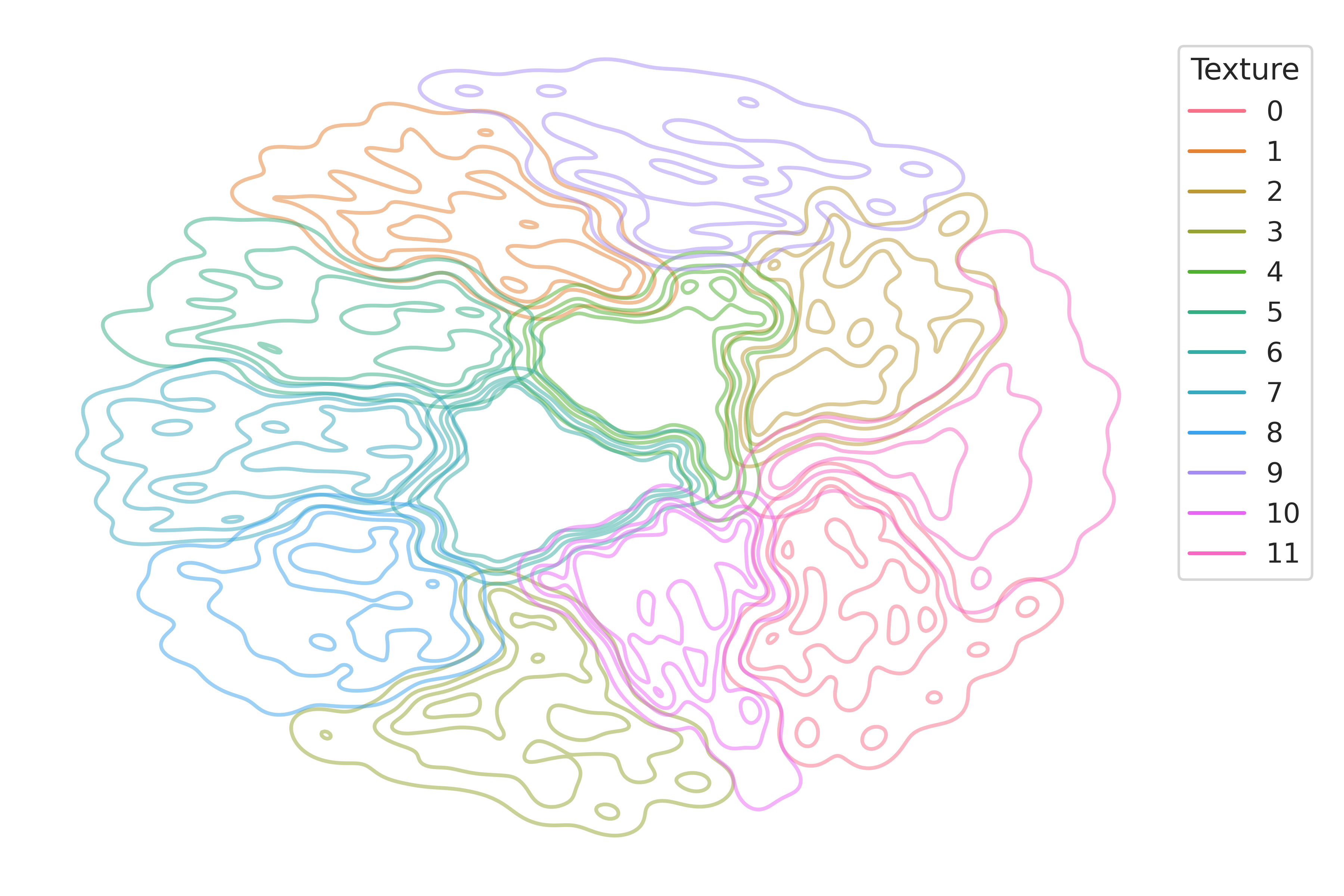}
            \caption{TransReID~\cite{he2021transreid}:
            While the embedding is more compact, the individual clusters overlap more often.
            }
            \label{fig:reid:transreid}
        \end{subfigure}
    \caption{
        \textbf{KDE of the Global Texture Embedding.
        }
        We show the KDE~\cite{rosenblatt1956remarks,parzen1962on} of the t-SNE of the global texture codes.
        The contour lines show the distribution of latent codes for camera views from our novel~\dataabr dataset.
        %
        The individual cows cluster nicely.
        See~\cref{sec:experiments:reID} for a discussion of the results.
    }
    \label{fig:reid}
\end{figure}

\subheading{Results}
%
In \cref{fig:reid:netepu} we show~\paperabr's kernel density estimation (KDE)~\cite{rosenblatt1956remarks,parzen1962on} of the t-SNE~\cite{JMLR:v9:vandermaaten08a} of the global texture embedding from~\eqref{eq:global-texture-embedding-for-reid} of twelve cows from our novel synthetic~\dataabr dataset (cf.~\cref{sec:dataset}).
We encode the global texture codes of all $910$ poses, twelve textures and $24$ cameras.
While the contour lines show the distribution of global texture codes for the camera views from our dataset, the dots show the global codes for novel camera views.
In~\cref{fig:reid:transreid} we show the KDE of the global feature embedding $\mathbf{Z}$ generated with~\cite{he2021transreid} for a comparison.

We see that the textures cluster in such a way that they can be distinguished by both frameworks.
Both frameworks can distinguish the challenging textures~$1$ and~$11$ (cf.~\cref{fig:npnv:1,fig:npnv:11}) and~$9$ and~$10$ (cf.~\cref{fig:npnv:9,fig:npnv:10}).
While the embedding of~\cite{he2021transreid} is more compact, we observe only one slight overlap of texture clusters for~\paperabr.
No cluster overlap in the embedding helps in readily distinguishing all textures in all poses from all camera views.
%
Furthermore, note that~\paperabr's global texture latent codes of novel camera views (dots in~\cref{fig:reid:netepu}) fall into the corresponding clusters of the known camera views
(contour lines in~\cref{fig:reid:netepu}).
%

%
%

\subheading{Runtime}
We evaluate the runtime on our novel~\dataabr dataset with a workstation that has a nVidia Titan RTX, 64 GB DDR4 RAM, an Intel Xeon E5-2620 at 2.10GHz and a 2TB Samsung SSD 850.
We encode the global texture code from an input of size $768\times 1024\times 6$ (masked 2D color observation and learned NNOPCS map, cf.~\cref{sec:positional_encoding}).
This calculation includes to render our learned NNOPCS maps with a resolution of $768\times 1024\times 3$.
Since our NNOPCS maps are only defined within the object, the calculation also includes to render the learned mask with a resolution of $768\times 1024$.
We use a batch size of $1$ for all twelve textures of $100$ samples from our dataset seen from $24$ camera views.
In total, we thus encode the global texture code for $28800$ frames.
To calculate the runtime, we take the average over these frames.
We achieve a runtime of $0.5$ fps.
%

All together, this is why our proposed neural rendering pipeline offers an alternative approach to CNN- or transformer-based frameworks like~\cite{he2021transreid} for the task of re-identification in interactive tracking applications.

\section{Limitations and Future Work}
\label{sec:limitations_future_work}
At the moment, we learn the NNOPCS maps in a supervised manner, cf.~\cref{sec:positional_encoding}.
This limits our method to data that provide NNOPCS maps.
If we train on data that does not provide these NNOCPS maps, the quality of the outcome diminishs because the model lacks information on the geometry's shape, cf.~\cref{fig:ablation_study}.
In order to make our method more applicable, especially to real-world applications,
we aim to learn the NNOPCS maps in the future in an unsupervised manner.
%
Also, at this point our dataset contains a limited variety of textures. 
In the future we aim at extending our dataset to allow for better generalizability when it comes to new - unseen - textures.

\section{Conclusions}
In this paper, we present a neural rendering pipeline for textured articulated shapes.
%
%
We show that~\paperabr encodes a distinct global texture embedding (cf.~\cref{fig:reid:netepu}), which is computed from a learned NNOPCS map (cf.~\cref{sec:positional_encoding})
and 2D color information.
This global texture embedding can be used in a downstream task to identify individuals.
Our neural rendering-based re-identification process runs at interactive speed (cf.~\cref{sec:experiments:reID}),
and thus offers an alternative to CNN- or transformer-based approaches like~\cite{he2021transreid} in tracking applications.
To the best of our knowledge, we are the first to provide a framework for re-identification of articulated individuals based on neural rendering.
We also demonstrate~\paperabr's quality with realistic looking novel view and pose synthesis for different synthetic cow textures, cf.~\cref{fig:novel-pose-novel-views}.
Restricted by the availability of ground truth NNOPCS maps, the quality for real-world data synthesis is reduced, cf.~\cref{fig:ablation_study} (right pair).
We further demonstrate the flexibility of our model by applying a synthetic to real-world texture domain shift where we reconstruct the texture from a real-world 2D RGB image with a model trained on synthetic data only.
This makes our model useful for real-world applications with endangered animal species where available real-world data is limited and synthetic data can be generated using Blender (\url{www.blender.org}).

We hope that this work inspires other researchers to develop methods for neural rendering-based re-identification which work for humans and animals.
%

{\small
\subheading{Acknowledgements}
We acknowledge funding by the Deutsche Forschungsgemeinschaft (DFG, German Research Foundation) under Germany's Excellence Strategy -- EXC 2117 -- 422037984, and the Federal Ministry of Education and Research (BMBF) within the research program -- KI4KMU -- 01IS23046B.
}
%

\newpage

{\small
\bibliographystyle{ieee_fullname}
\bibliography{refs}
}

\clearpage
\appendix
\section{Architecture}
\begin{figure}[ht]
    \centerline{
        \includegraphics[width=\linewidth,trim={8cm 5cm 9cm 4cm},clip]{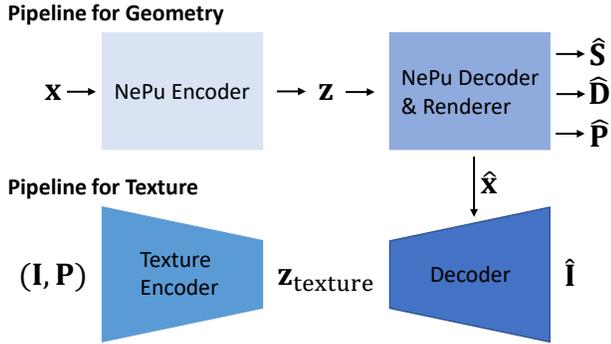}
    }
    \caption{
        \textbf{Complete Pipeline.}
        Our key idea in this work is to disentangle the pipelines for geometry (top) and texture information (bottom) of the shape.
        We modify the original NePu~\cite{NePu} pipeline to predict a NNOPCS map (cf. Sec. 3.1 in main paper) $\mathbf{\hat{P}}$ (top).
        The heart of our framework is an encoder-decoder network for the texture of an articulated shape (bottom).
        The detailed architecture of our texture encoder (bottom left) can be found in~\cref{fig:texture_encoder_architecture}.
        \cref{fig:supp:decoder} shows the decoder (bottom right) in more detail.
    }
    \label{fig:supp:architecture_overview}
\end{figure}
\begin{figure}[t]
    \centerline{
        \includegraphics[width=\linewidth]{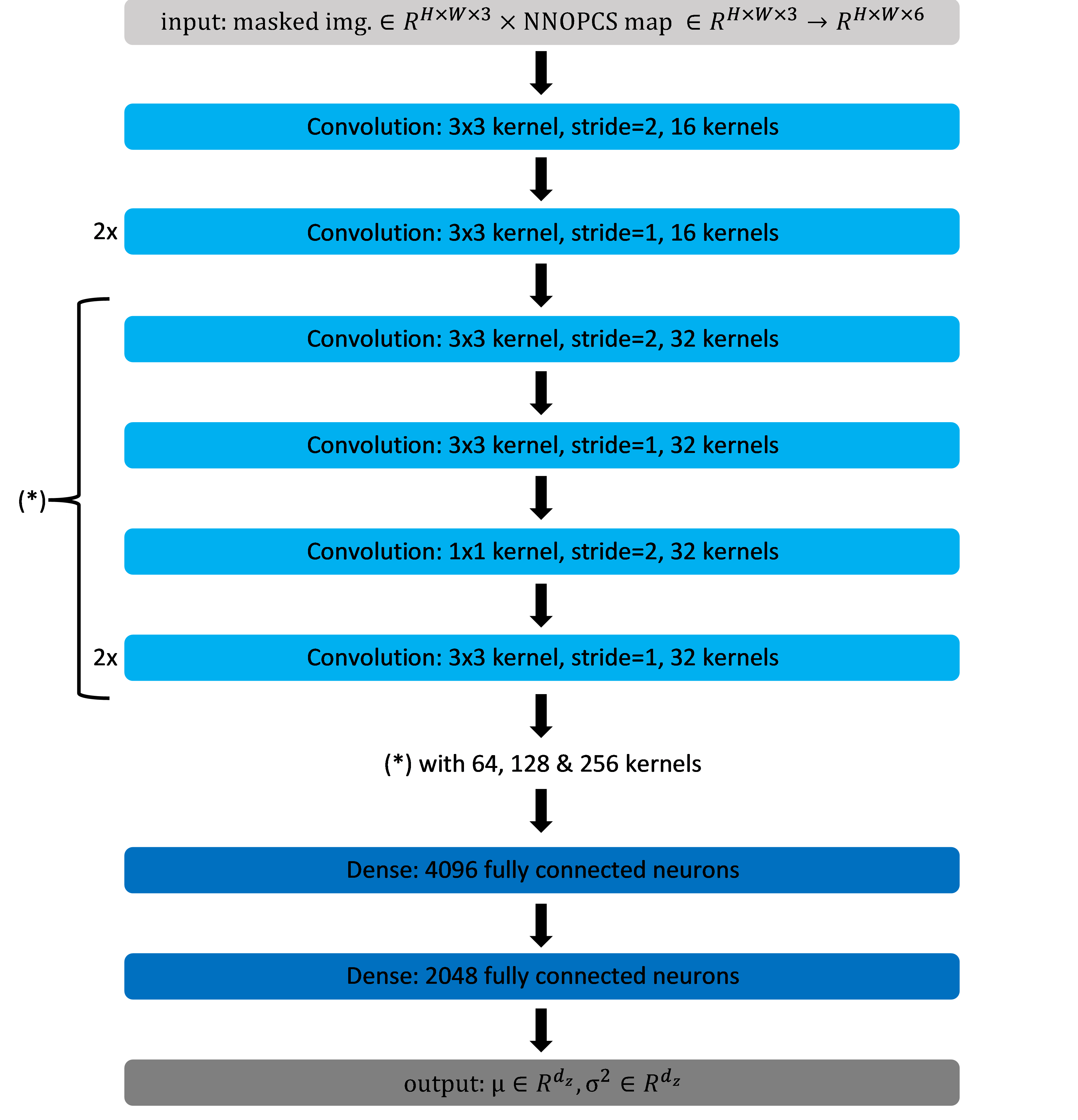}
    }
    \caption{
        \textbf{Architecture of Our Texture Encoder.}
        The texture encoder
        takes a masked RGB image and the NNOPCS map (cf. Sec. 3.1 in main paper) as input and outputs a mean $\mu \in \mathbb{R}^{d_z}$ and standard deviation $\sigma^2\in \mathbb{R}^{d_z}$ using $23$ convolutional and two dense layers.
    }
    \label{fig:texture_encoder_architecture}
\end{figure}
\begin{figure}[t]
    \centerline{
        \includegraphics[width=\linewidth,trim={6cm 6cm 8cm 6cm},clip]{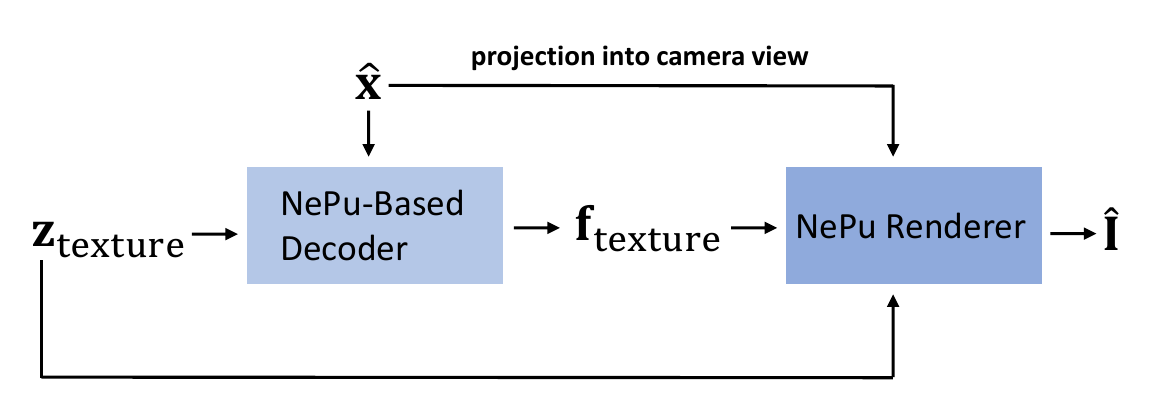}
    }
    \caption{
        \textbf{Decoder.}
        We show the decoder from~\cref{fig:supp:architecture_overview}, bottom right, in more detail.
        For further details, we refer to~\cite{NePu}.
    }
    \label{fig:supp:decoder}
\end{figure}
As mentioned in the main paper, our key idea in this work is to disentangle the pipelines for geometry and texture information of the shape.
The original NePu~\cite{NePu} pipeline is modified (cf.~\cref{fig:supp:architecture_overview}, top) to produce a NNOPCS map~$\mathbf{\hat{P}}$ (cf. Sec. 3.1 in main paper, similar to~\cite{NOCS}) instead of a final image, which essentially encodes within the image plane the complete geometric information about the shape relevant for rendering this particular view.
In addition, we also predict the silhouette $\mathbf{\hat{S}}$ and depth map $\mathbf{\hat{D}}$ like the original NePu pipeline.
The heart of our framework is an encoder-decoder network for the texture of an articulated shape (cf.~\cref{fig:supp:architecture_overview}, bottom).
The detailed architecture of our texture encoder (cf.~\cref{fig:supp:architecture_overview}, bottom left) can be found in~\cref{fig:texture_encoder_architecture}.
\cref{fig:supp:decoder} shows the decoder (cf.~\cref{fig:supp:architecture_overview}, bottom right) in more detail.

\subsection{Additional Implementation Details}
As mentioned in the main paper, in order to disentangle geometry and texture (cf.~\cref{fig:supp:architecture_overview}), we augment the local and global features for geometry $\mathbf{f} \in \mathbb{R}^{d_f}$ and $\mathbf{z} \in \mathbb{R}^{d_z}$, respectively, with local and global features $\mathbf{f}_{\text{texture}} \in \mathbb{R}^{d_f}$ and $\mathbf{z}_{\text{texture}}  \in \mathbb{R}^{d_z}$, respectively, for the texture.
In all experiments, we choose $d_f = 256$ and $d_z = 1024$.
For the other implementation details, we refer to~\cite{NePu}.

\subsection{Additional Ablation Study}
\begin{table}[ht]
    \centering
    \small
    \begin{tabular}{c|c}
        \toprule
         & Color PSNR [dB] \\
        \midrule
        w/o $\mathbf{f}_{\text{texture}}$ & $14.74$ \\
        \rowcolor{verylightgray} w/ $\mathbf{f}_{\text{texture}}$ & $\mathbf{19.35}$ \\
        \bottomrule
    \end{tabular}
    \caption{{\em Ablation study on the influence of the local texture features}.
         We report the PSNR [dB] for the reconstructed RGB images on our~\dataabr test set.
         We report results for an architecture that uses local texture features $\mathbf{f}_{\text{texture}}$ and one that does not.
         Best result is bold.
         Configuration of this paper highlighted in gray.
    }
    \label{tab:supp:ablation-study}
\end{table}
We perform an ablation study on the influence of the local texture features $\mathbf{f}_{\text{texture}} \in \mathbb{R}^{d_f}$.
We run an experiment on our novel~\dataabr data where we do not augment the local features for geometry $\mathbf{f}$ with local features for texture $\mathbf{f}_{\text{texture}}$.
In~\cref{tab:supp:ablation-study} we report the color PSNR [dB] over all~\dataabr test samples.
We see that augmenting the local features for geometry $\mathbf{f}$ with $\mathbf{f}_{\text{texture}}$ achieves a better color PSNR on average by $4.61$ dB (cf.~\cref{tab:supp:ablation-study}).

\begin{figure*}[t]
    \centering
    \begin{subfigure}[b]{0.495\linewidth}
        \centering
        \includegraphics[width=\linewidth]{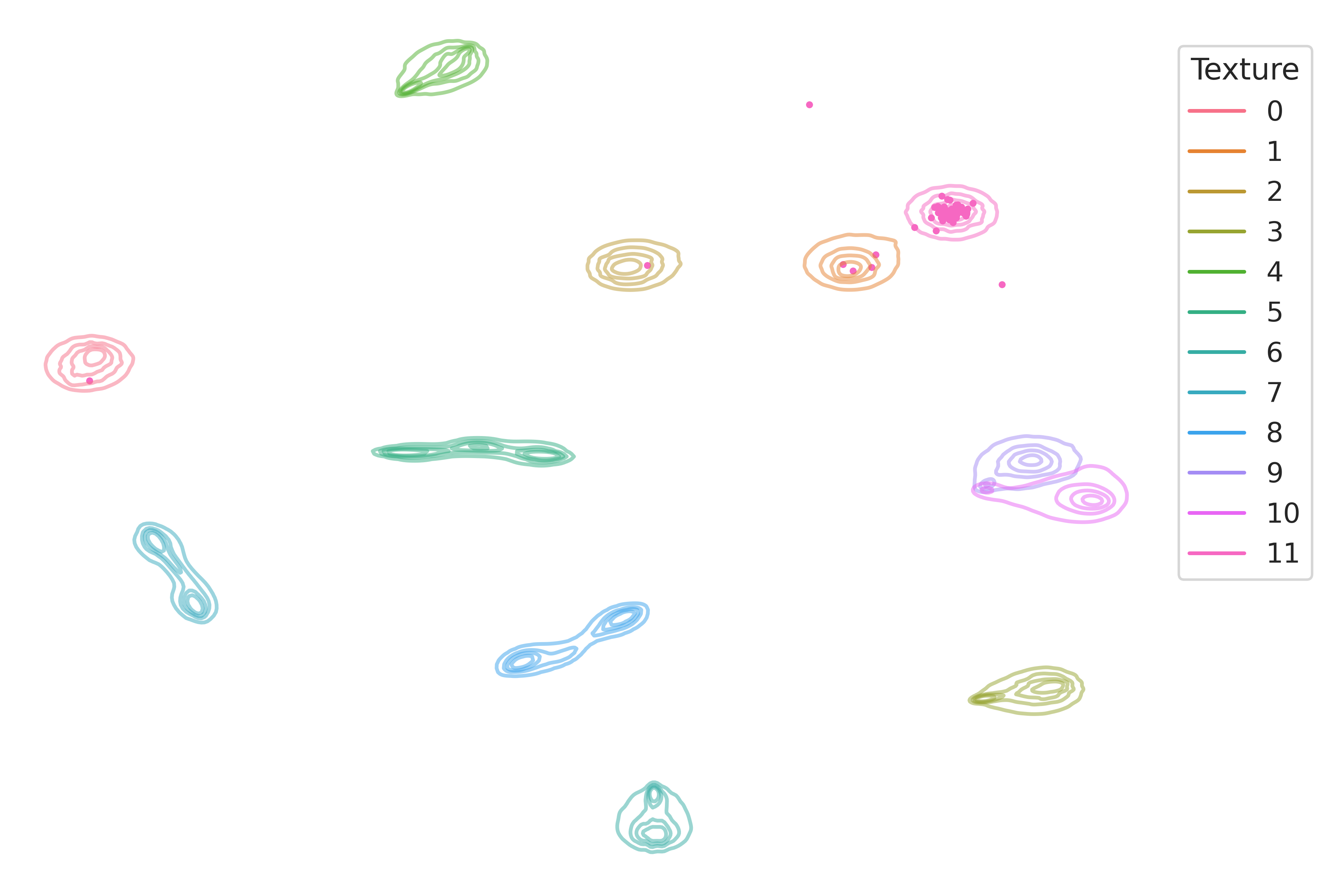}
        \caption{Complete Masks}
        \label{fig:reid:occ:complete_masks}
    \end{subfigure}
    \hfill
    \begin{subfigure}[b]{0.495\linewidth}
        \centering
        \includegraphics[width=\linewidth]{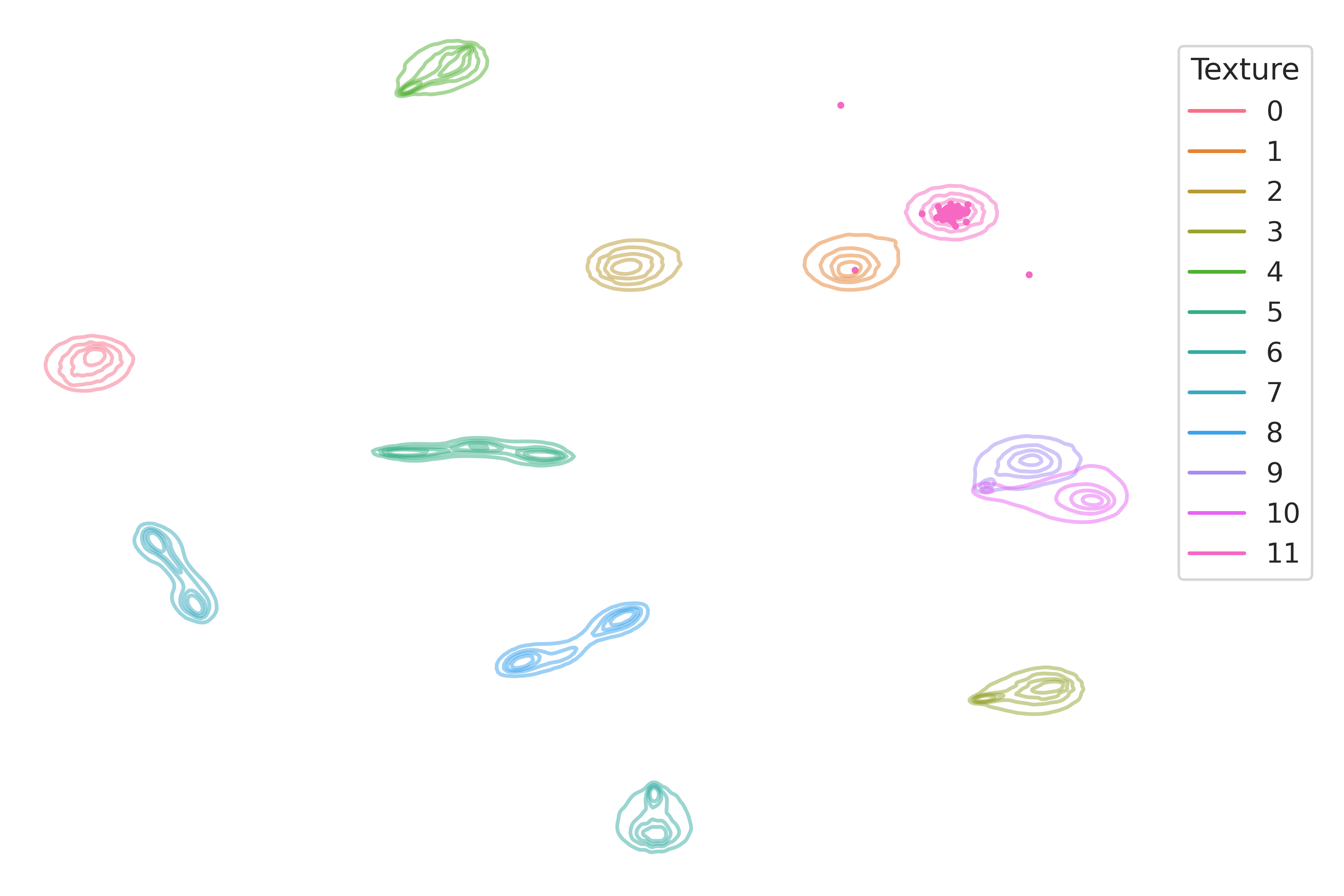}
        \caption{Occluded Masks}
        \label{fig:reid:occ:occ_masks}
    \end{subfigure}
    \caption{
        \textbf{KDE of the Global Texture Codes: Occlusions.}
        The contour lines depict the KDE of the t-SNE of latent codes of our training examples.
        The dots depict the position of the latent code for partially occluded examples.
        %
        The instances where the Holstein (texture $0$) partially occludes the Limousine (texture $11$) cow (cf.~\cref{fig:dataset_occ:image}) cluster correctly into the cluster of the Limousine cow.
        See~\cref{sec:supp:reid} for a discussion of the results.
    }
    \label{fig:reid:occ}
\end{figure*}

\section{\dataabr Dataset}
\label{sec:supp:dataset}
In~\cref{fig:supp:dataset} we show example frames from our novel texture dataset.
This dataset extends~\cite{NePu_dataset} with its Holstein cow by eleven cow textures.
The texture pairs $9$ and $10$ (cf.~\cref{fig:supp:dataset:9,fig:supp:dataset:10} respectively) and $1$ and $11$ (cf.~\cref{fig:supp:dataset:1,fig:supp:dataset:11} respectively) look similar and are thus challenging to distinguish. 

\subheading{Occlusions}
\begin{figure}[t]
    \centering
    \begin{subfigure}[b]{0.49\linewidth}
        \centering
        \includegraphics[width=\textwidth]{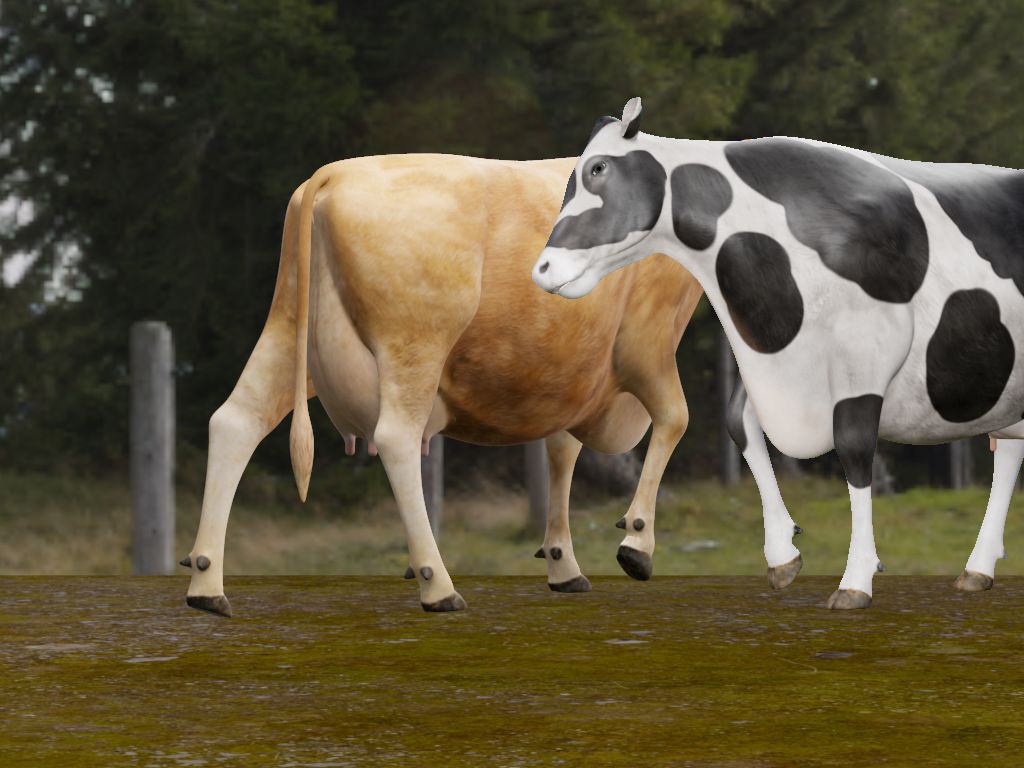}
        \caption{RGB Image}
        \label{fig:dataset_occ:image}
    \end{subfigure}
    \hfill
    \begin{subfigure}[b]{0.49\linewidth}
        \centering
        \includegraphics[width=\textwidth]{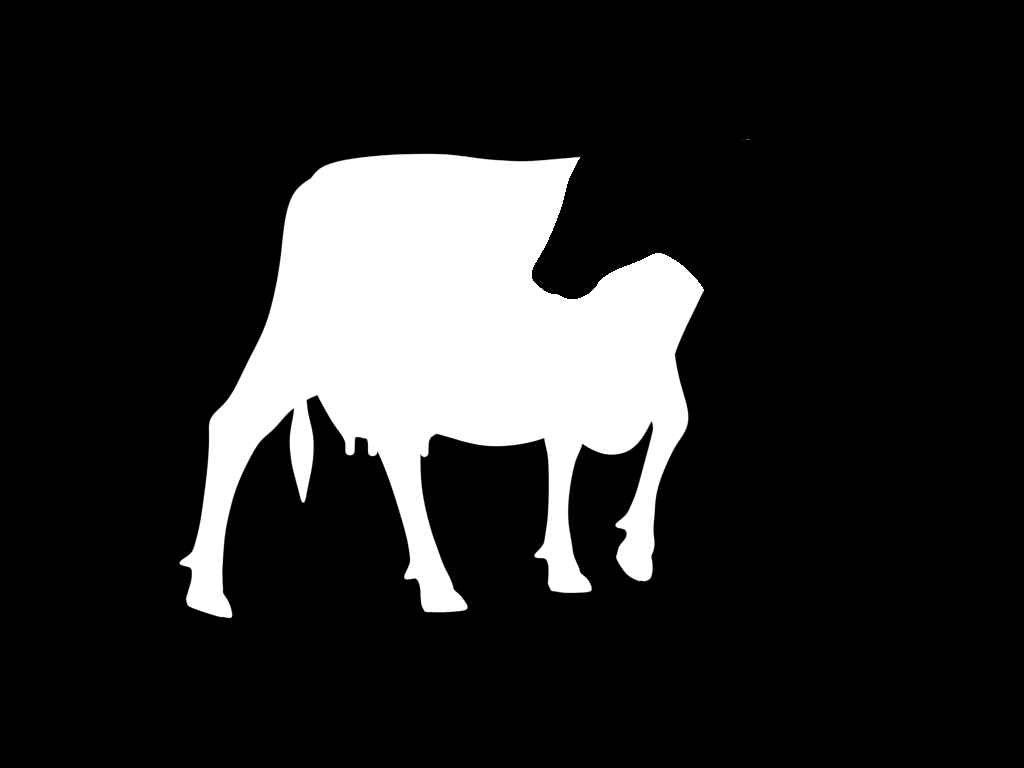}
        \caption{Occluded Mask}
        \label{fig:dataset_occ:occ_mask}
    \end{subfigure}
    \caption{
        \textbf{Example of Occluded Instances.}
        The Holstein (texture $0$) partially occludes the Limousine (texture $11$) cow.
        In total, we provide $50$ occluded instances in our novel texture dataset captured from two camera views.
    }
    \label{fig:dataset_occlusions}
\end{figure}
Our novel synthetic texture dataset also contains instances of occlusions.
In these samples the Holstein (texture $0$) partially occludes the Limousine (texture $11$) cow (short text description of cow textures in~\cref{tab:supp:quantitative-color}).
For these cases we provide the occluded (cf.~\cref{fig:dataset_occ:occ_mask}) and complete silhouettes.
We show one example in~\cref{fig:dataset_occlusions}.

\section{Additional Results}
%

\subsection{Neural Texture Rendering}
\begin{figure*}[ht!]
\vspace{3mm}
    \centering
    \begin{subfigure}[b]{0.24\linewidth}
        \centering
        \includegraphics[width=\textwidth]{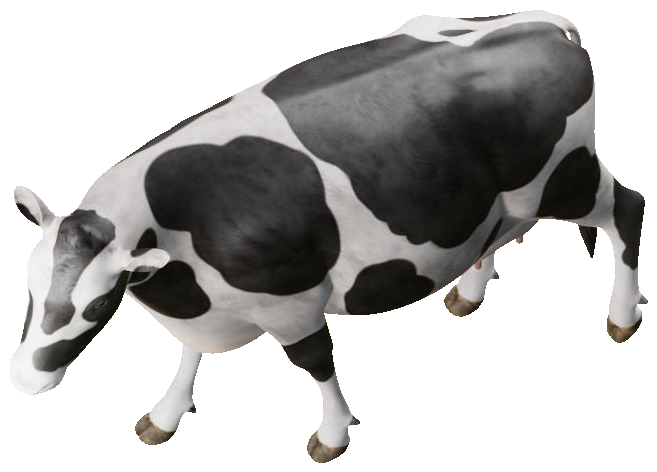}
        \caption{GT: Texture 0}
        \label{fig:color:gt0}
    \end{subfigure}
    \hfill
    \begin{subfigure}[b]{0.24\linewidth}
        \centering
        \includegraphics[width=\textwidth]{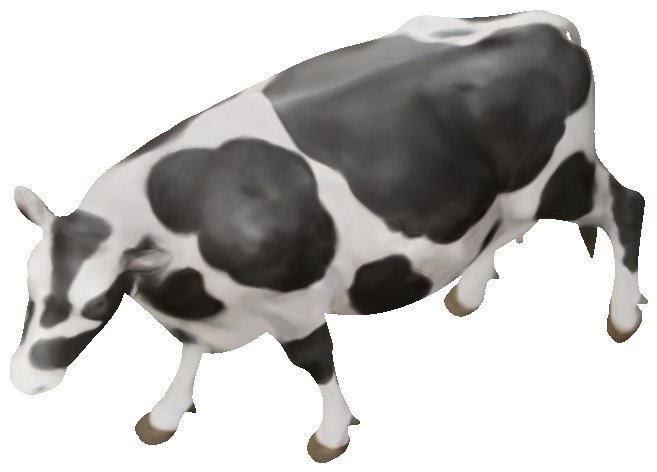}
        \caption{Rec.: Texture 0}
        \label{fig:color:rec0}
    \end{subfigure}
    \hfill
    \begin{subfigure}[b]{0.24\linewidth}
        \centering
        \includegraphics[width=\textwidth]{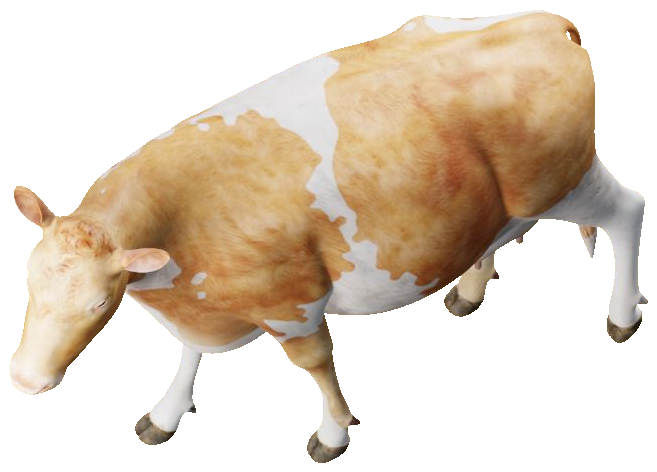}
        \caption{GT: Texture 2}
        \label{fig:color:gt2}
    \end{subfigure}
    \hfill
    \begin{subfigure}[b]{0.24\linewidth}
        \centering
        \includegraphics[width=\textwidth]{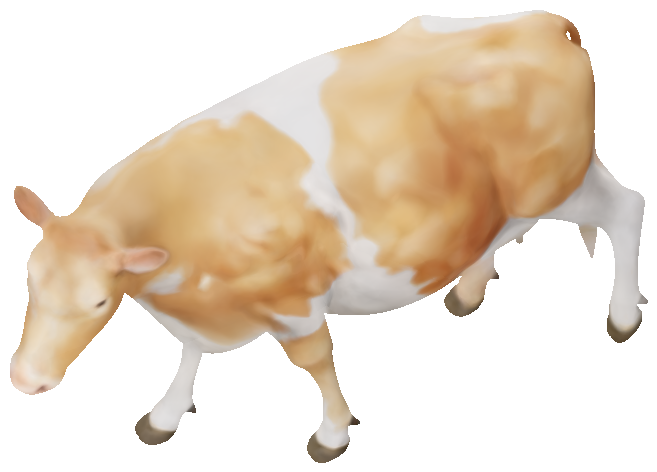}
        \caption{Rec.: Texture 2}
        \label{fig:color:rec2}
    \end{subfigure}
    \hfill
    \begin{subfigure}[b]{0.24\linewidth}
        \centering
        \includegraphics[width=\textwidth]{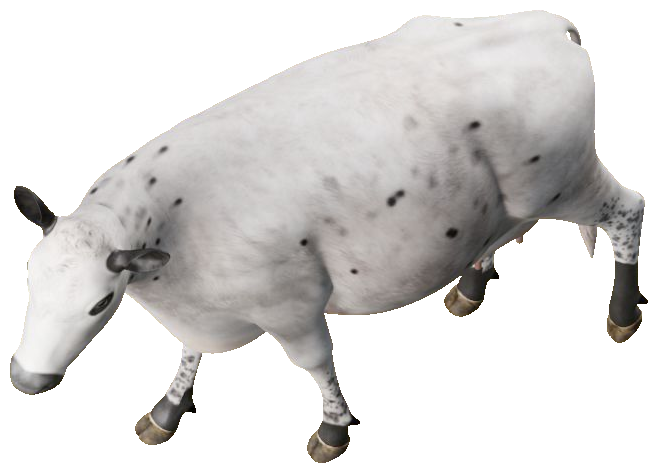}
        \caption{GT: Texture 3}
        \label{fig:color:gt3}
    \end{subfigure}
    \hfill
    \begin{subfigure}[b]{0.24\linewidth}
        \centering
        \includegraphics[width=\textwidth]{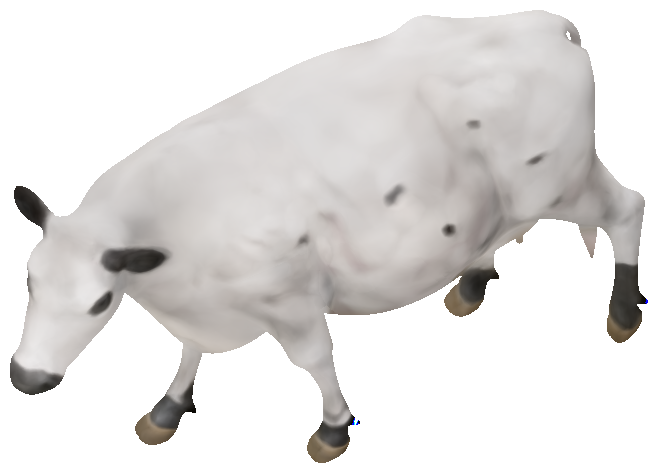}
        \caption{Rec.: Texture 3}
        \label{fig:color:rec3}
    \end{subfigure}
    \hfill
    \begin{subfigure}[b]{0.24\linewidth}
        \centering
        \includegraphics[width=\textwidth]{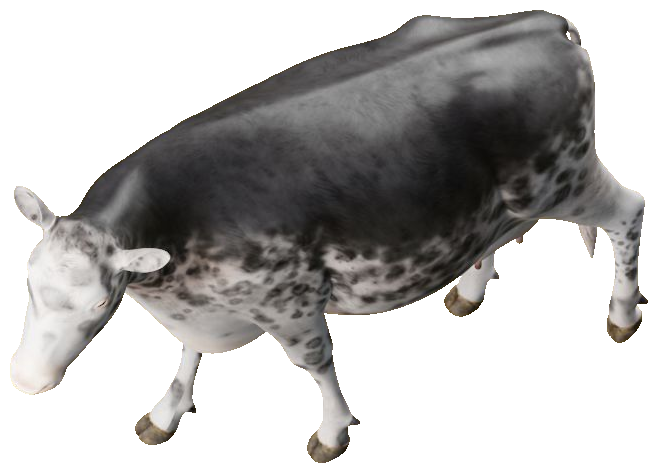}
        \caption{GT: Texture 4}
        \label{fig:color:gt4}
    \end{subfigure}
    \hfill
    \begin{subfigure}[b]{0.24\linewidth}
        \centering
        \includegraphics[width=\textwidth]{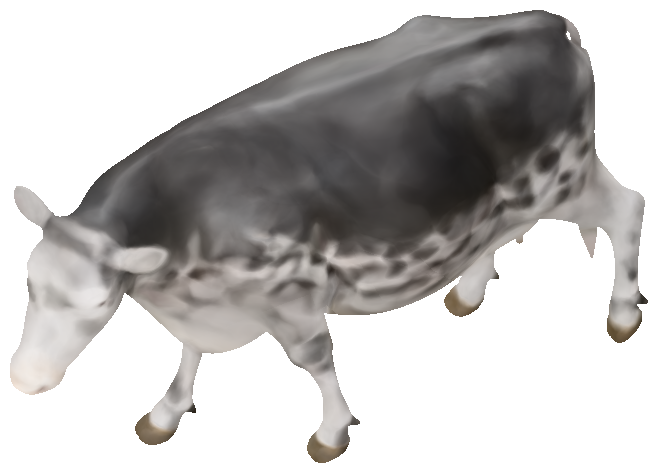}
        \caption{Rec.: Texture 4}
        \label{fig:color:rec4}
    \end{subfigure}
    \hfill
    \begin{subfigure}[b]{0.24\linewidth}
        \centering
        \includegraphics[width=\textwidth]{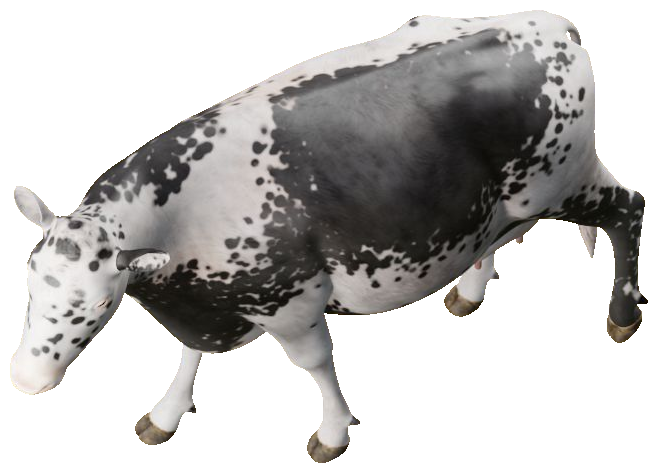}
        \caption{GT: Texture 5}
        \label{fig:color:gt5}
    \end{subfigure}
    \hfill
    \begin{subfigure}[b]{0.24\linewidth}
        \centering
        \includegraphics[width=\textwidth]{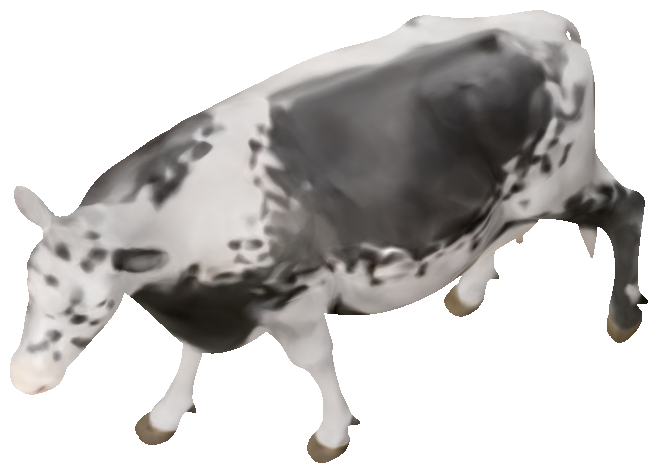}
        \caption{Rec.: Texture 5}
        \label{fig:color:rec5}
    \end{subfigure}
    \hfill
    \begin{subfigure}[b]{0.24\linewidth}
        \centering
        \includegraphics[width=\textwidth]{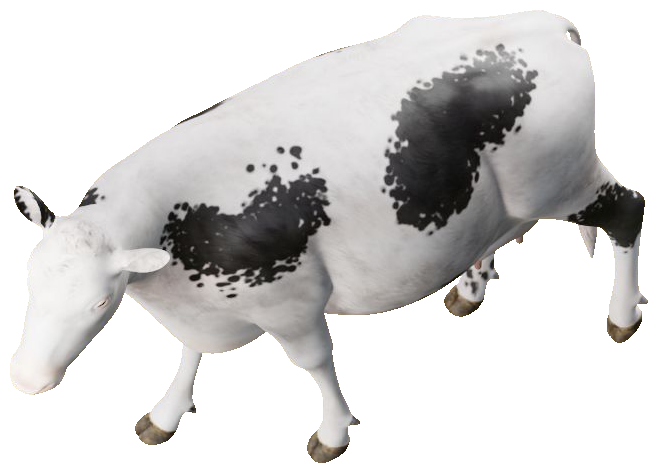}
        \caption{GT: Texture 6}
        \label{fig:color:gt6}
    \end{subfigure}
    \hfill
    \begin{subfigure}[b]{0.24\linewidth}
        \centering
        \includegraphics[width=\textwidth]{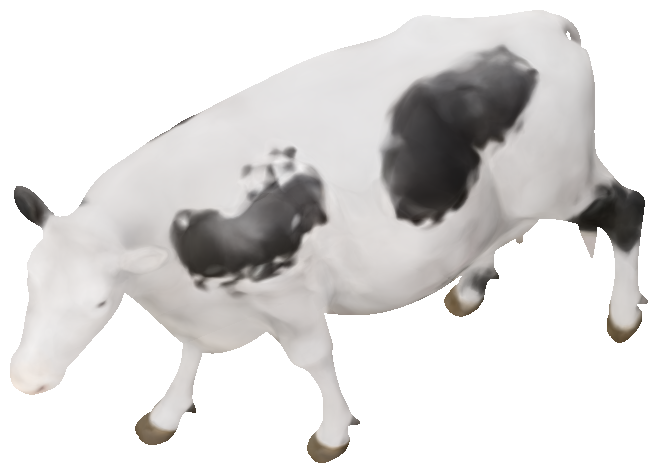}
        \caption{Rec.: Texture 6}
        \label{fig:color:rec6}
    \end{subfigure}
    \hfill
    \begin{subfigure}[b]{0.24\linewidth}
        \centering
        \includegraphics[width=\textwidth]{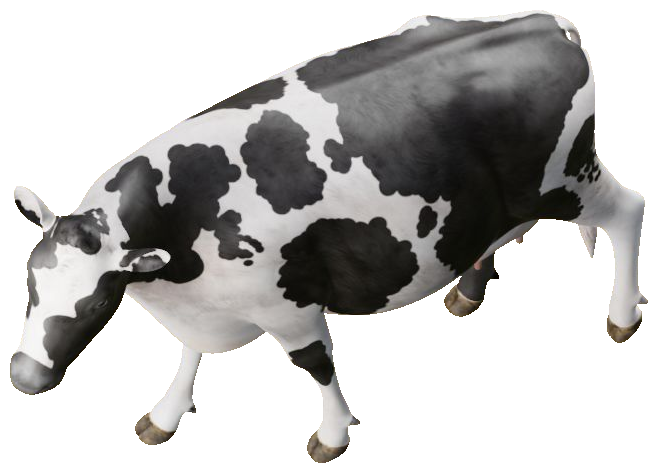}
        \caption{GT: Texture 7}
        \label{fig:color:gt7}
    \end{subfigure}
    \hfill
    \begin{subfigure}[b]{0.24\linewidth}
        \centering
        \includegraphics[width=\textwidth]{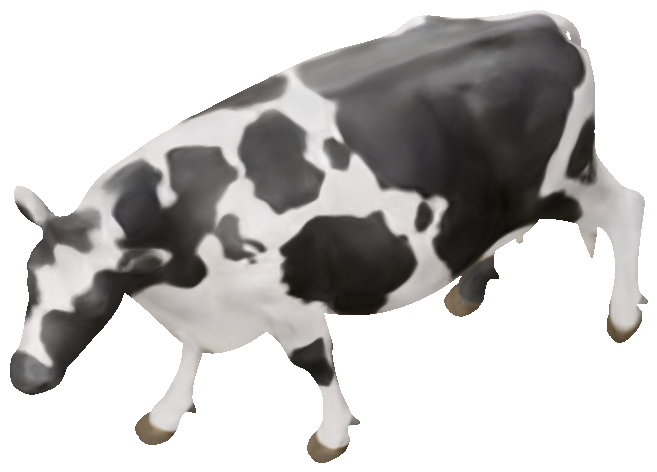}
        \caption{Rec.: Texture 7}
        \label{fig:color:rec7}
    \end{subfigure}
    \hfill
    \begin{subfigure}[b]{0.24\linewidth}
        \centering
        \includegraphics[width=\textwidth]{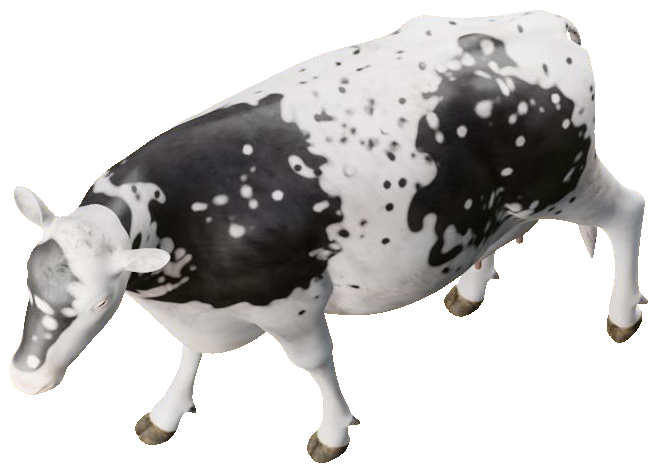}
        \caption{GT: Texture 8}
        \label{fig:color:gt8}
    \end{subfigure}
    \hfill
    \begin{subfigure}[b]{0.24\linewidth}
        \centering
        \includegraphics[width=\textwidth]{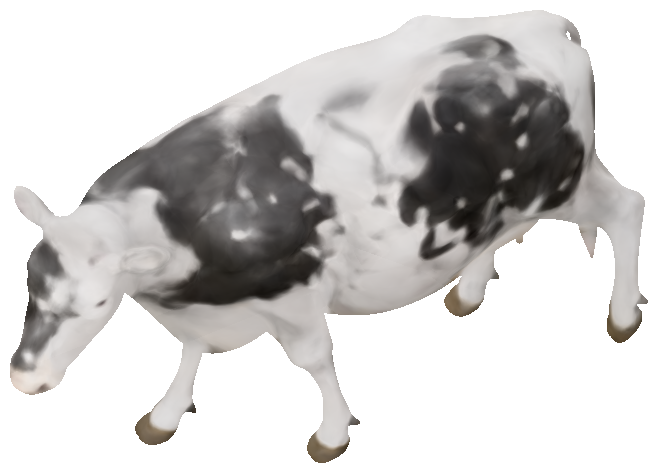}
        \caption{Rec.: Texture 8}
        \label{fig:color:rec8}
    \end{subfigure}
    \hfill
    \begin{subfigure}[b]{0.24\linewidth}
        \centering
        \includegraphics[width=\textwidth]{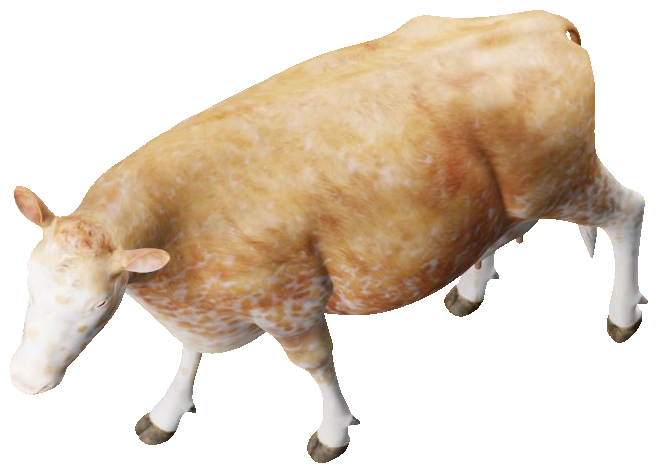}
        \caption{GT: Texture 1}
        \label{fig:color:gt1}
    \end{subfigure}
    \hfill
    \begin{subfigure}[b]{0.24\linewidth}
        \centering
        \includegraphics[width=\textwidth]{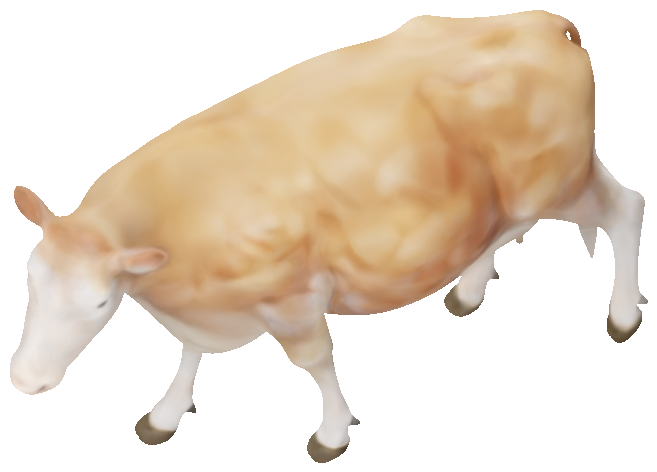}
        \caption{Rec.: Texture 1}
        \label{fig:color:rec1}
    \end{subfigure}
    \hfill
    \begin{subfigure}[b]{0.24\linewidth}
        \centering
        \includegraphics[width=\textwidth]{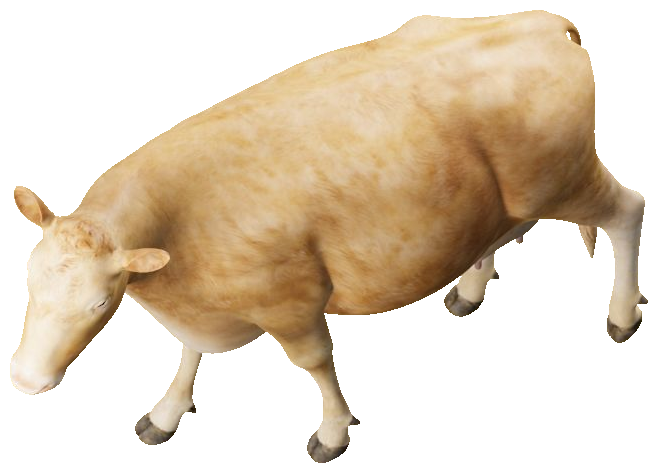}
        \caption{GT: Texture 11}
        \label{fig:color:gt11}
    \end{subfigure}
    \hfill
    \begin{subfigure}[b]{0.24\linewidth}
        \centering
        \includegraphics[width=\textwidth]{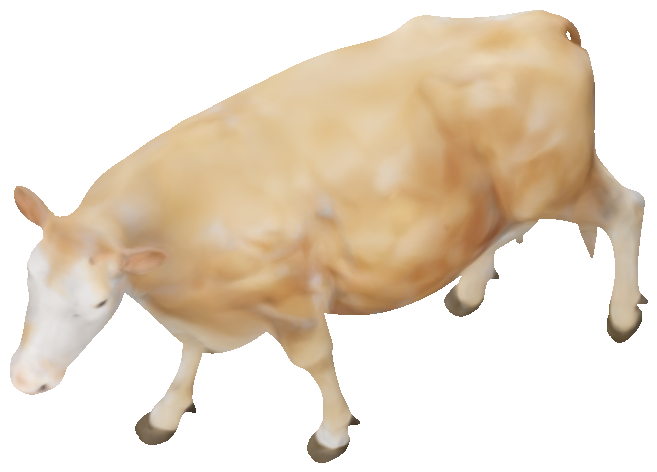}
        \caption{Rec.: Texture 11}
        \label{fig:color:rec11}
    \end{subfigure}
    \hfill
    \begin{subfigure}[b]{0.24\linewidth}
        \centering
        \includegraphics[width=\textwidth]{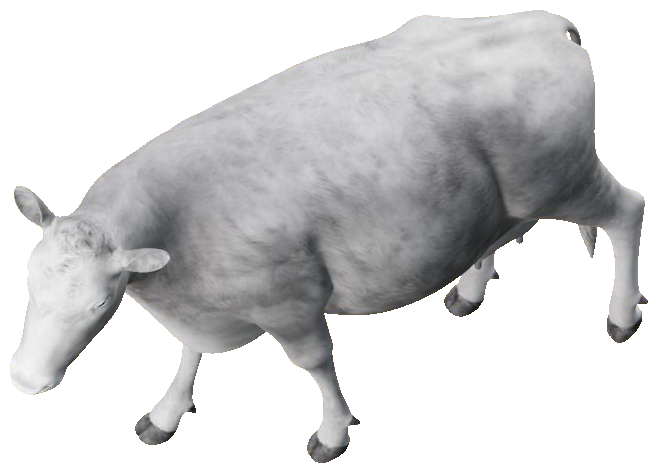}
        \caption{GT: Texture 9}
        \label{fig:color:gt9}
    \end{subfigure}
    \hfill
    \begin{subfigure}[b]{0.24\linewidth}
        \centering
        \includegraphics[width=\textwidth]{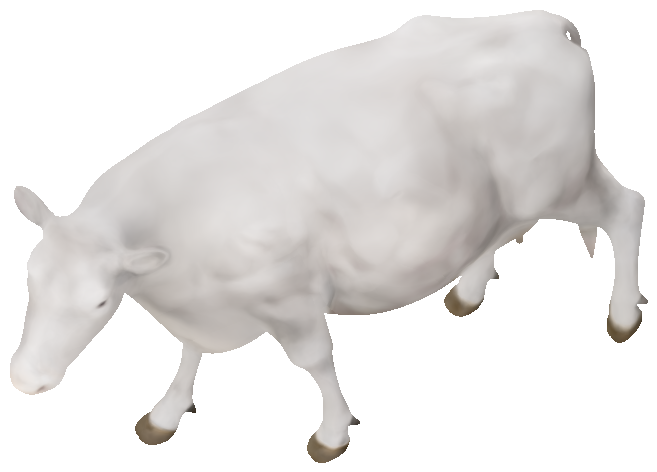}
        \caption{Rec.: Texture 9}
        \label{fig:color:rec9}
    \end{subfigure}
    \hfill
    \begin{subfigure}[b]{0.24\linewidth}
        \centering
        \includegraphics[width=\textwidth]{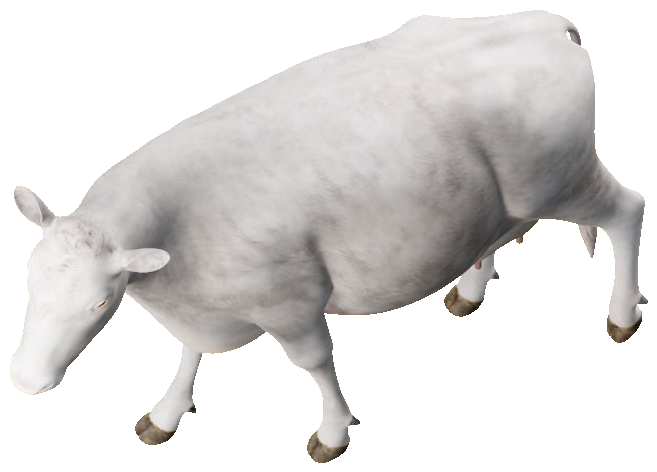}
        \caption{GT: Texture 10}
        \label{fig:color:gt10}
    \end{subfigure}
    \hfill
    \begin{subfigure}[b]{0.24\linewidth}
        \centering
        \includegraphics[width=\textwidth]{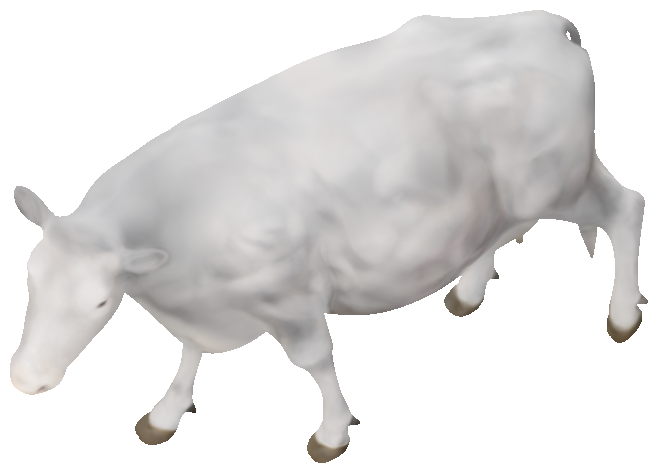}
        \caption{Rec.: Texture 10}
        \label{fig:color:rec10}
    \end{subfigure}
    \caption{
        \textbf{Novel Pose Synthesis.}
        We reconstruct images for novel poses from our~\dataabr test set, which have not been seen during training.
        We show one example of ground truth (GT) and our reconstruction (Rec.) for each distinct texture in our dataset.
    }
    \label{fig:color}
\end{figure*}

\begin{table}[t]
    \centering
    \small
    \begin{tabular}{cc|c}
        \toprule
        Sy. Cows & Texture & Color PSNR [dB] \\
        \midrule
        Texture 0 & Holstein-Friesian & $17.39$ \\
        Texture 1 & Brown Spotted & $20.90$ \\
        Texture 2 & Brown Freckled & $20.73$ \\
        Texture 3 & White W. Few Black Freckles & $21.04$ \\
        Texture 4 & Black Freckled & $18.98$ \\
        Texture 5 & Holstein-Friesian W. Freckles & $17.42$ \\
        Texture 6 & Holstein-Friesian W. Freckles & $18.64$ \\
        Texture 7 & Holstein-Friesian & $16.51$ \\
        Texture 8 & Holstein-Friesian W. Freckles & $17.44$ \\
        Texture 9 & Gray & $20.38$ \\
        Texture 10 & White & $20.86$ \\
        Texture 11 & Limousin & $21.91$ \\
        \midrule
        \textbf{Average} & & $\mathbf{19.35}$ \\
        \bottomrule
    \end{tabular}
    \caption{{\em Detailed quantitative results for novel pose synthesis on our~\dataabr test set}.
         We report PSNR [dB] for the reconstructed RGB images.
    }
    \label{tab:supp:quantitative-color}
\end{table}
\subheading{Novel Pose Synthesis}
We report detailed quantitative results for novel pose synthesis on our~\dataabr test set in~\cref{tab:supp:quantitative-color}.
%
Qualitative results for novel pose synthesis with~\paperabr on our~\dataabr test set are shown in~\cref{fig:color}.
The dataset contains two challenging texture pairs, which are
textures~$9, 10$ (cf.~\cref{fig:color:gt9,fig:color:rec9,fig:color:gt10,fig:color:rec10}) and $1, 11$ (cf.~\cref{fig:color:gt1,fig:color:rec1,fig:color:gt11,fig:color:rec11}).
While the reconstructions of these pairs look similar here, this is also the case for the ground truth.
%

\begin{figure*}[t]
    \centering
    \begin{subfigure}[b]{0.49\linewidth}
        \centering
        \includegraphics[width=0.49\textwidth,trim={3cm 4cm 2cm 4cm},clip]{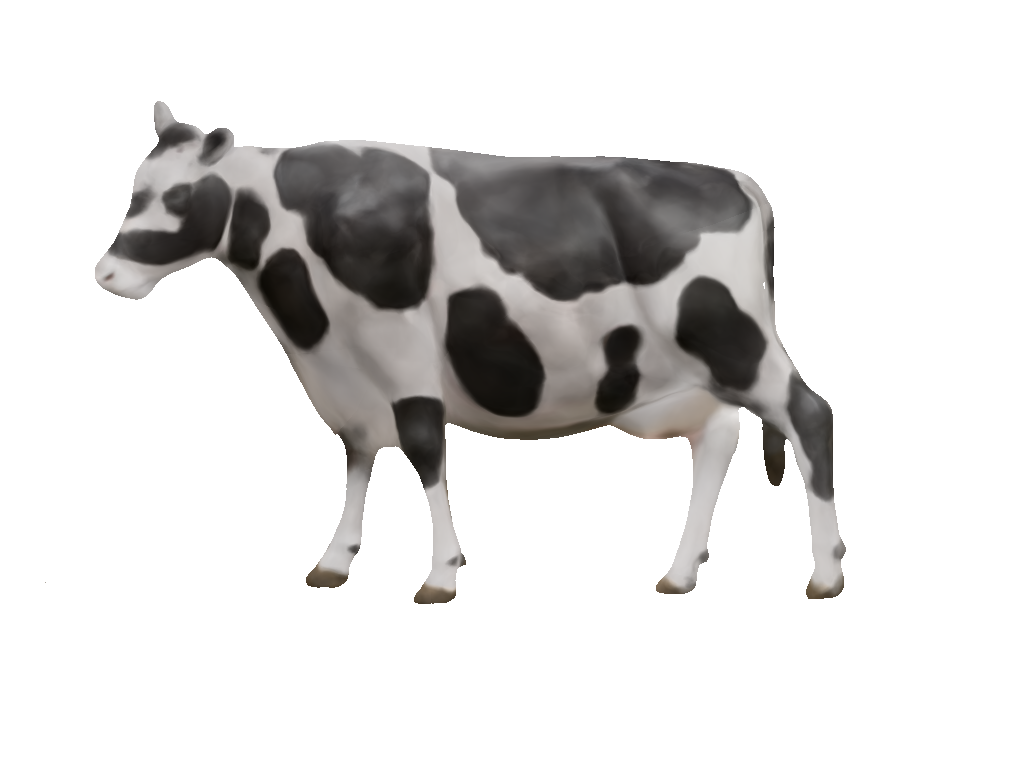}
        \includegraphics[width=0.49\textwidth,trim={3cm 4cm 2cm 4cm},clip]{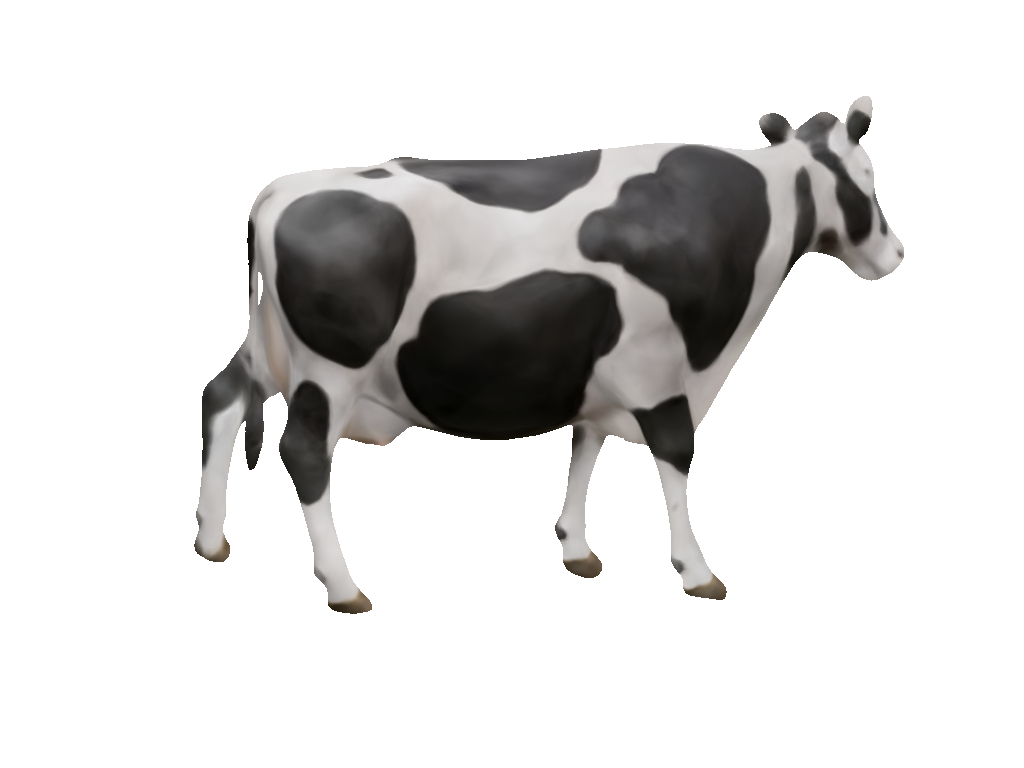}
        \caption{Texture $0$}
    \end{subfigure}
    \hfill
    \begin{subfigure}[b]{0.49\linewidth}
        \centering
        \includegraphics[width=0.49\textwidth,trim={3cm 4cm 2cm 4cm},clip]{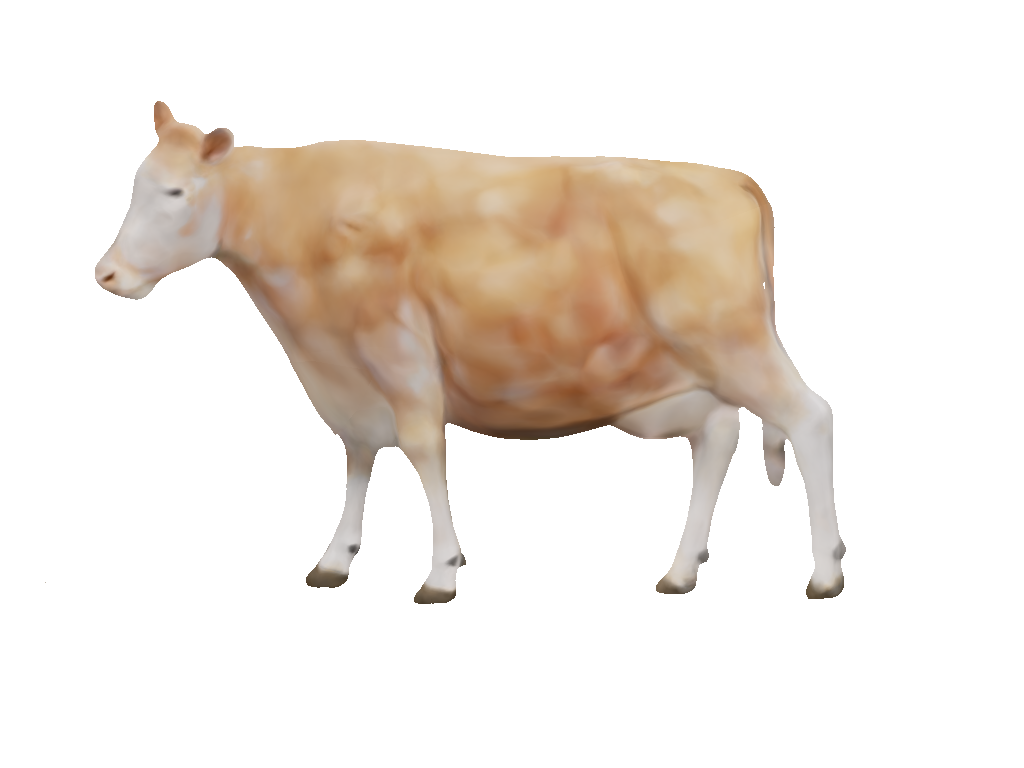}
        \includegraphics[width=0.49\textwidth,trim={3cm 4cm 2cm 4cm},clip]{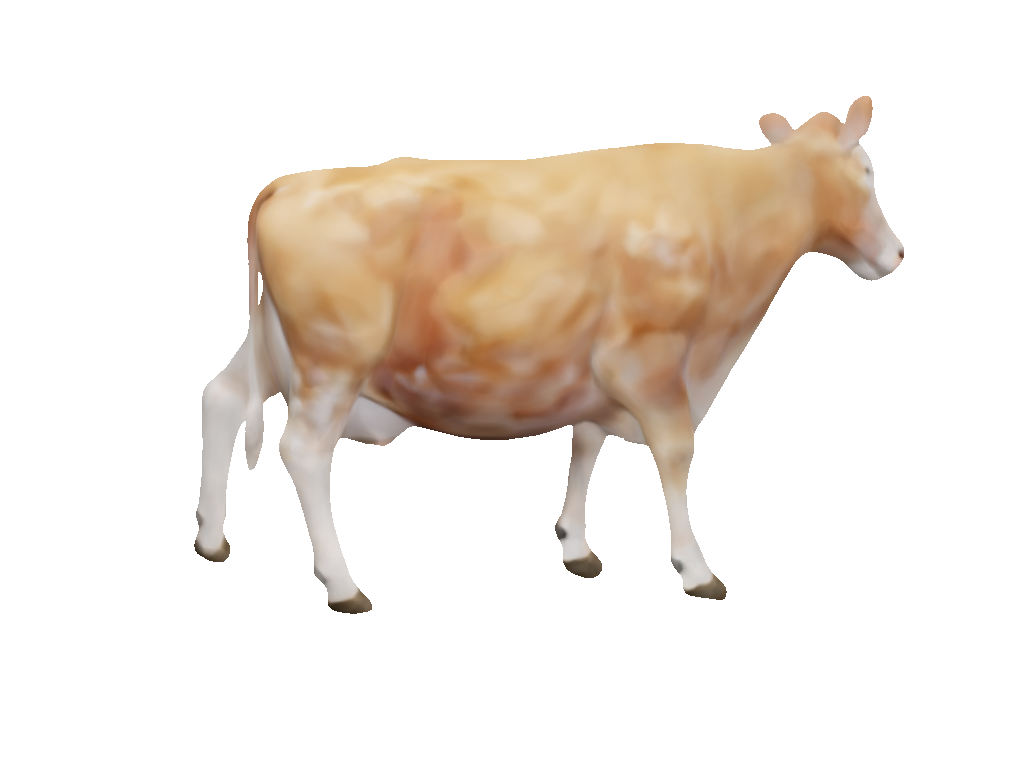}
        \caption{Texture $1$}
        \label{fig:supp:npnv:1}
    \end{subfigure}
    \hfill
    \begin{subfigure}[b]{0.49\linewidth}
        \centering
        \includegraphics[width=0.49\textwidth,trim={3cm 4cm 2cm 4cm},clip]{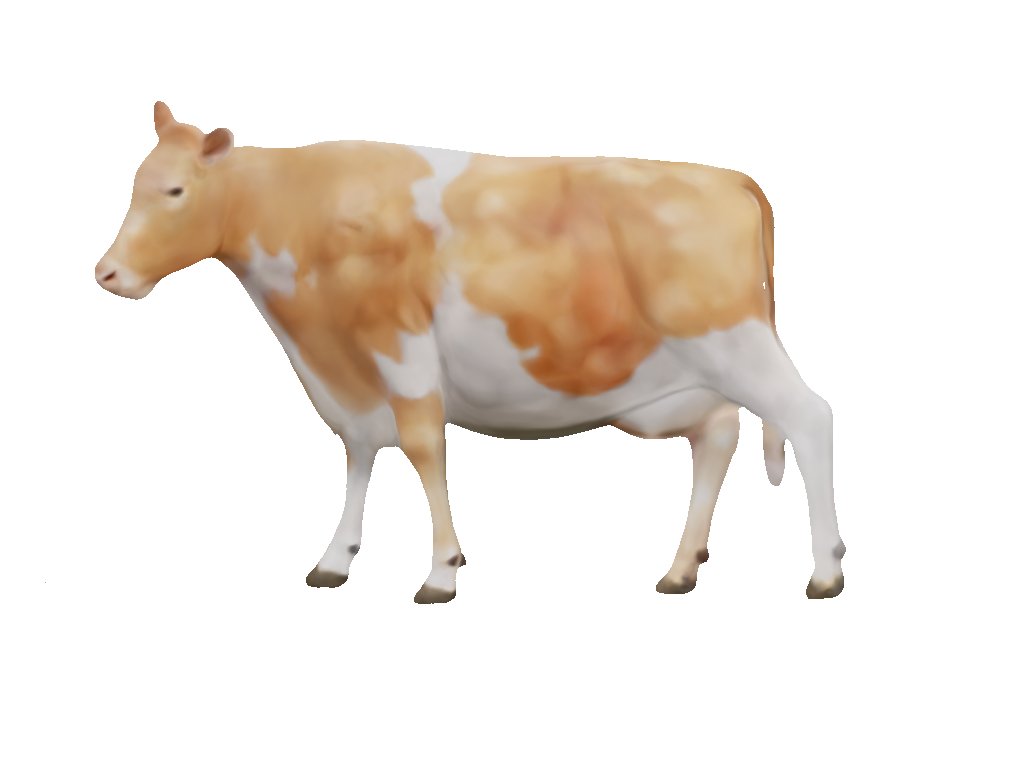}
        \includegraphics[width=0.49\textwidth,trim={3cm 4cm 2cm 4cm},clip]{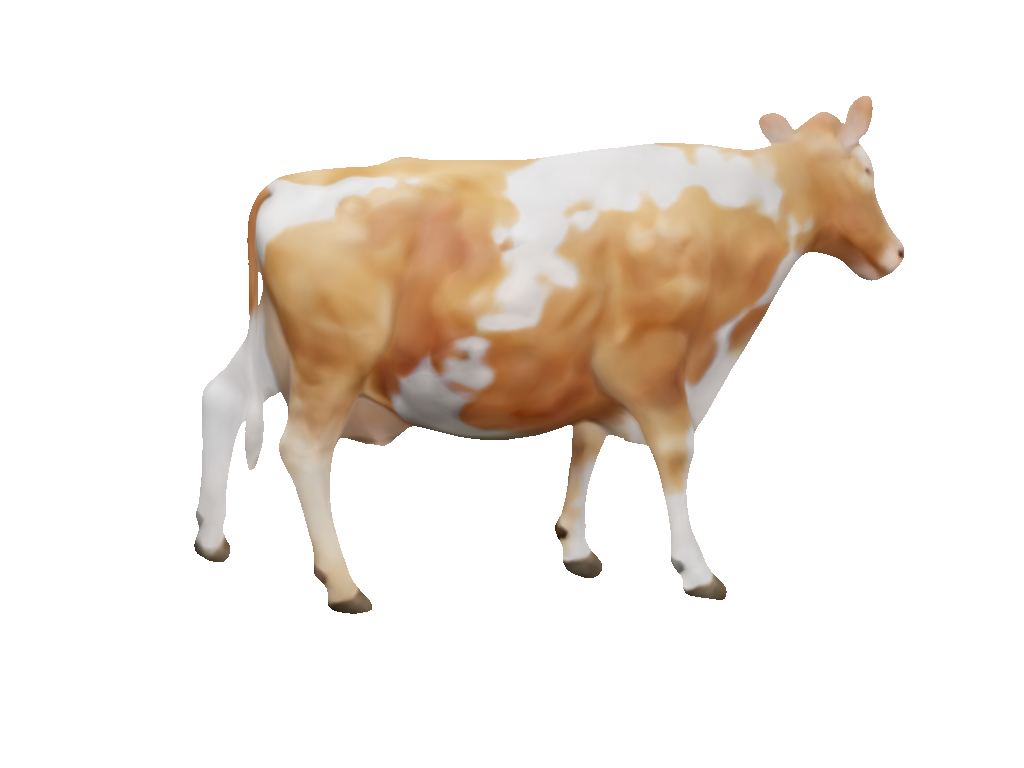}
        \caption{Texture $2$}
    \end{subfigure}
    \hfill
    \begin{subfigure}[b]{0.49\linewidth}
        \centering
        \includegraphics[width=0.49\textwidth,trim={3cm 4cm 2cm 4cm},clip]{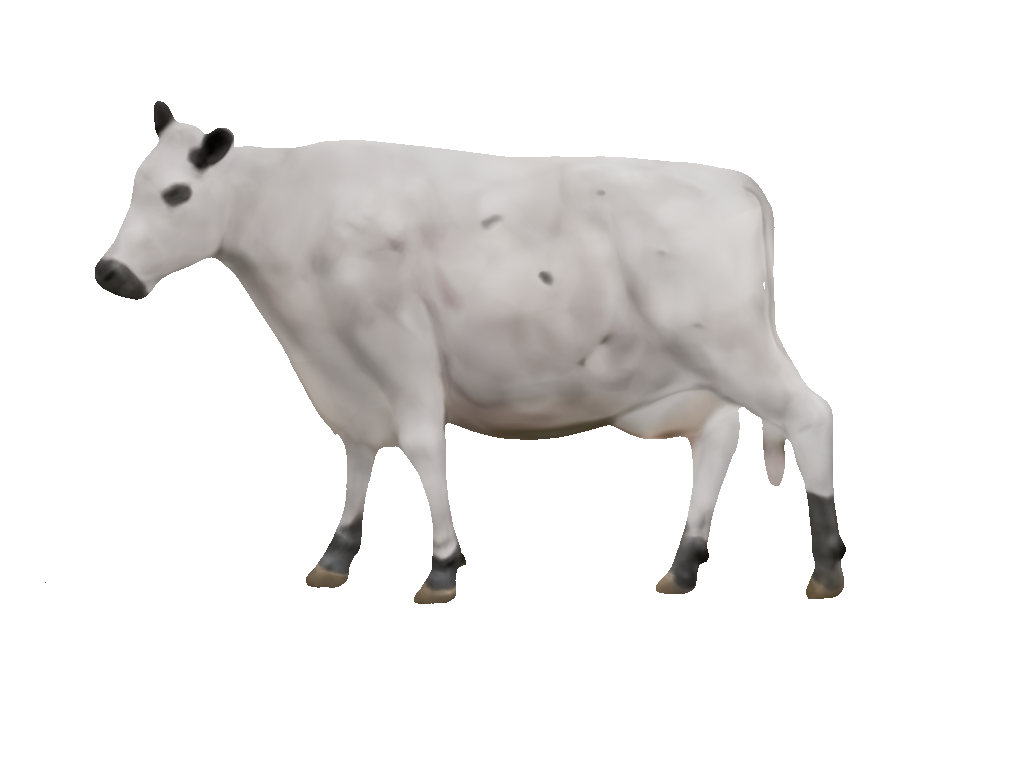}
        \includegraphics[width=0.49\textwidth,trim={3cm 4cm 2cm 4cm},clip]{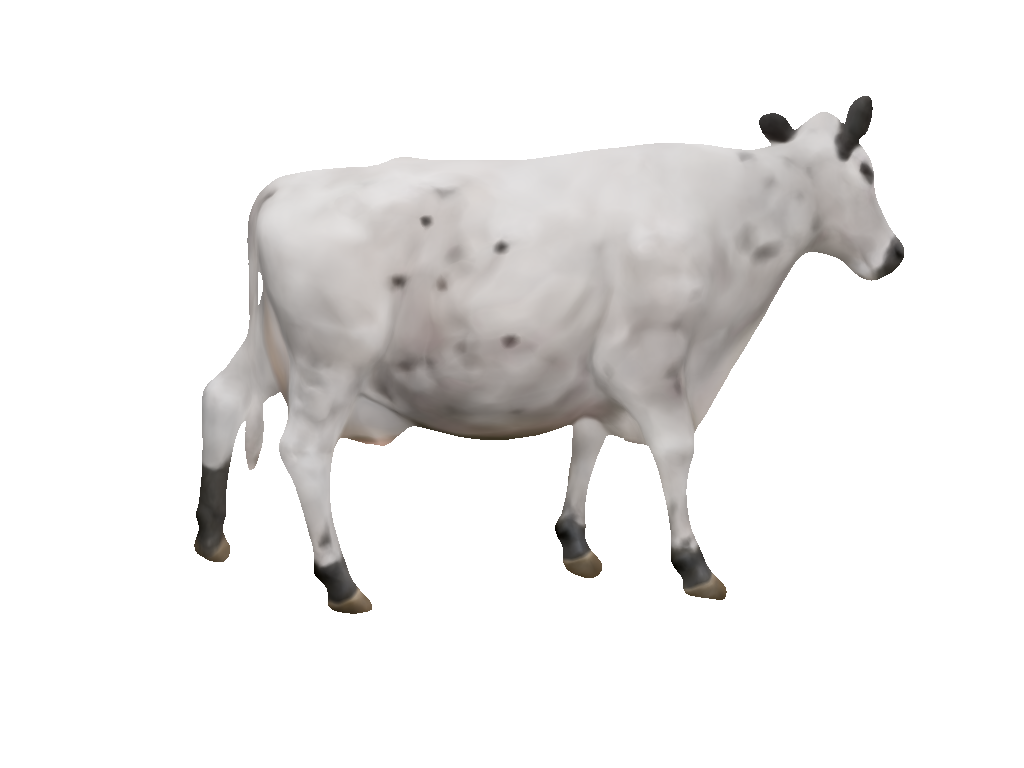}
        \caption{Texture $3$}
    \end{subfigure}
    \hfill
    \begin{subfigure}[b]{0.49\linewidth}
        \centering
        \includegraphics[width=0.49\textwidth,trim={3cm 4cm 2cm 4cm},clip]{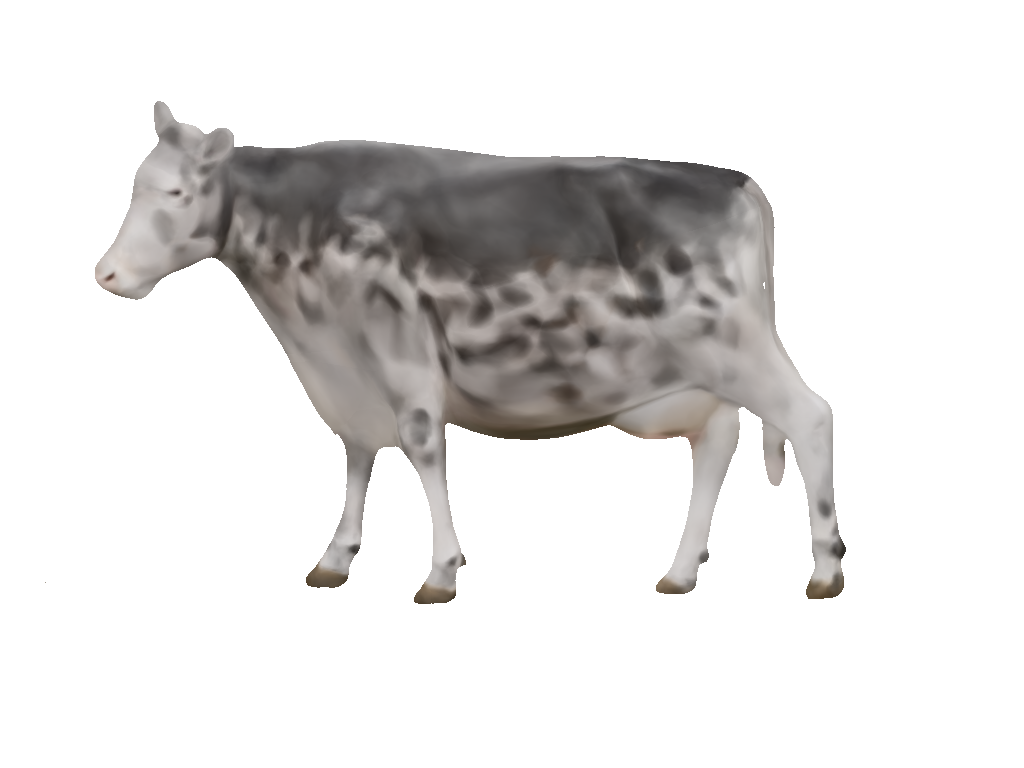}
        \includegraphics[width=0.49\textwidth,trim={3cm 4cm 2cm 4cm},clip]{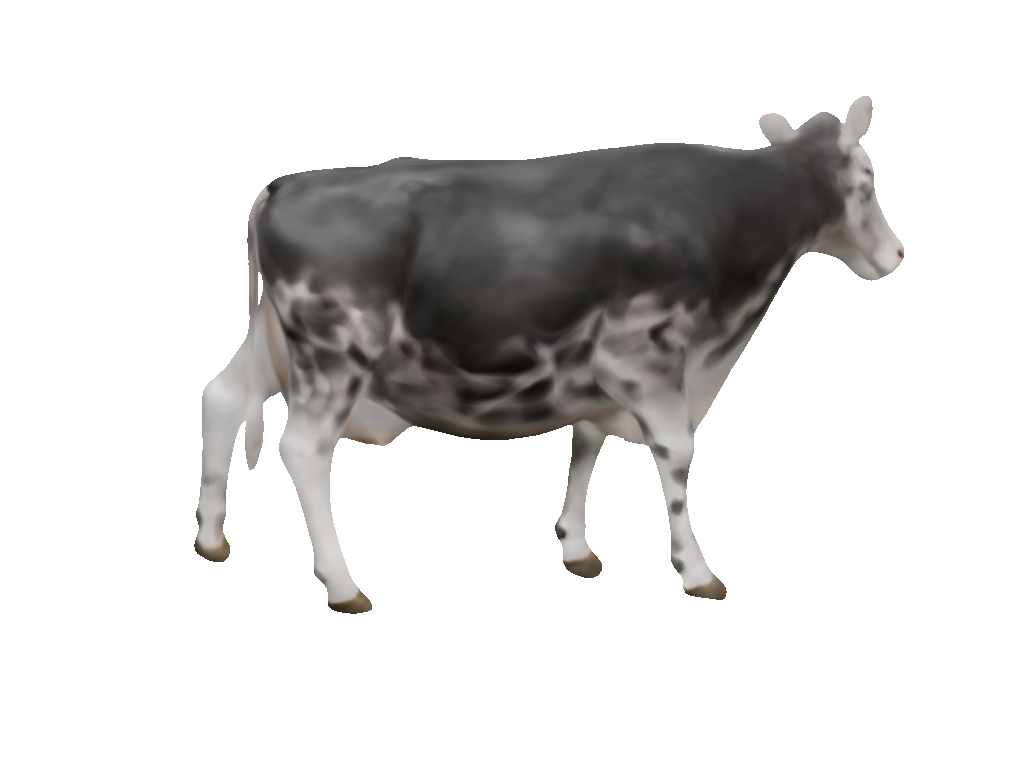}
        \caption{Texture $4$}
    \end{subfigure}
    \hfill
    \begin{subfigure}[b]{0.49\linewidth}
        \centering
        \includegraphics[width=0.49\textwidth,trim={3cm 4cm 2cm 4cm},clip]{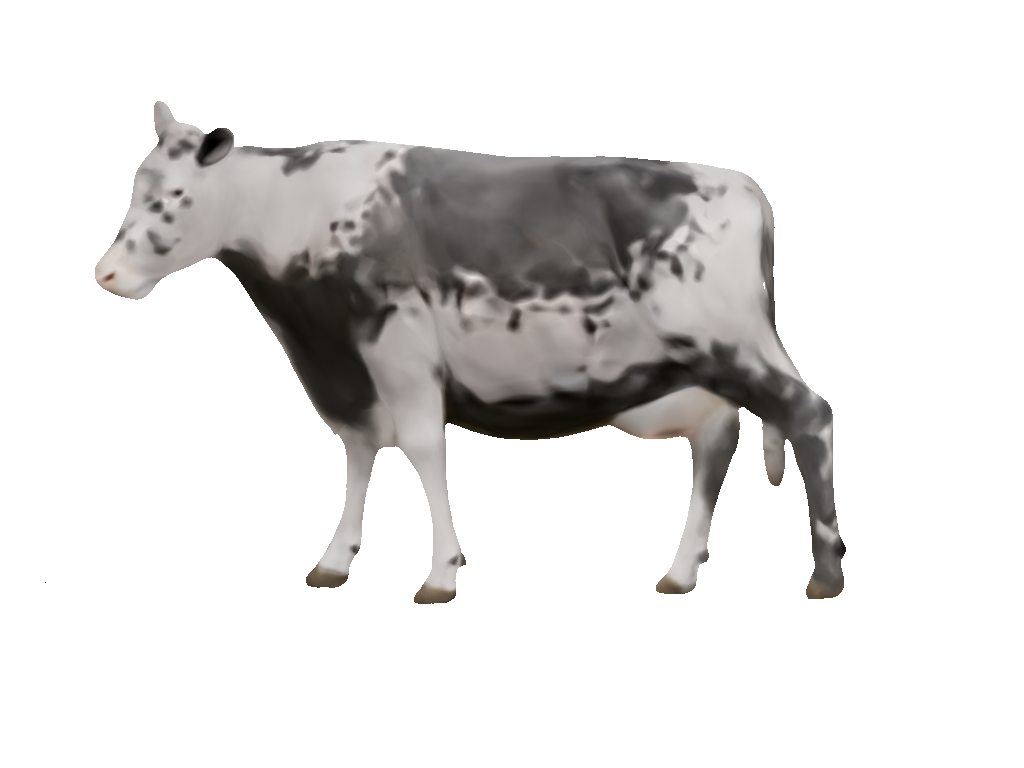}
        \includegraphics[width=0.49\textwidth,trim={3cm 4cm 2cm 4cm},clip]{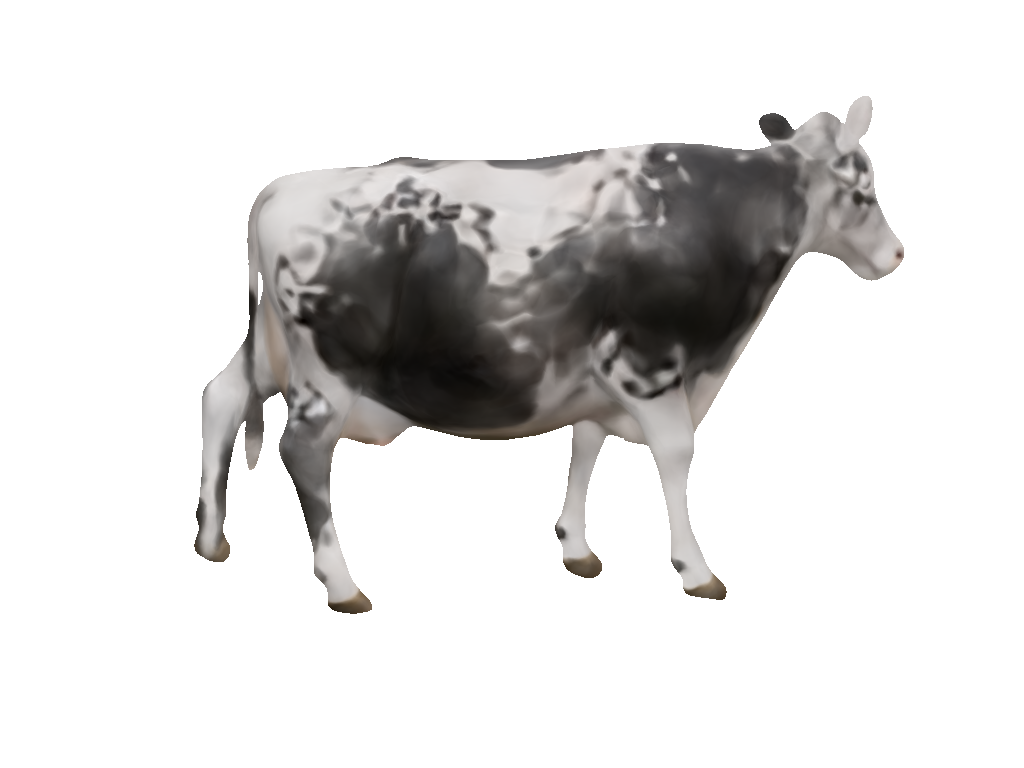}
        \caption{Texture $5$}
    \end{subfigure}
    \hfill
    \begin{subfigure}[b]{0.49\linewidth}
        \centering
        \includegraphics[width=0.49\textwidth,trim={3cm 4cm 2cm 4cm},clip]{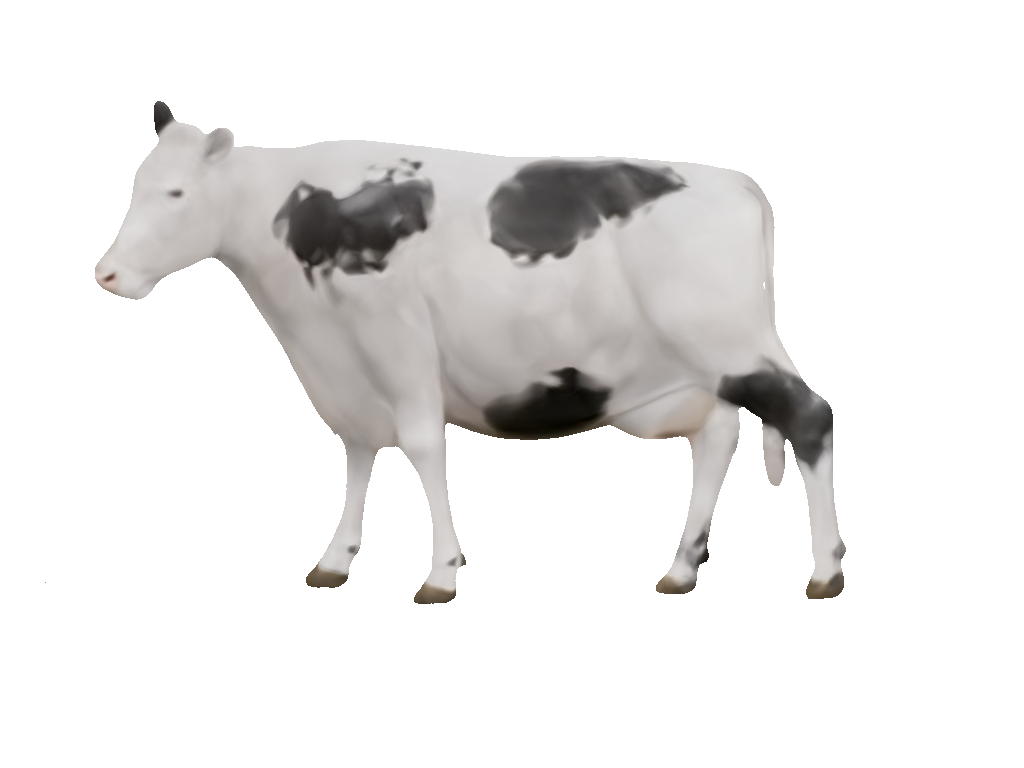}
        \includegraphics[width=0.49\textwidth,trim={3cm 4cm 2cm 4cm},clip]{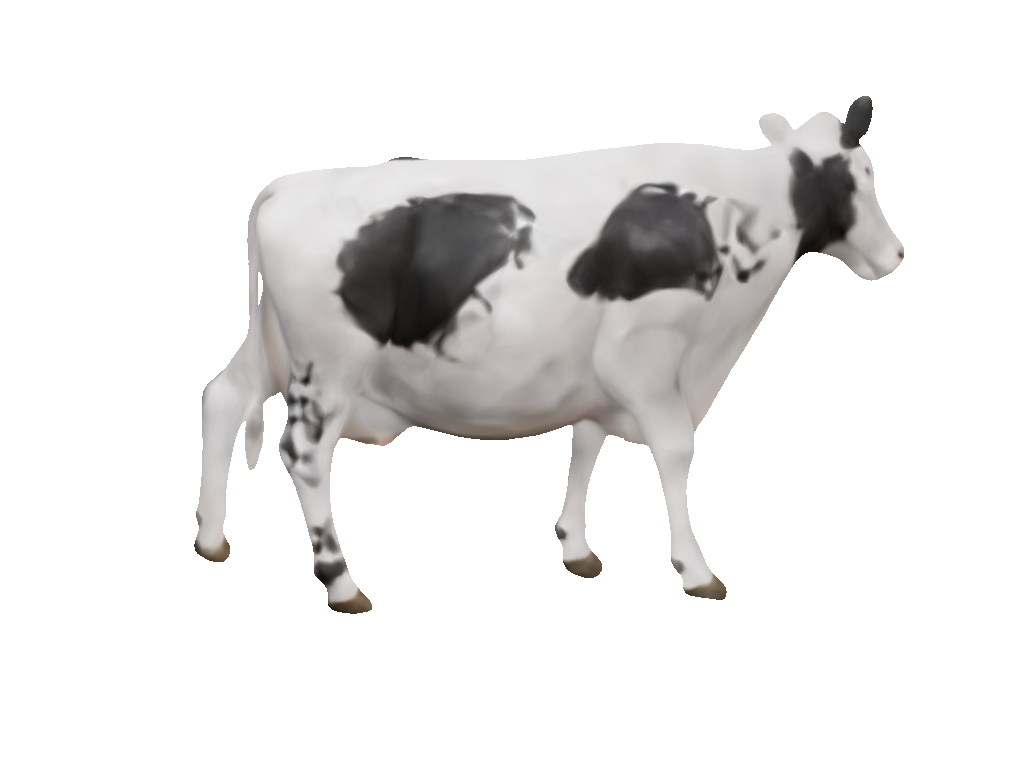}
        \caption{Texture $6$}
    \end{subfigure}
    \hfill
        \begin{subfigure}[b]{0.49\linewidth}
        \centering
        \includegraphics[width=0.49\textwidth,trim={3cm 4cm 2cm 4cm},clip]{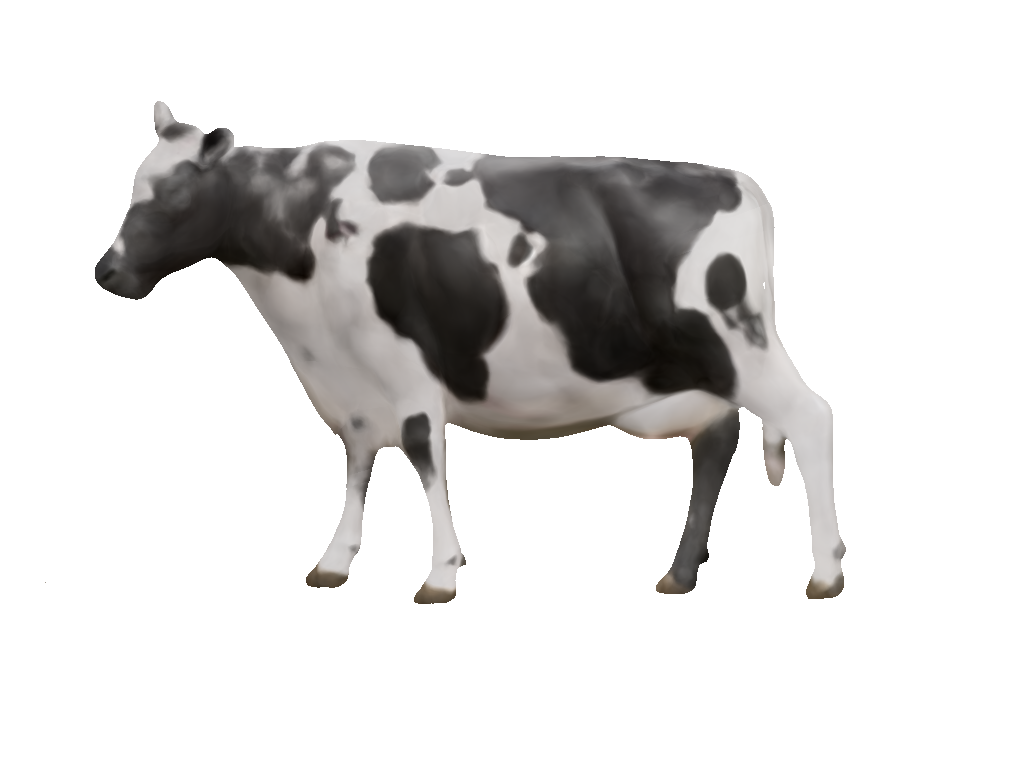}
        \includegraphics[width=0.49\textwidth,trim={3cm 4cm 2cm 4cm},clip]{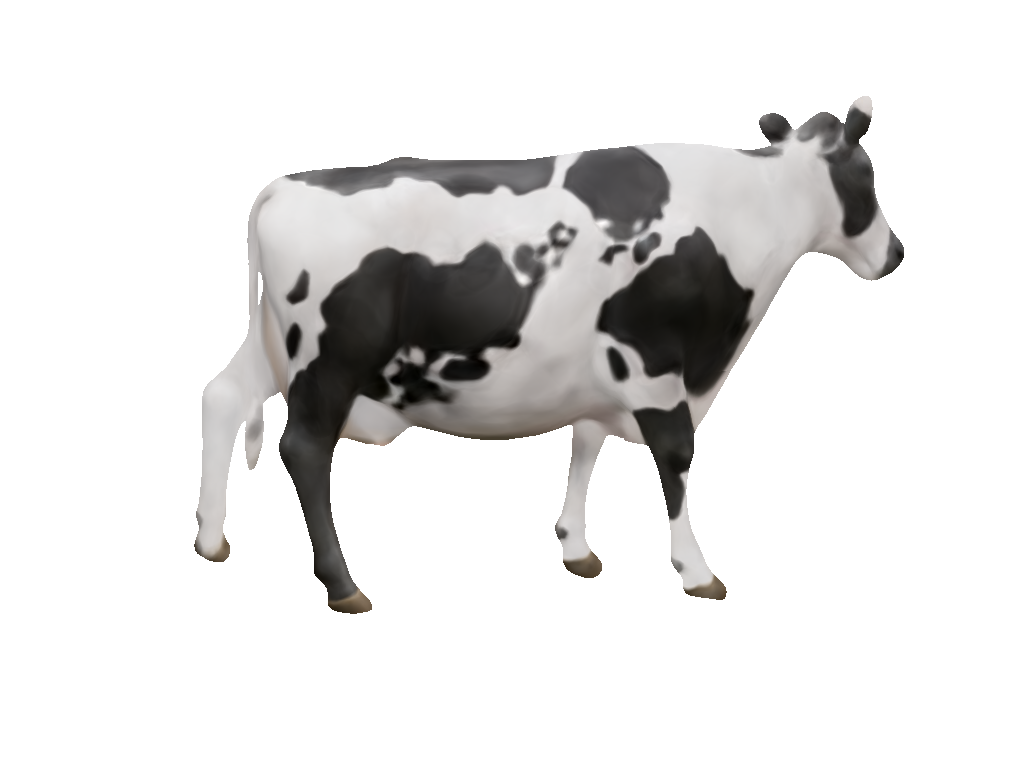}
        \caption{Texture $7$}
    \end{subfigure}
    \hfill
        \begin{subfigure}[b]{0.49\linewidth}
        \centering
        \includegraphics[width=0.49\textwidth,trim={3cm 4cm 2cm 4cm},clip]{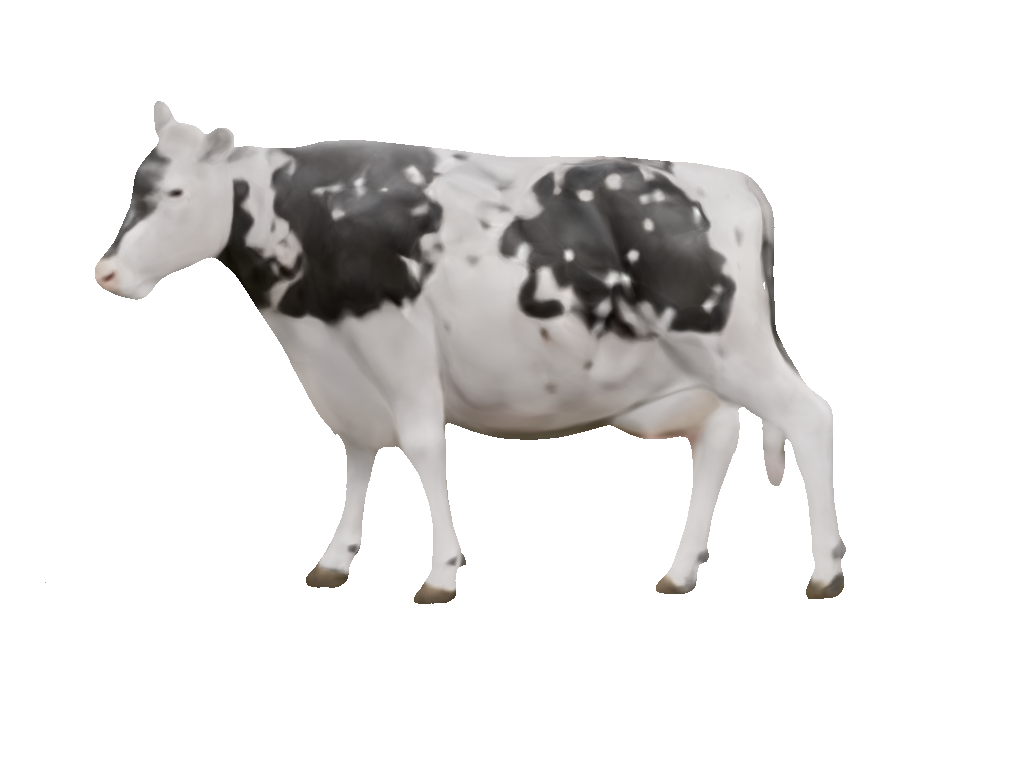}
        \includegraphics[width=0.49\textwidth,trim={3cm 4cm 2cm 4cm},clip]{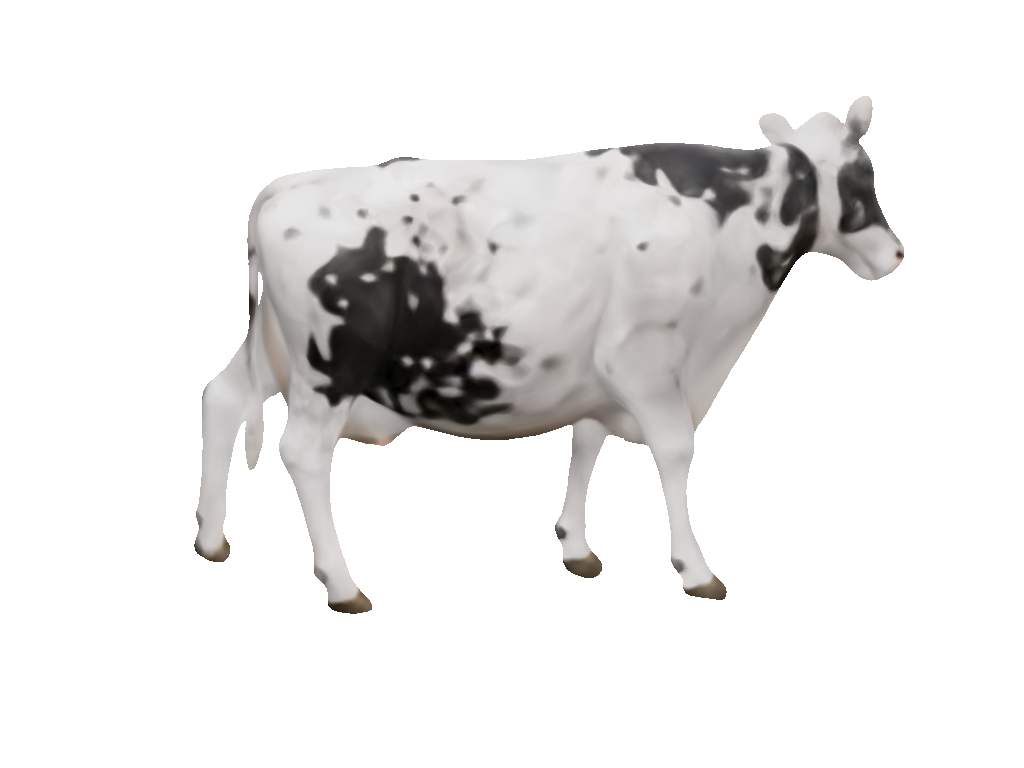}
        \caption{Texture $8$}
    \end{subfigure}
    \hfill
        \begin{subfigure}[b]{0.49\linewidth}
        \centering
        \includegraphics[width=0.49\textwidth,trim={3cm 4cm 2cm 4cm},clip]{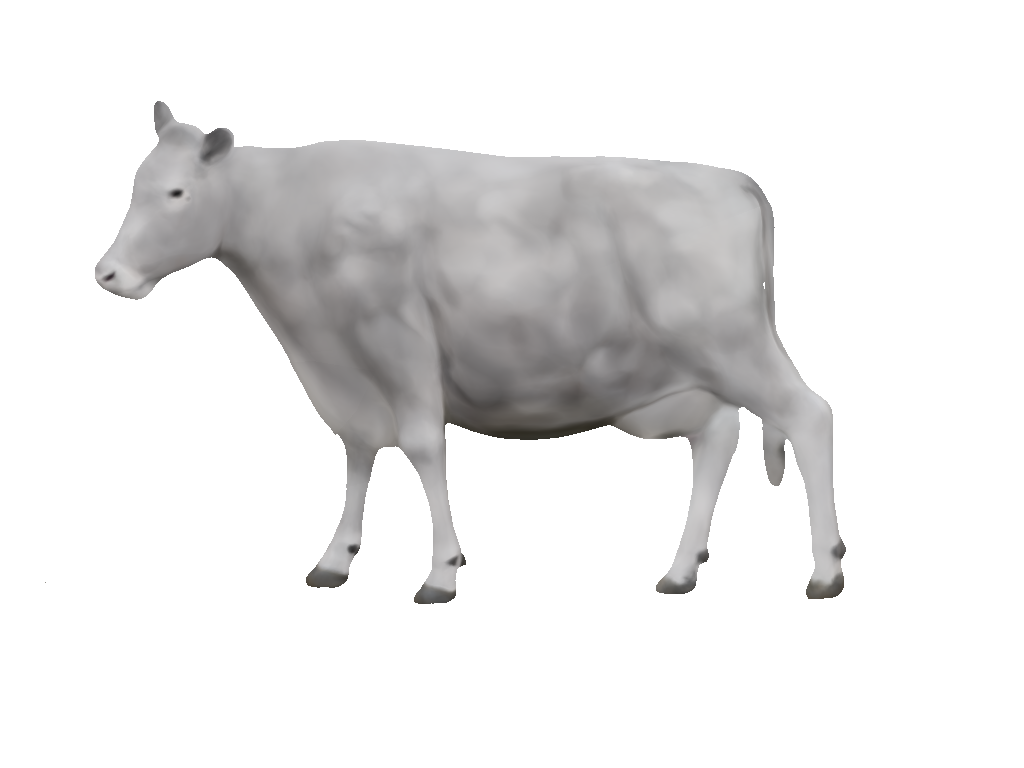}
        \includegraphics[width=0.49\textwidth,trim={3cm 4cm 2cm 4cm},clip]{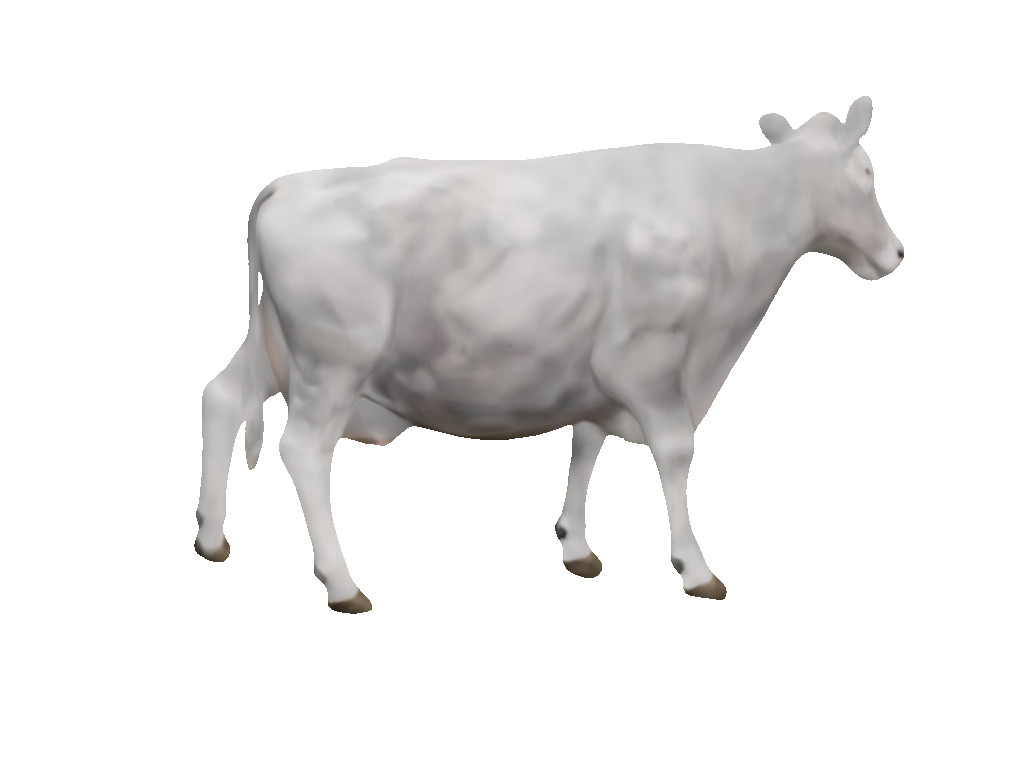}
        \caption{Texture $9$}
        \label{fig:supp:npnv:9}
    \end{subfigure}
    \hfill
        \begin{subfigure}[b]{0.49\linewidth}
        \centering
        \includegraphics[width=0.49\textwidth,trim={3cm 4cm 2cm 4cm},clip]{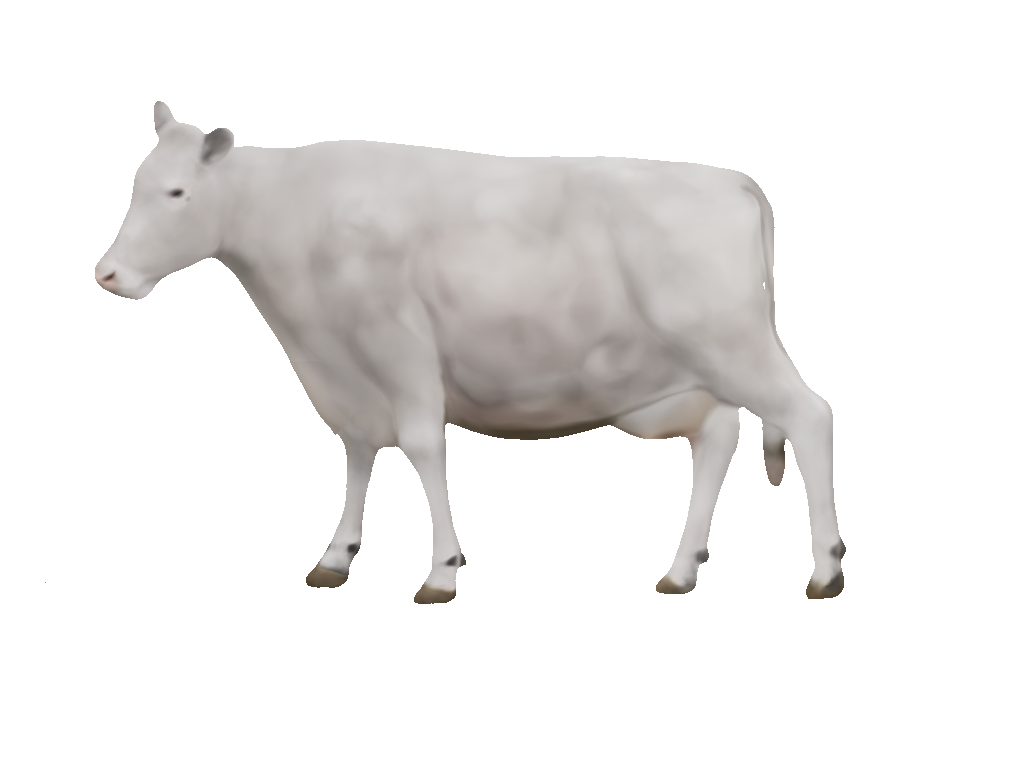}
        \includegraphics[width=0.49\textwidth,trim={3cm 4cm 2cm 4cm},clip]{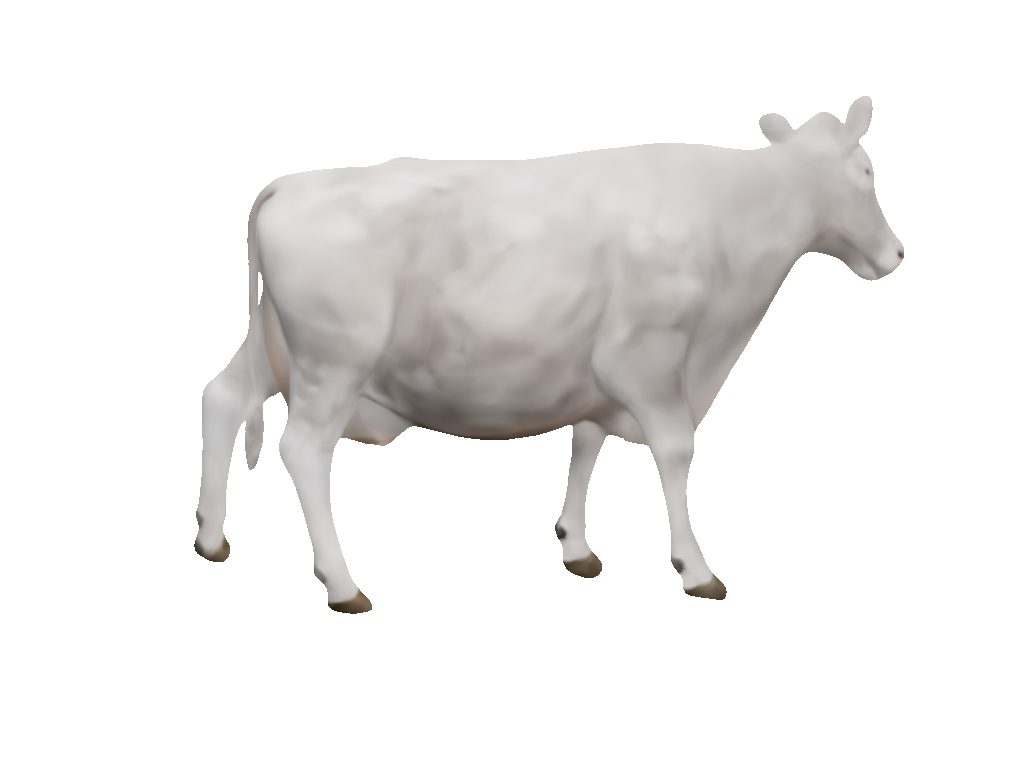}
        \caption{Texture $10$}
        \label{fig:supp:npnv:10}
    \end{subfigure}
    \hfill
        \begin{subfigure}[b]{0.49\linewidth}
        \centering
        \includegraphics[width=0.49\textwidth,trim={3cm 4cm 2cm 4cm},clip]{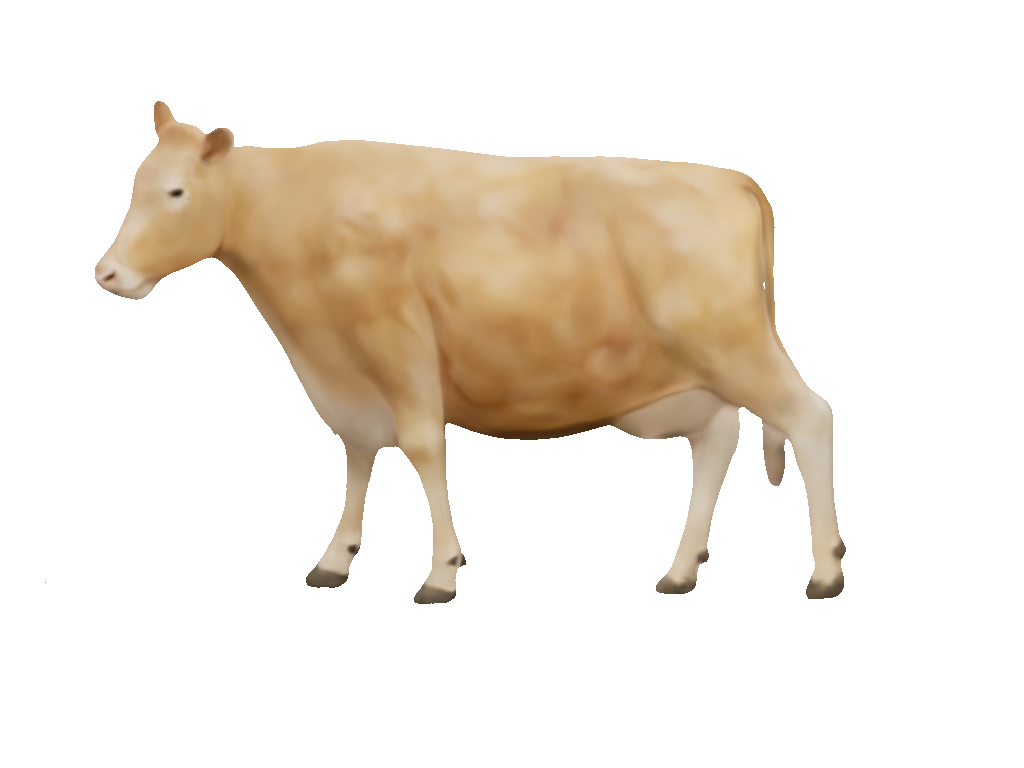}
        \includegraphics[width=0.49\textwidth,trim={3cm 4cm 2cm 4cm},clip]{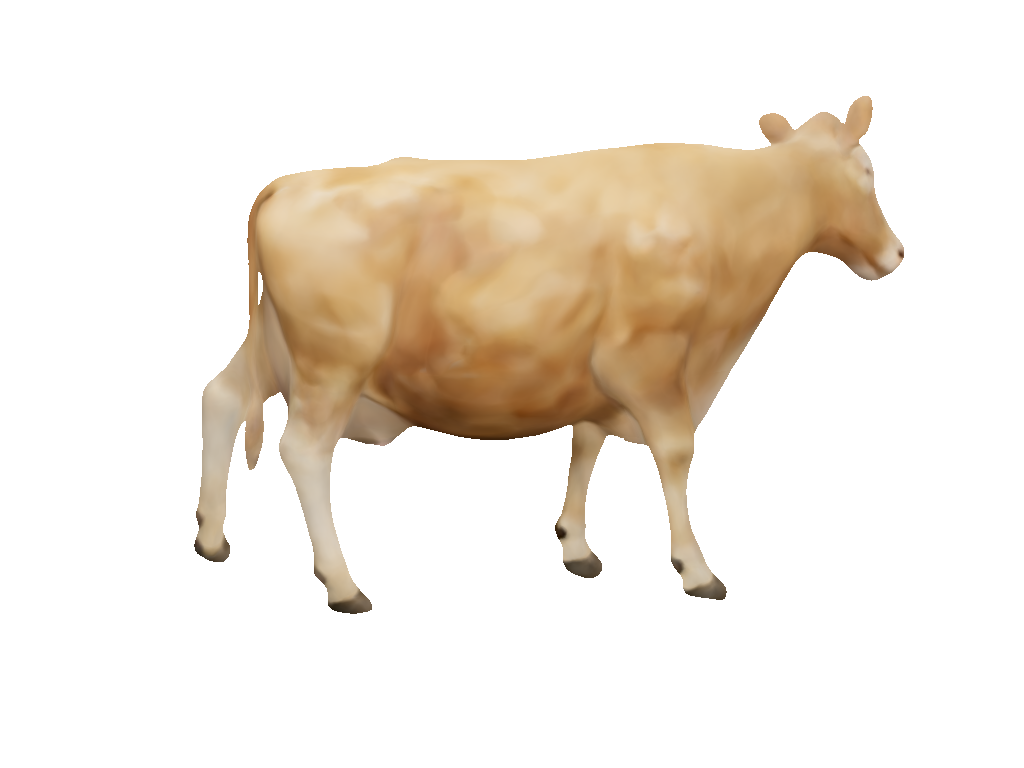}
        \caption{Texture $11$}
        \label{fig:supp:npnv:11}
    \end{subfigure}
    \caption{
        \textbf{Novel Pose and Novel View Synthesis.}
        We reconstruct images for novel views and novel poses from our~\dataabr test set, which have not been seen during training.
        We show two examples for each distinct texture in our dataset.
    }
    \label{fig:supp:novel-pose-novel-views}
\end{figure*}

\subheading{Novel Pose and Novel View Synthesis}
We also reconstruct images for novel views and novel poses from our~\dataabr test set, which have not been seen during training.
In~\cref{fig:supp:novel-pose-novel-views} we show two examples for each distinct texture in our dataset.
We note that this time even the challenging texture pairs $9$, $10$ and $1$, $11$ are reconstructed correctly (cf.~\cref{fig:supp:npnv:9,fig:supp:npnv:10} and~\cref{fig:supp:npnv:1,fig:supp:npnv:11} respectively).

\subsection{Re-identification}
\label{sec:supp:reid}
\subheading{Occlusions}
In~\cref{fig:reid:occ} we compare the clusters of our novel~\dataabr data to the embedding for occluded instances (cf.~\cref{sec:supp:dataset}).
In the first case we assume complete masks for the cow in question, which include RGB pixels from the occluding cow.
In the second case we assume occluded masks, without interfering texture information.
In both cases, the predicted codes fall in the correct cluster, showing that our method is robust against occlusions in the RGB and mask image.
%
%

%
%
%

\begin{figure*}[t]
    \vspace{0.35cm}
    \centering
    \begin{subfigure}[b]{\linewidth}
        \centering
        \includegraphics[width=0.24\textwidth,trim={2cm 4cm 5cm 2cm},clip]{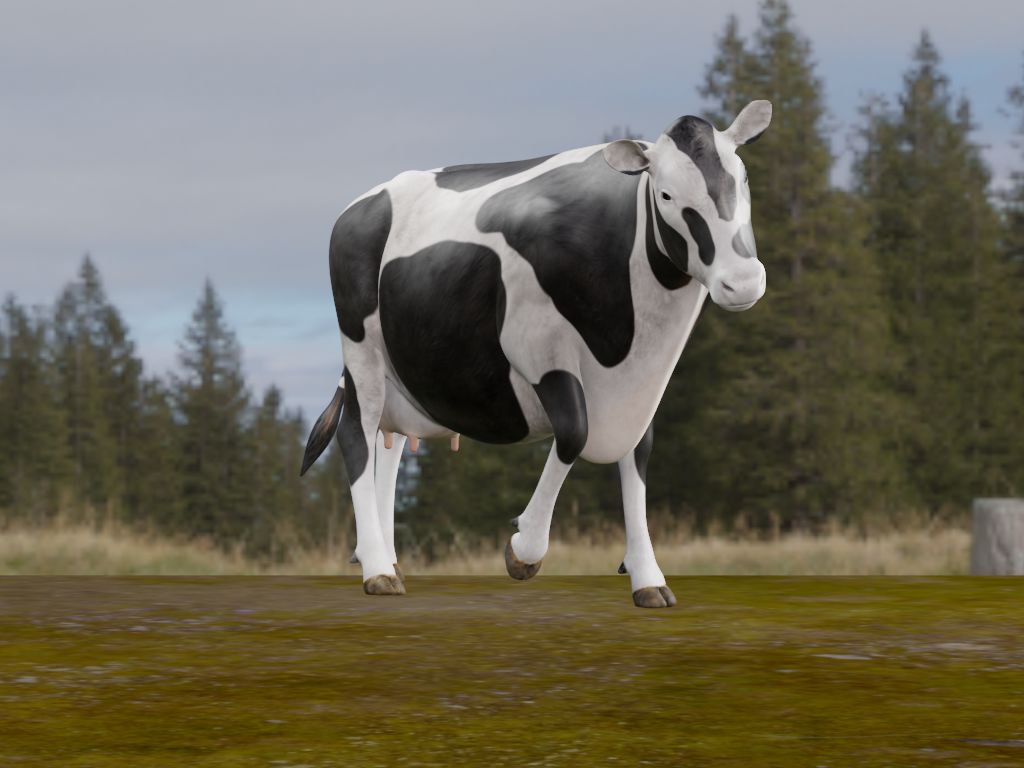}
        \includegraphics[width=0.24\textwidth,trim={2cm 4cm 5cm 2cm},clip]{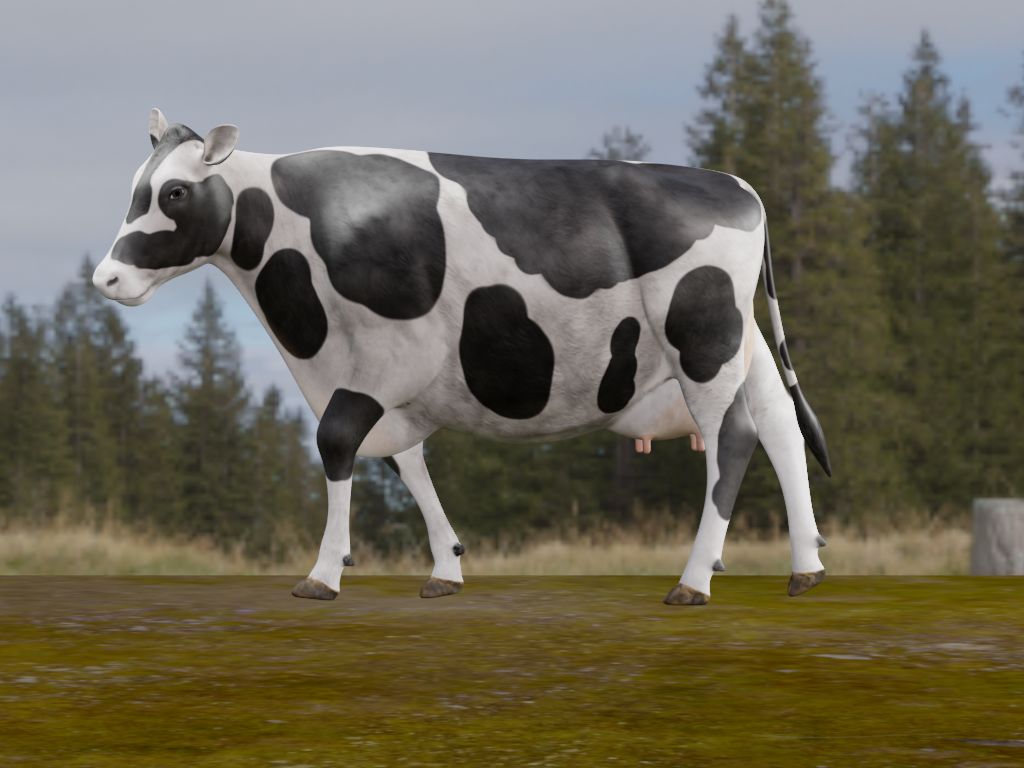}
        \includegraphics[width=0.24\textwidth,trim={2cm 4cm 5cm 2cm},clip]{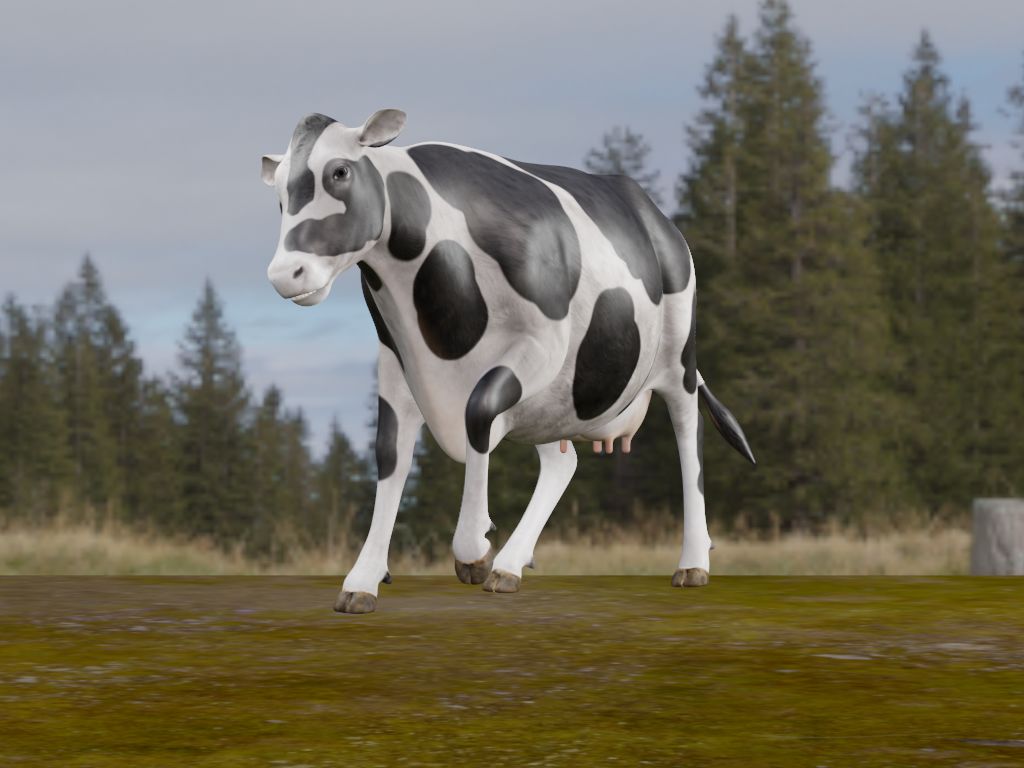}
        \includegraphics[width=0.24\textwidth,trim={2cm 4cm 5cm 2cm},clip]{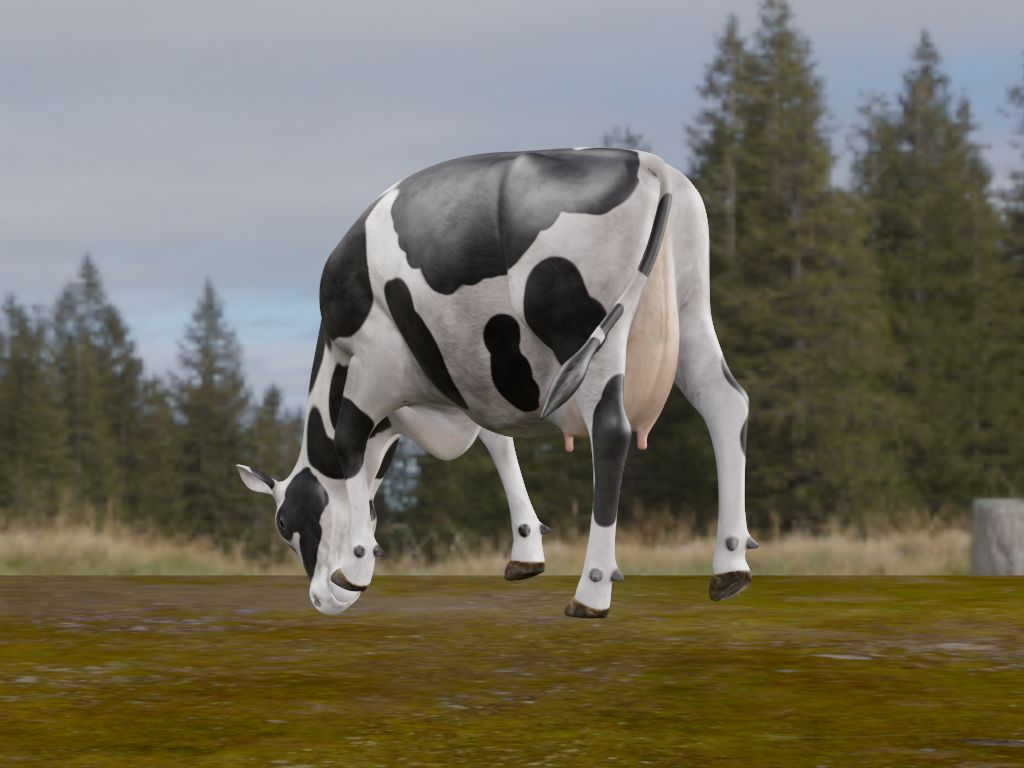}
        \caption{Texture $0$}
    \end{subfigure}
    \hfill
    \begin{subfigure}[b]{\linewidth}
        \centering
        \includegraphics[width=0.24\textwidth,trim={2cm 4cm 5cm 2cm},clip]{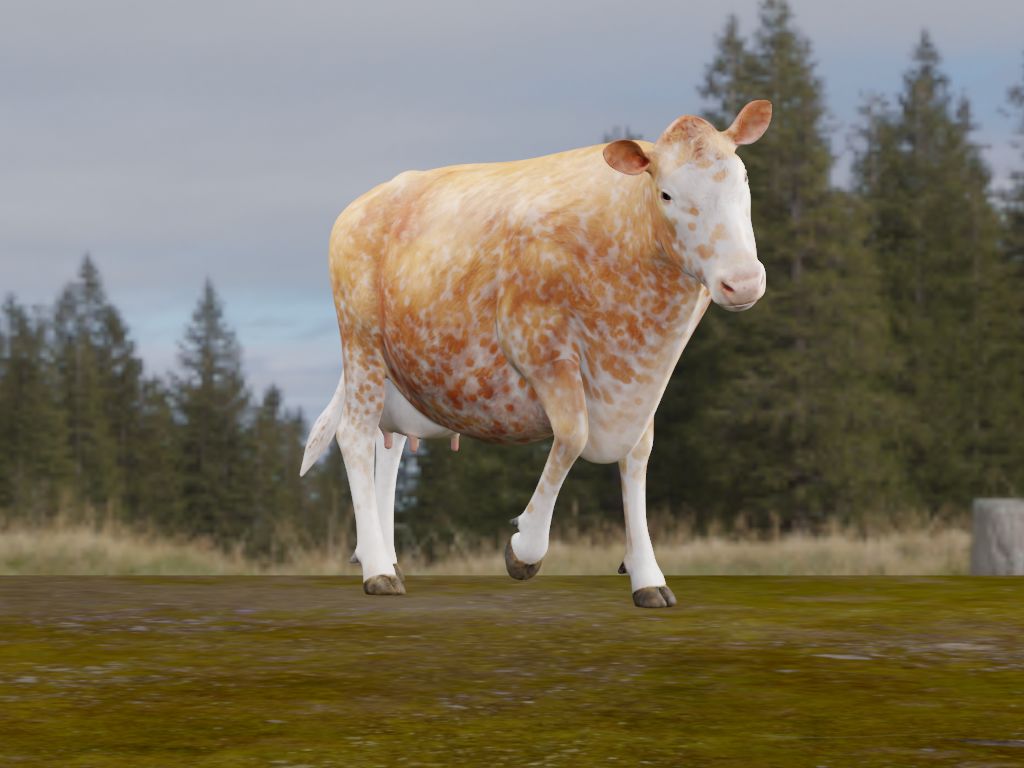}
        \includegraphics[width=0.24\textwidth,trim={2cm 4cm 5cm 2cm},clip]{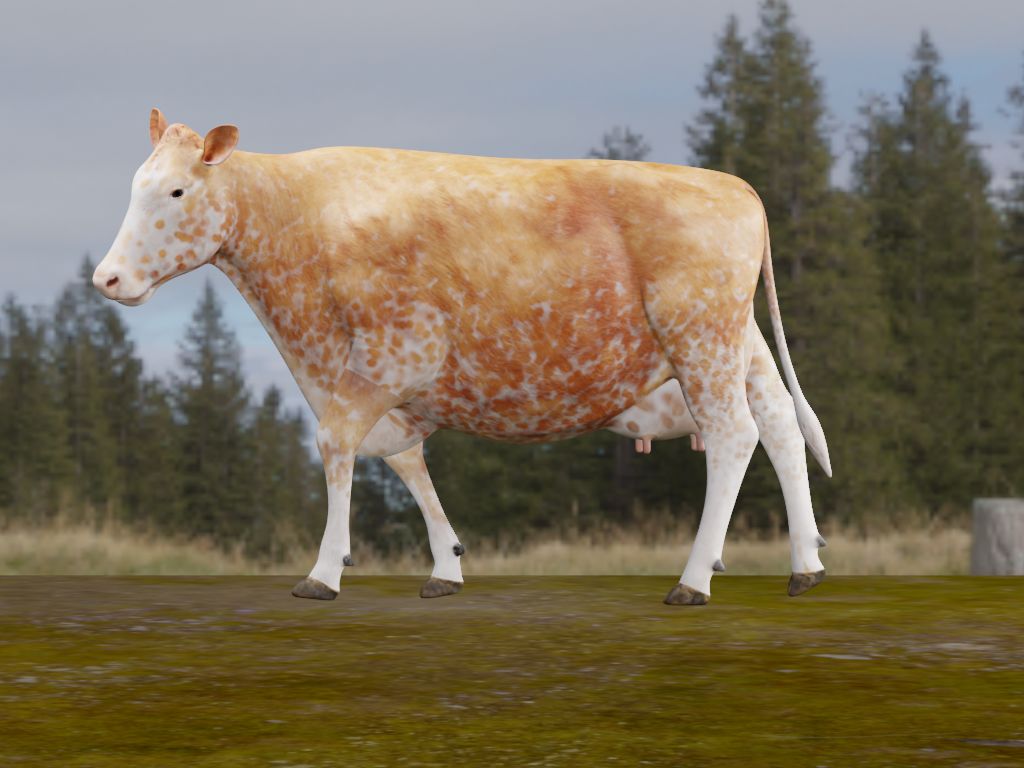}
        \includegraphics[width=0.24\textwidth,trim={2cm 4cm 5cm 2cm},clip]{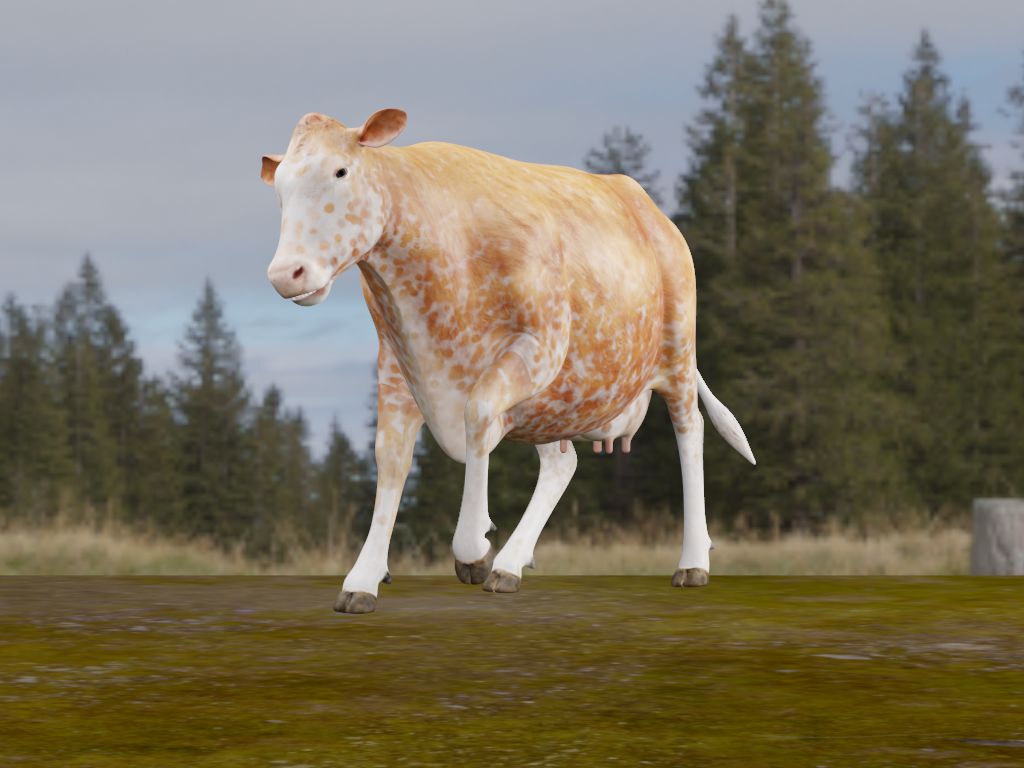}
        \includegraphics[width=0.24\textwidth,trim={2cm 4cm 5cm 2cm},clip]{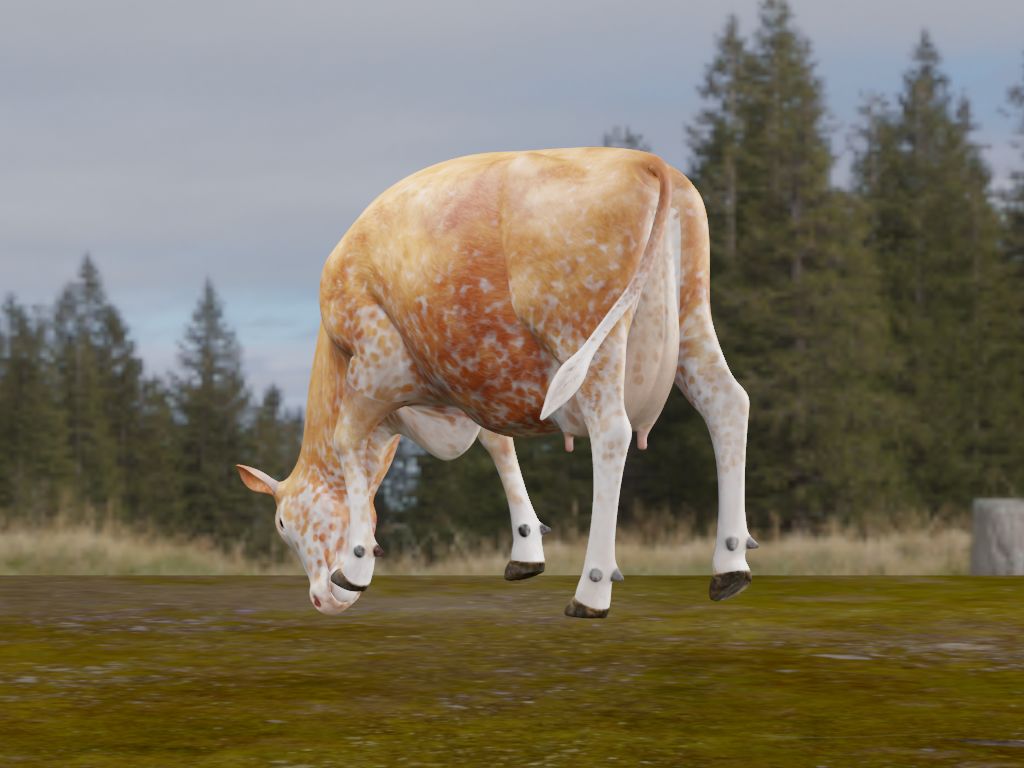}
        \caption{Texture $1$}
        \label{fig:supp:dataset:1}
    \end{subfigure}
    \hfill
    \begin{subfigure}[b]{\linewidth}
        \centering
        \includegraphics[width=0.24\textwidth,trim={2cm 4cm 5cm 2cm},clip]{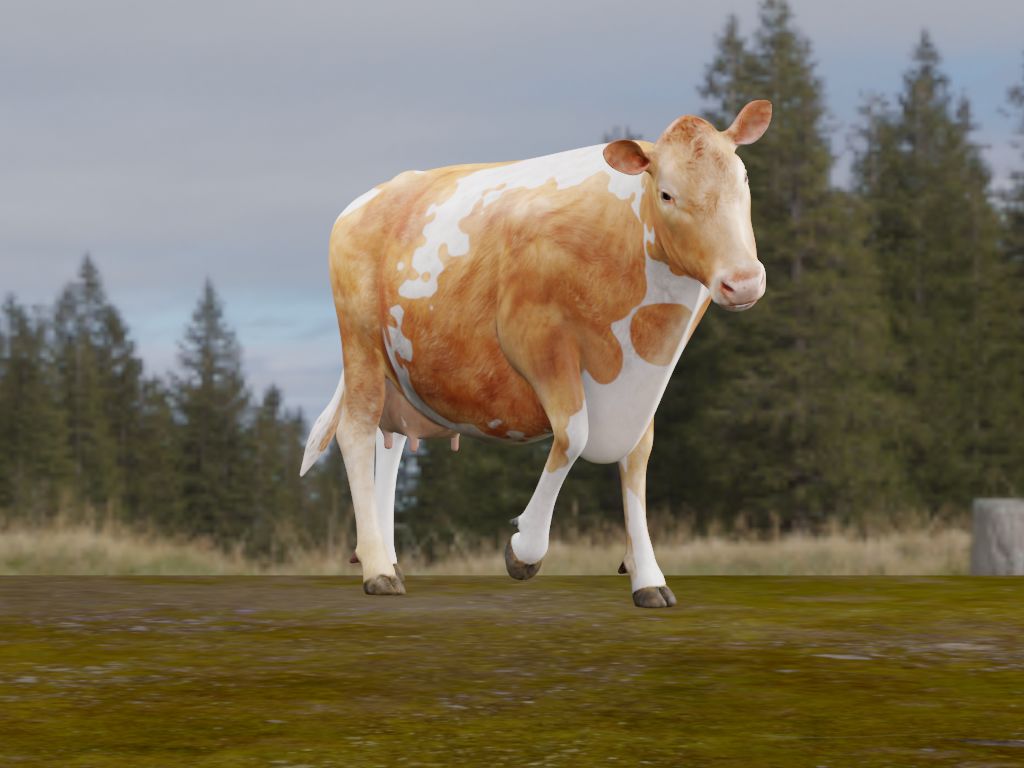}
        \includegraphics[width=0.24\textwidth,trim={2cm 4cm 5cm 2cm},clip]{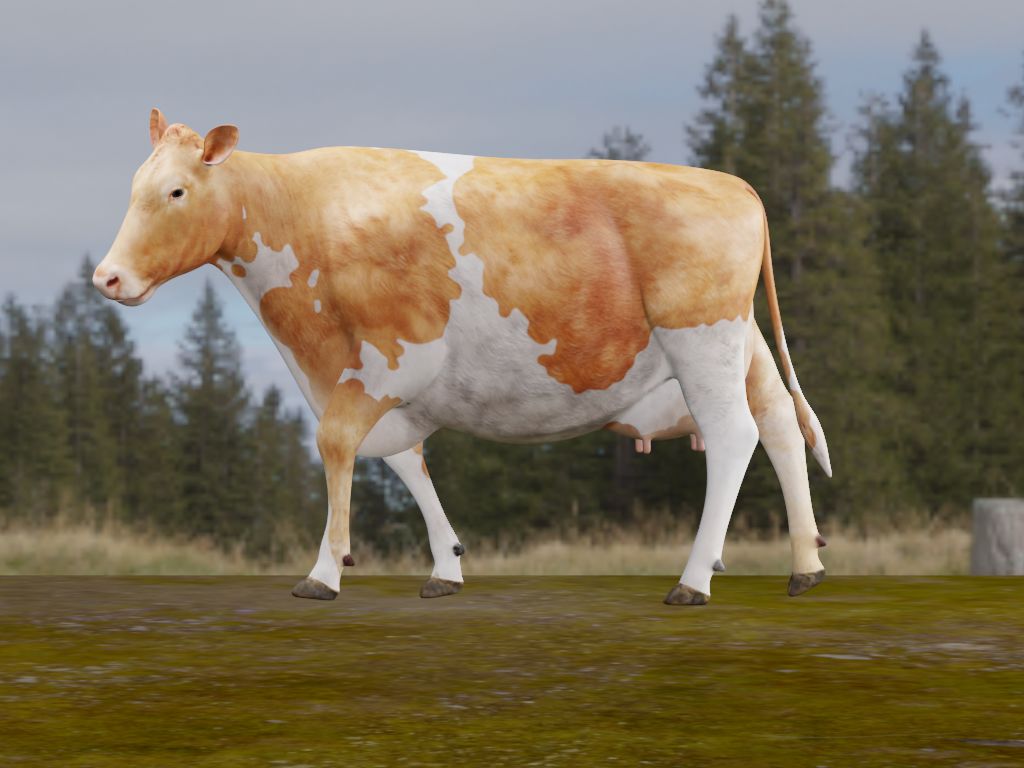}
        \includegraphics[width=0.24\textwidth,trim={2cm 4cm 5cm 2cm},clip]{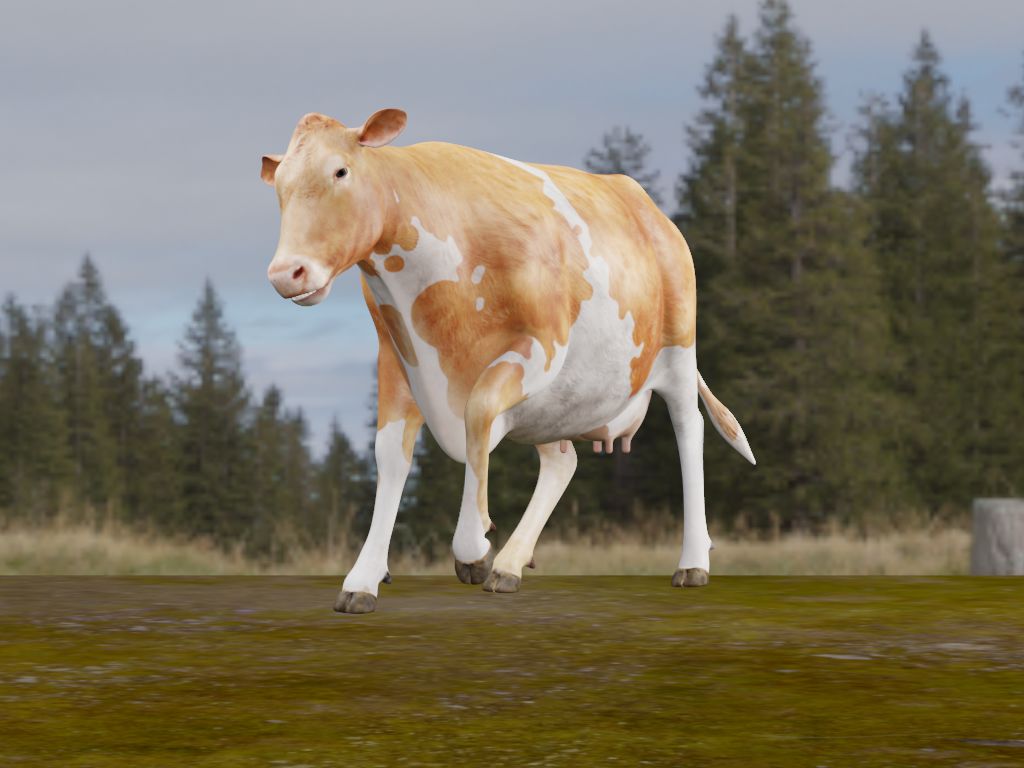}
        \includegraphics[width=0.24\textwidth,trim={2cm 4cm 5cm 2cm},clip]{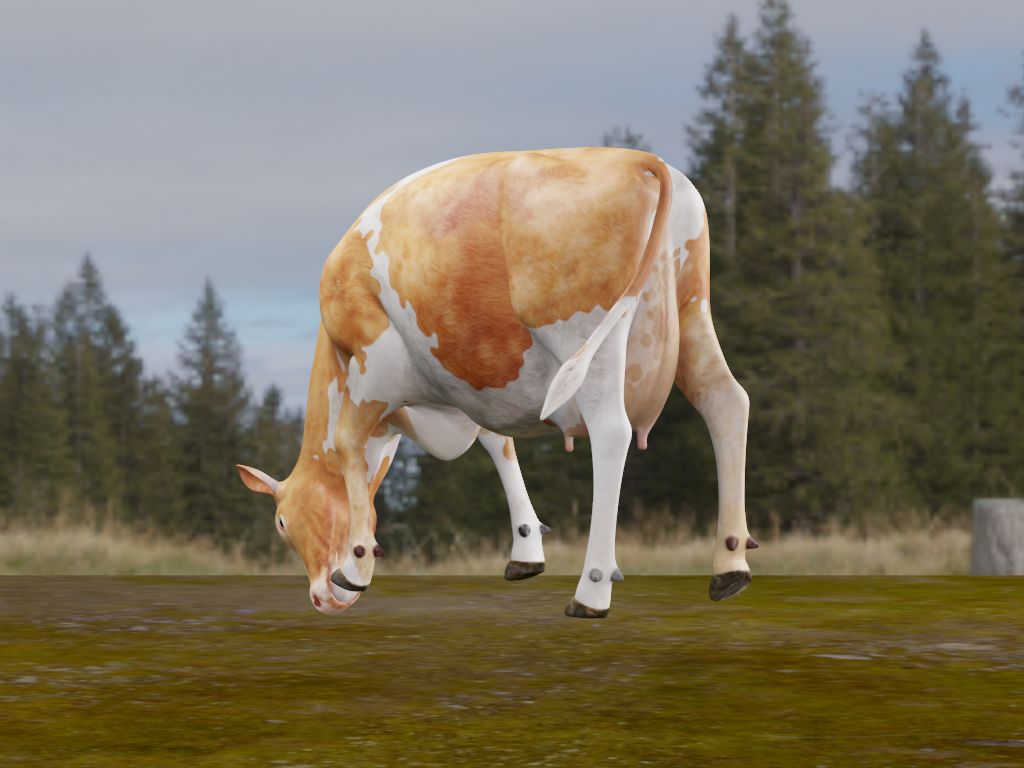}
        \caption{Texture $2$}
    \end{subfigure}
    \hfill
    \begin{subfigure}[b]{\linewidth}
        \centering
        \includegraphics[width=0.24\textwidth,trim={2cm 4cm 5cm 2cm},clip]{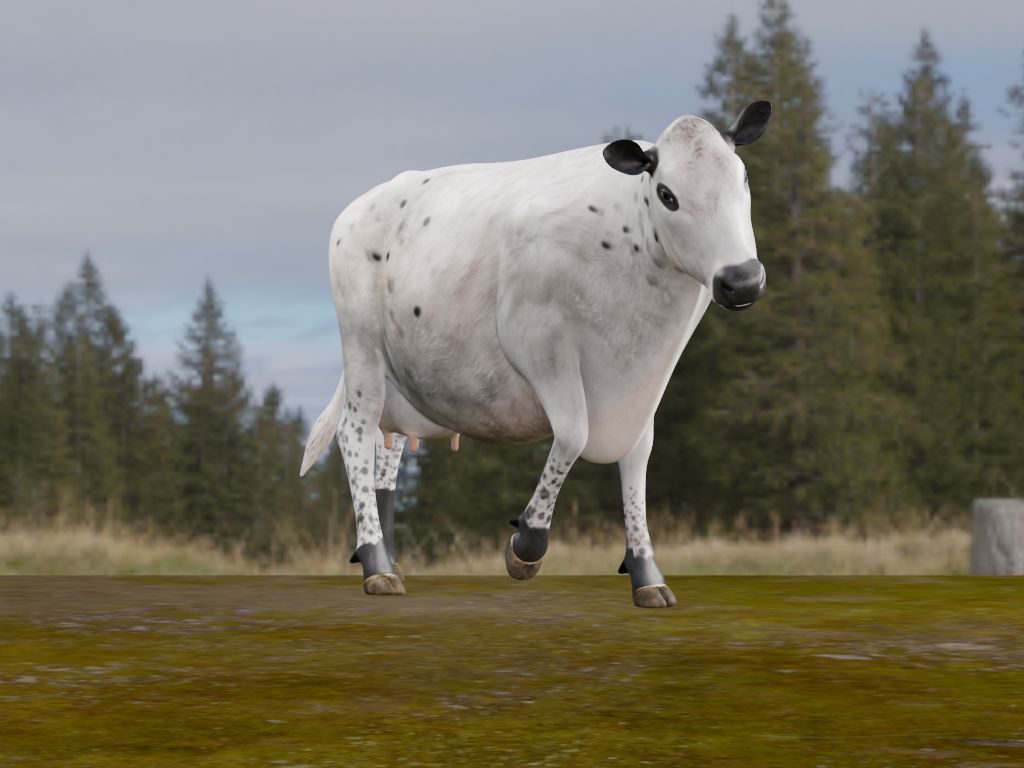}
        \includegraphics[width=0.24\textwidth,trim={2cm 4cm 5cm 2cm},clip]{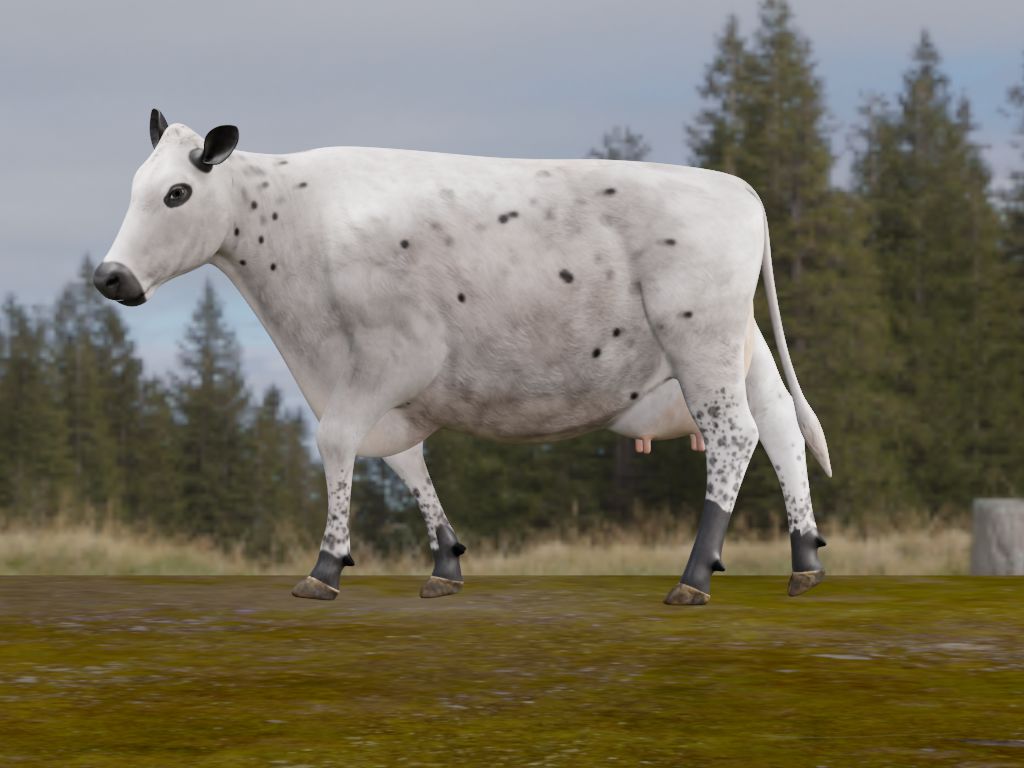}
        \includegraphics[width=0.24\textwidth,trim={2cm 4cm 5cm 2cm},clip]{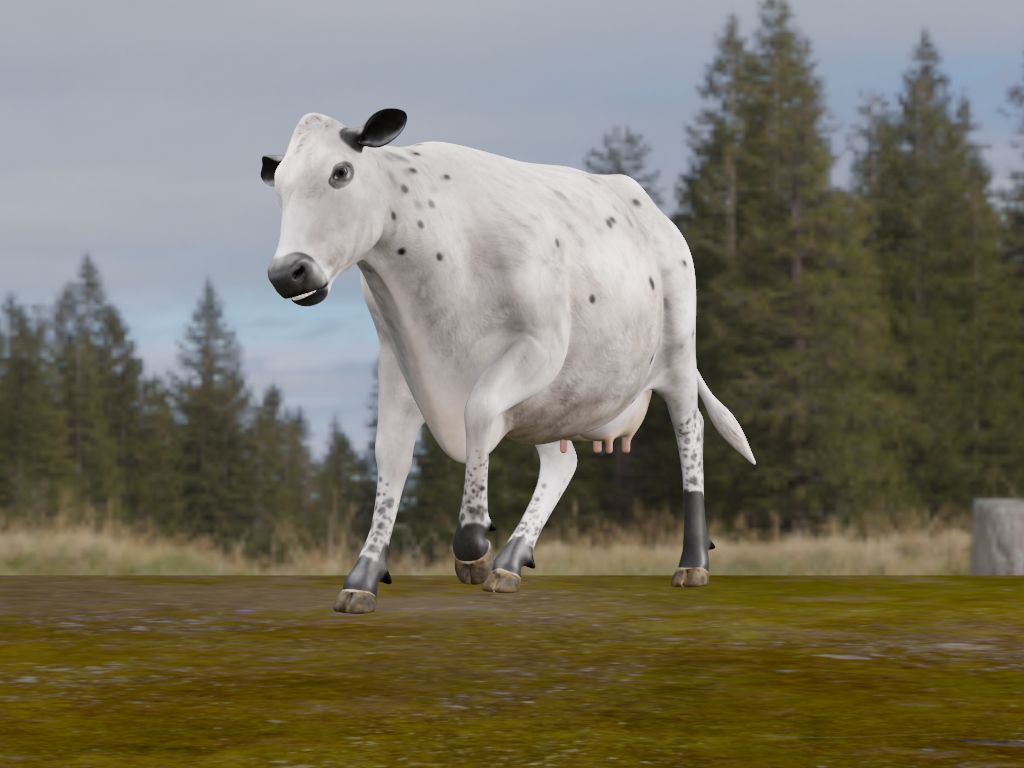}
        \includegraphics[width=0.24\textwidth,trim={2cm 4cm 5cm 2cm},clip]{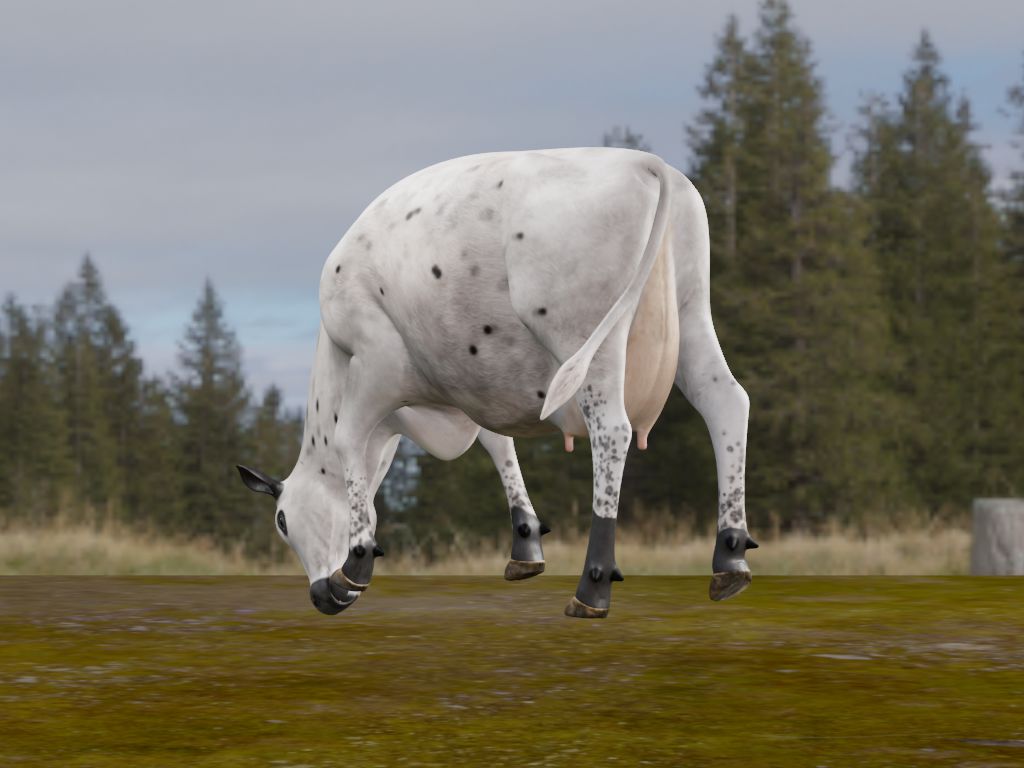}
        \caption{Texture $3$}
    \end{subfigure}
    \hfill
    \begin{subfigure}[b]{\linewidth}
        \centering
        \includegraphics[width=0.24\textwidth,trim={2cm 4cm 5cm 2cm},clip]{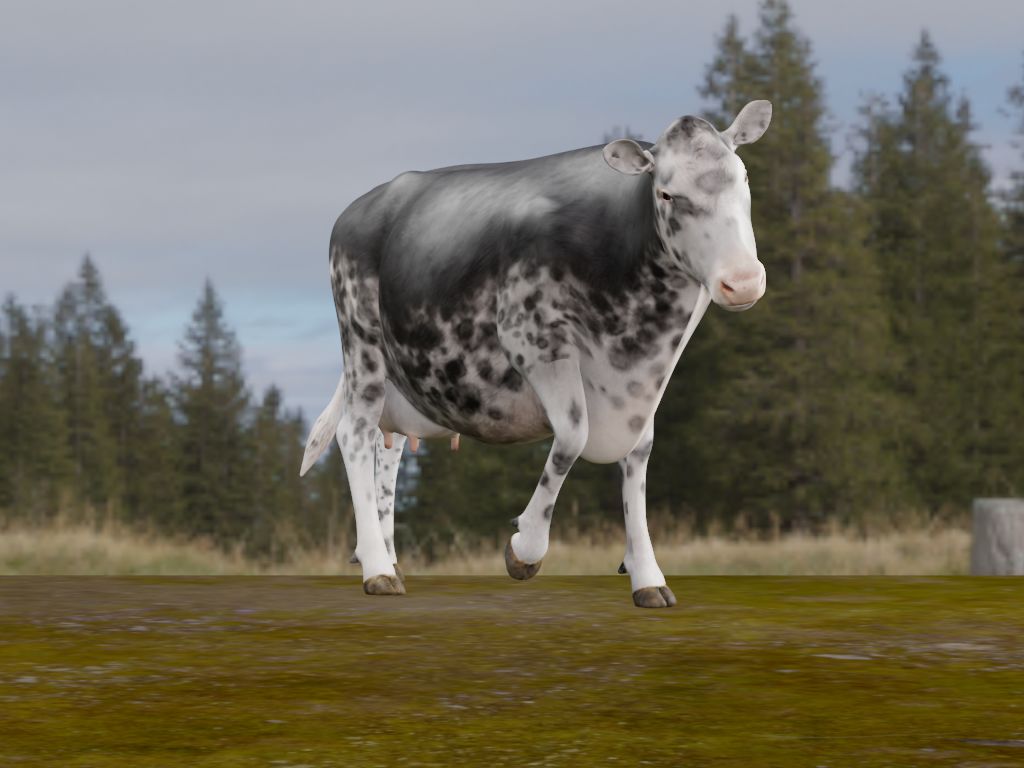}
        \includegraphics[width=0.24\textwidth,trim={2cm 4cm 5cm 2cm},clip]{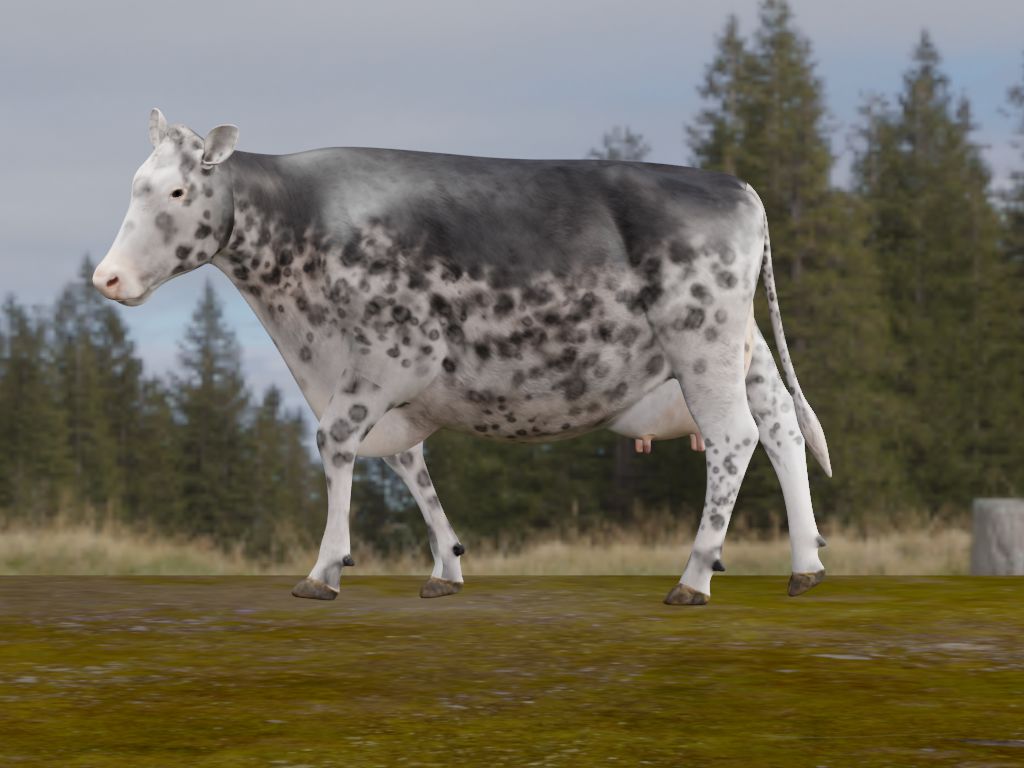}
        \includegraphics[width=0.24\textwidth,trim={2cm 4cm 5cm 2cm},clip]{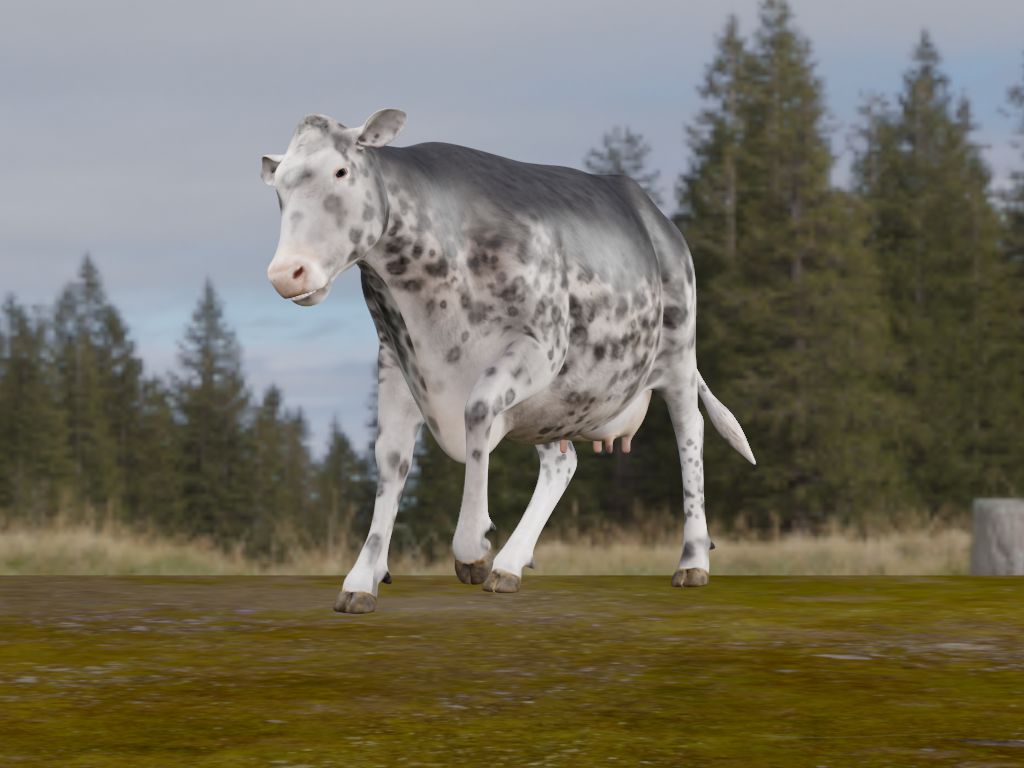}
        \includegraphics[width=0.24\textwidth,trim={2cm 4cm 5cm 2cm},clip]{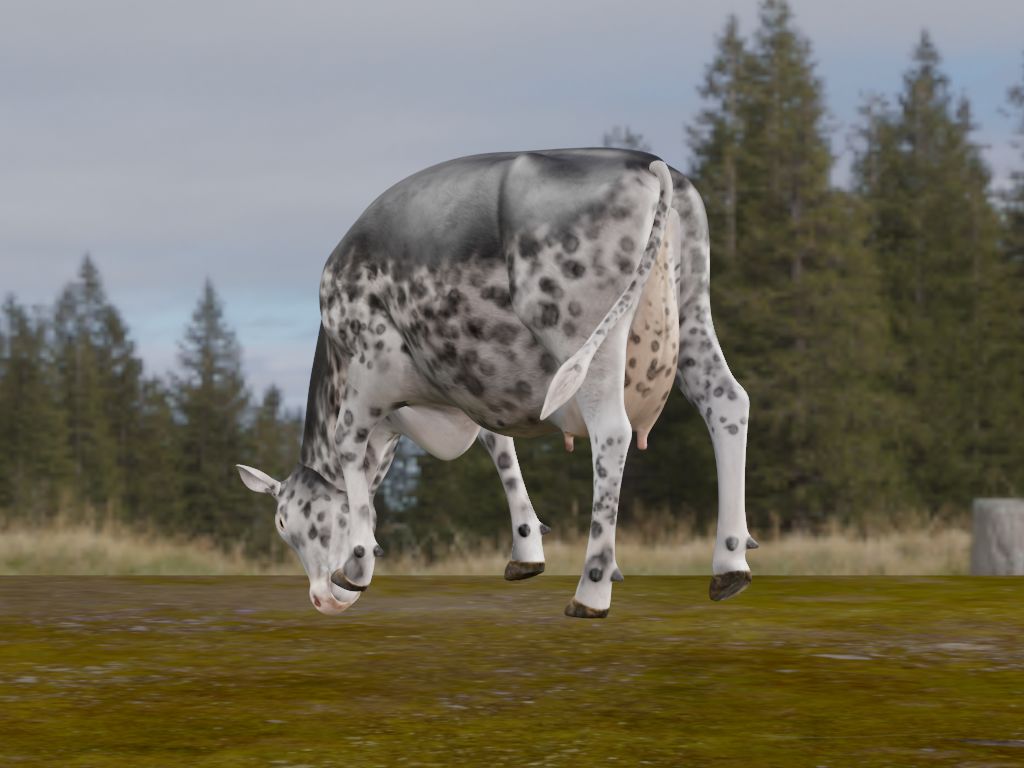}
        \caption{Texture $4$}
    \end{subfigure}
    \hfill
    \begin{subfigure}[b]{\linewidth}
        \centering
        \includegraphics[width=0.24\textwidth,trim={2cm 4cm 5cm 2cm},clip]{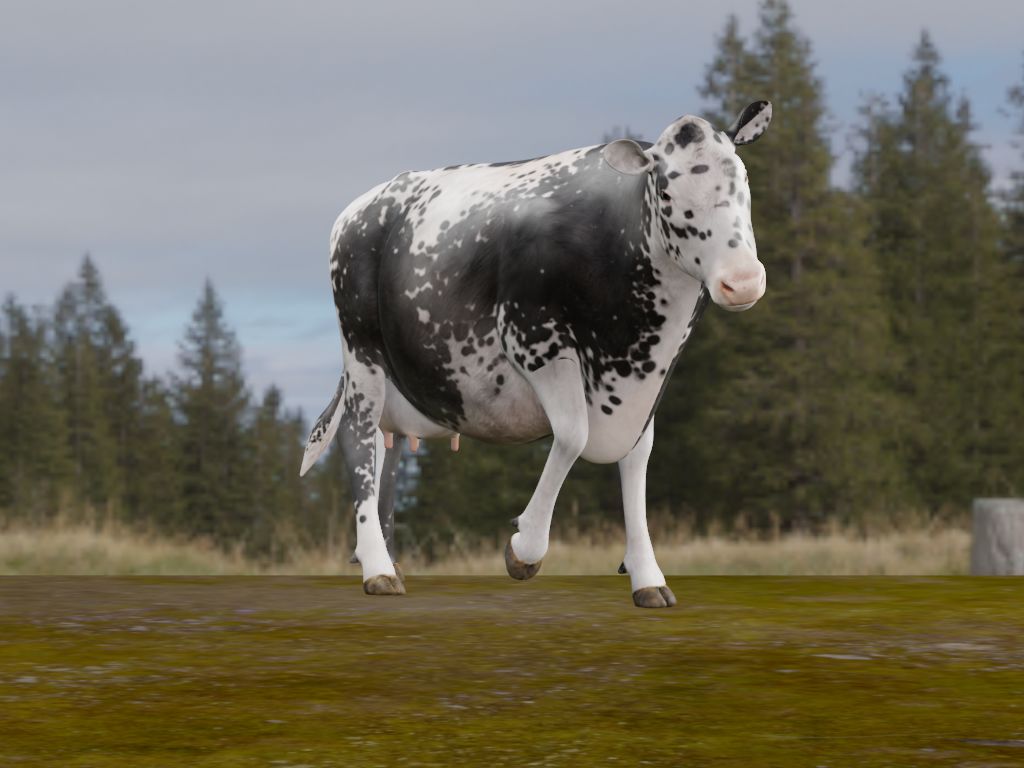}
        \includegraphics[width=0.24\textwidth,trim={2cm 4cm 5cm 2cm},clip]{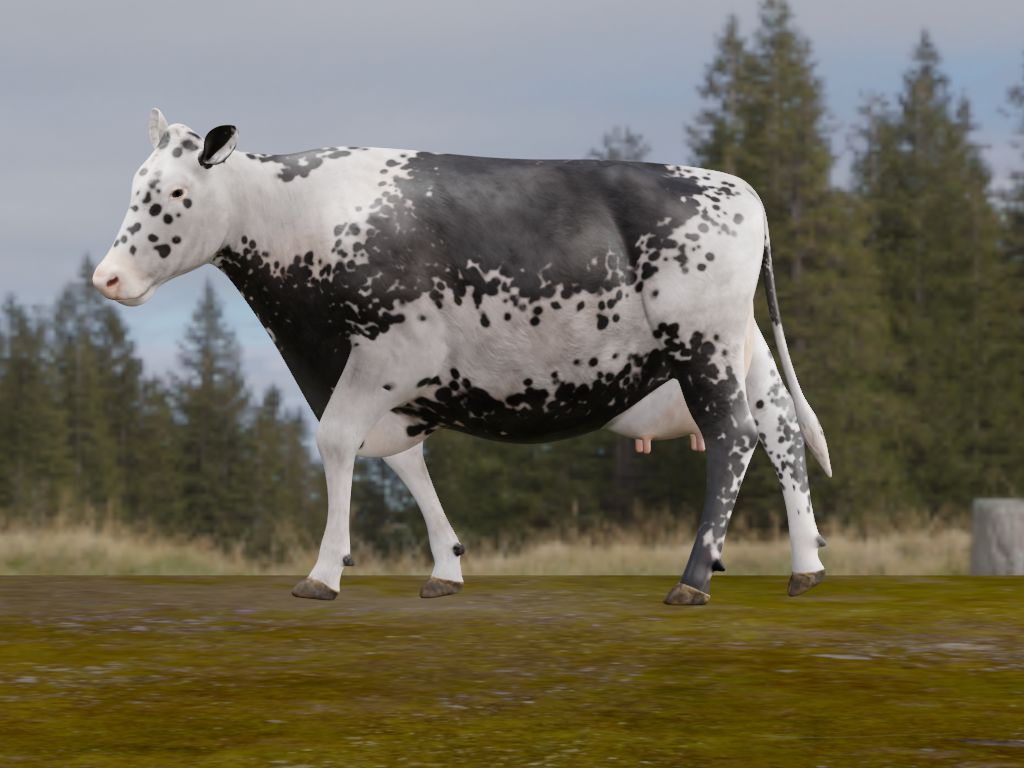}
        \includegraphics[width=0.24\textwidth,trim={2cm 4cm 5cm 2cm},clip]{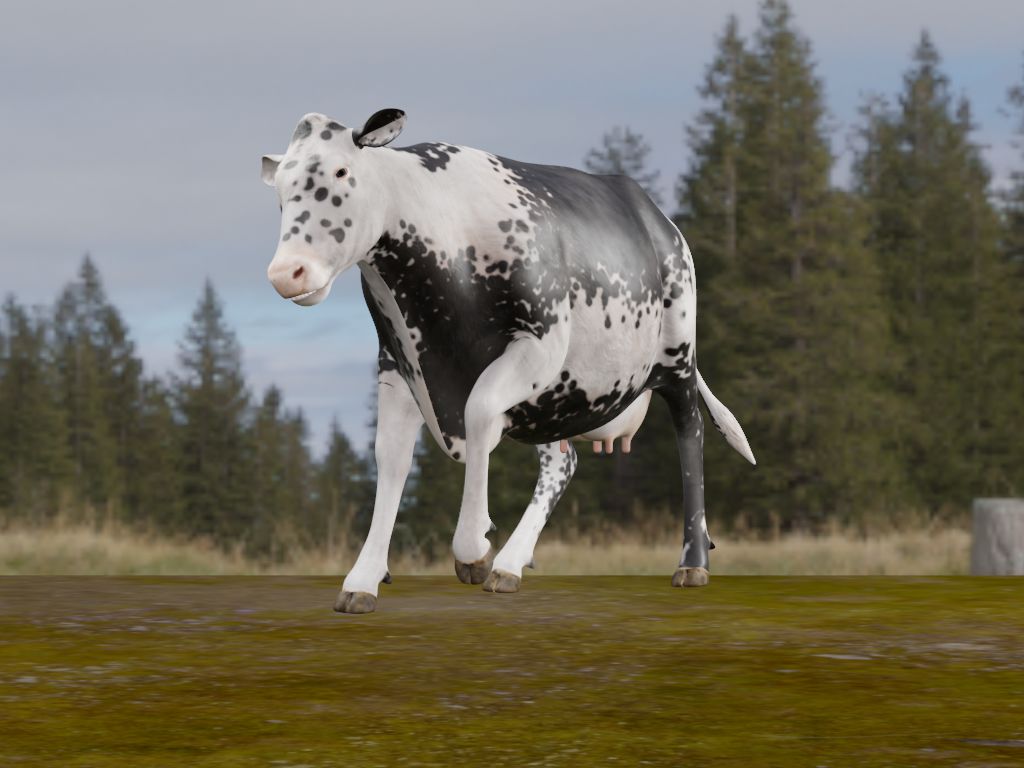}
        \includegraphics[width=0.24\textwidth,trim={2cm 4cm 5cm 2cm},clip]{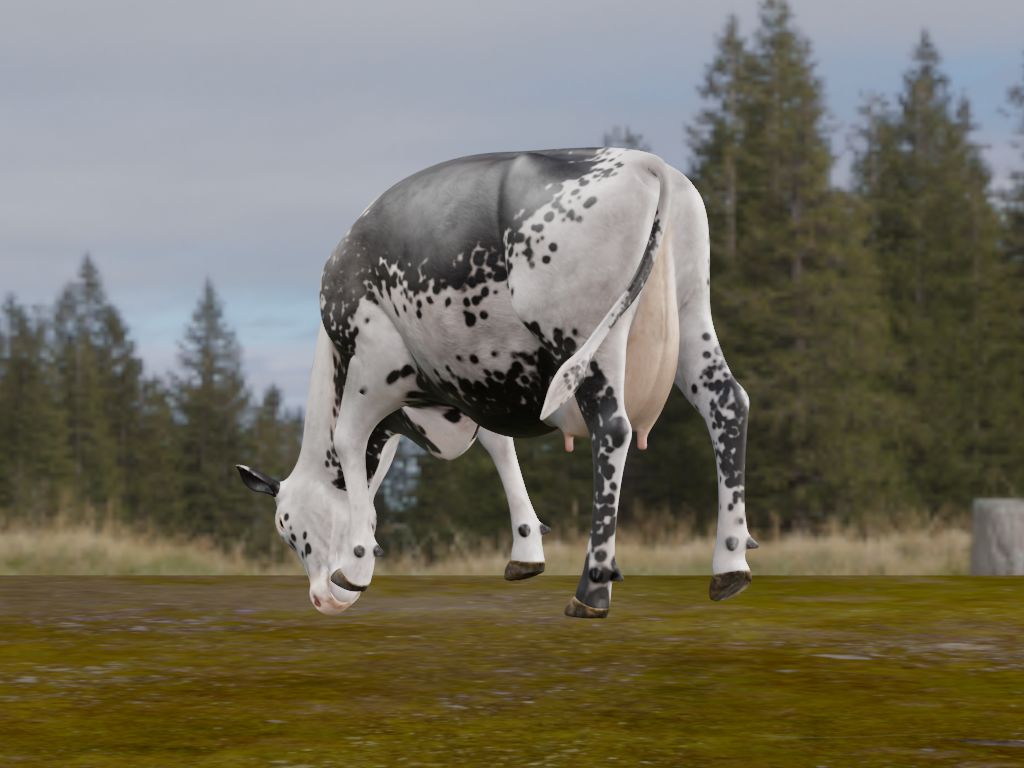}
        \caption{Texture $5$}
    \end{subfigure}
    \caption{
        \textbf{Dataset.}
        We show four examples for each distinct texture in our novel~\dataabr dataset.
    }
\end{figure*}
\begin{figure*}[t]\ContinuedFloat
    \vspace{0.35cm}
    \centering
    \begin{subfigure}[b]{\linewidth}
        \centering
        \includegraphics[width=0.24\textwidth,trim={2cm 4cm 5cm 2cm},clip]{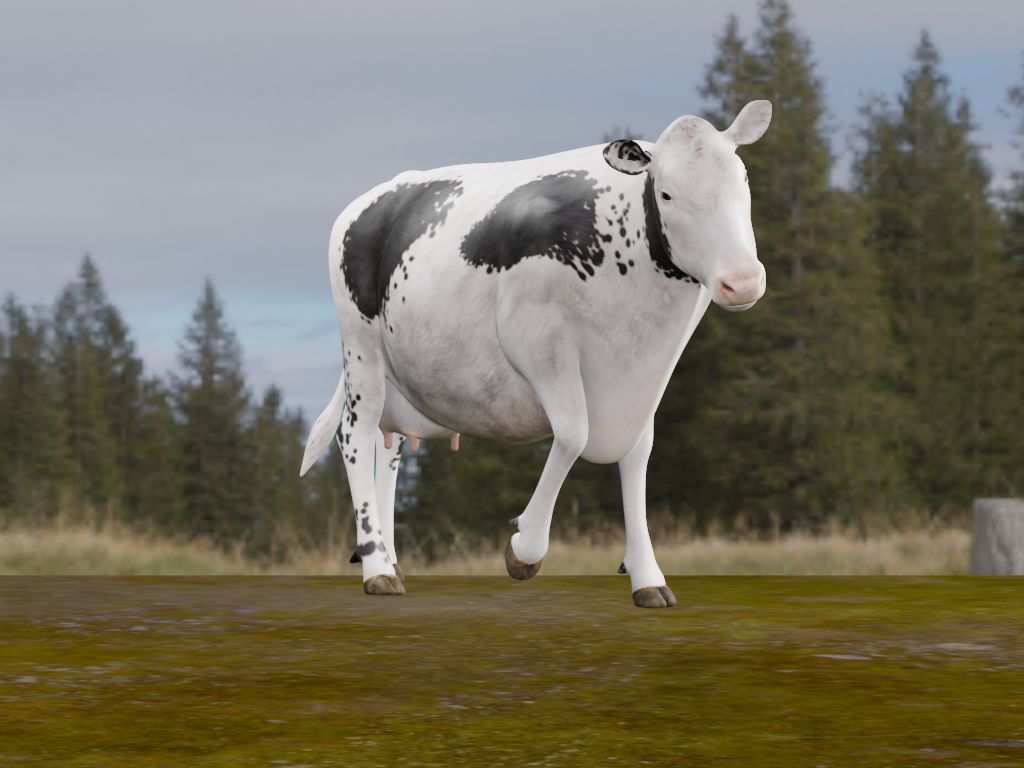}
        \includegraphics[width=0.24\textwidth,trim={2cm 4cm 5cm 2cm},clip]{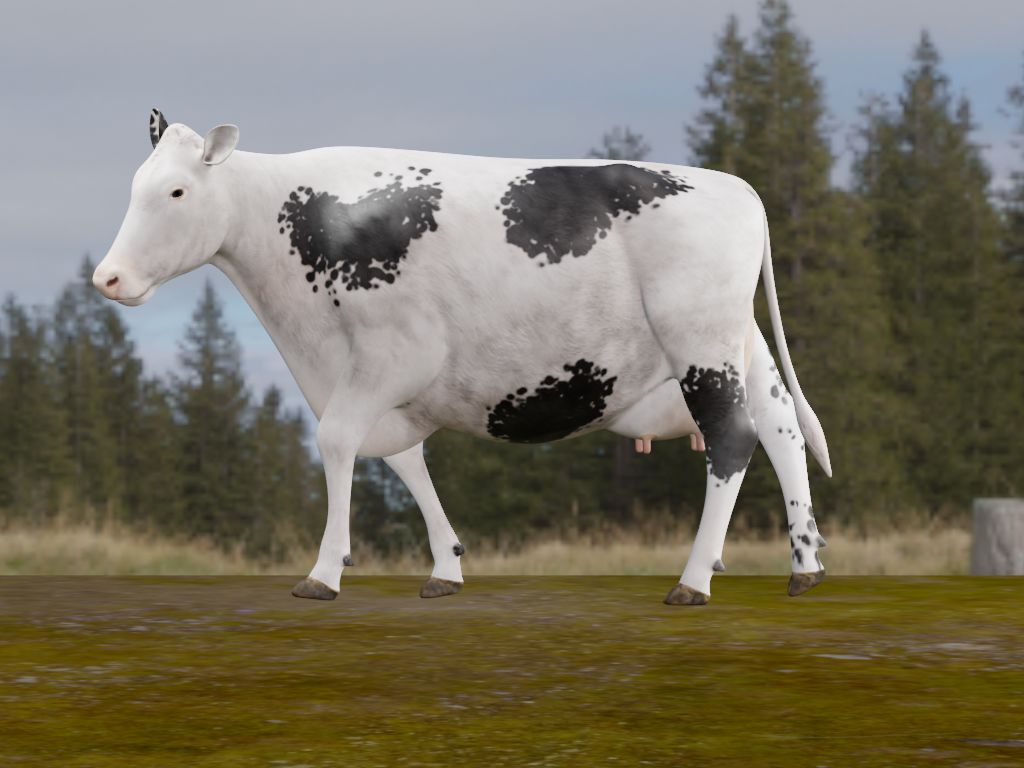}
        \includegraphics[width=0.24\textwidth,trim={2cm 4cm 5cm 2cm},clip]{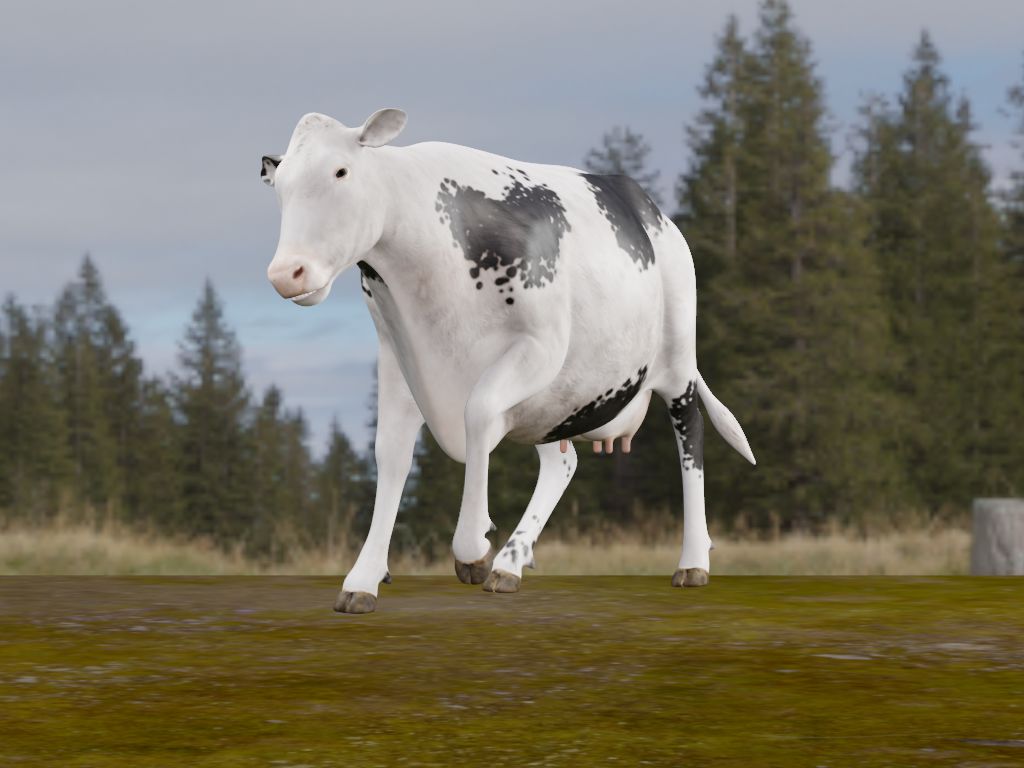}
        \includegraphics[width=0.24\textwidth,trim={2cm 4cm 5cm 2cm},clip]{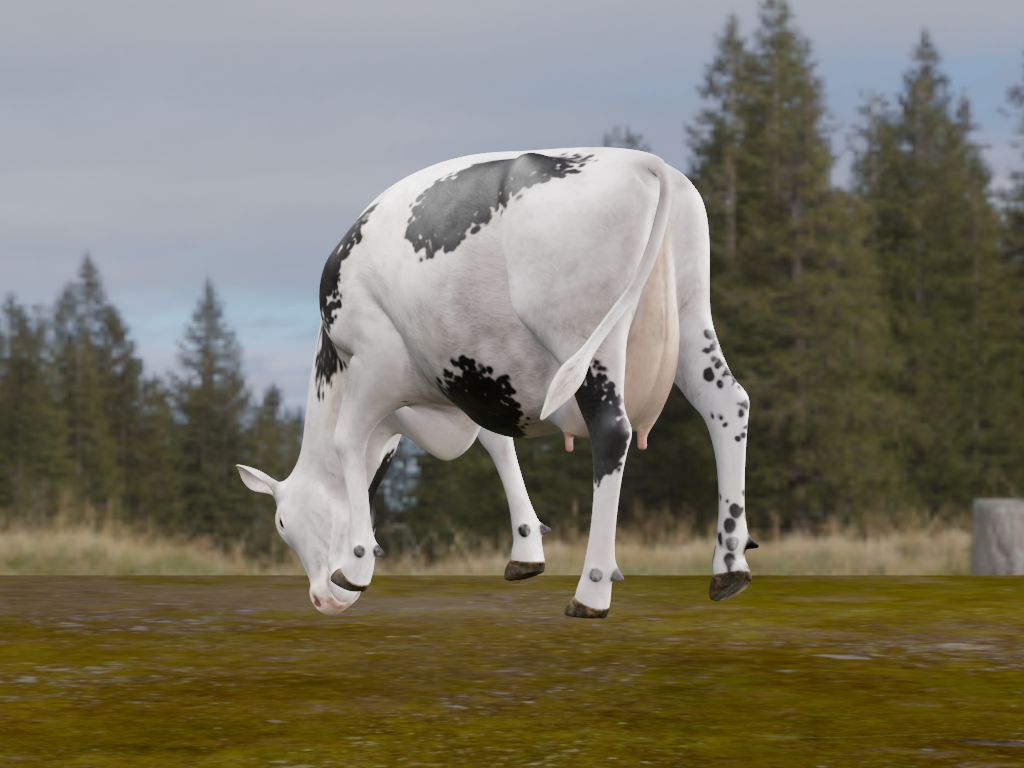}
        \caption{Texture $6$}
    \end{subfigure}
    \hfill
        \begin{subfigure}[b]{\linewidth}
        \centering
        \includegraphics[width=0.24\textwidth,trim={2cm 4cm 5cm 2cm},clip]{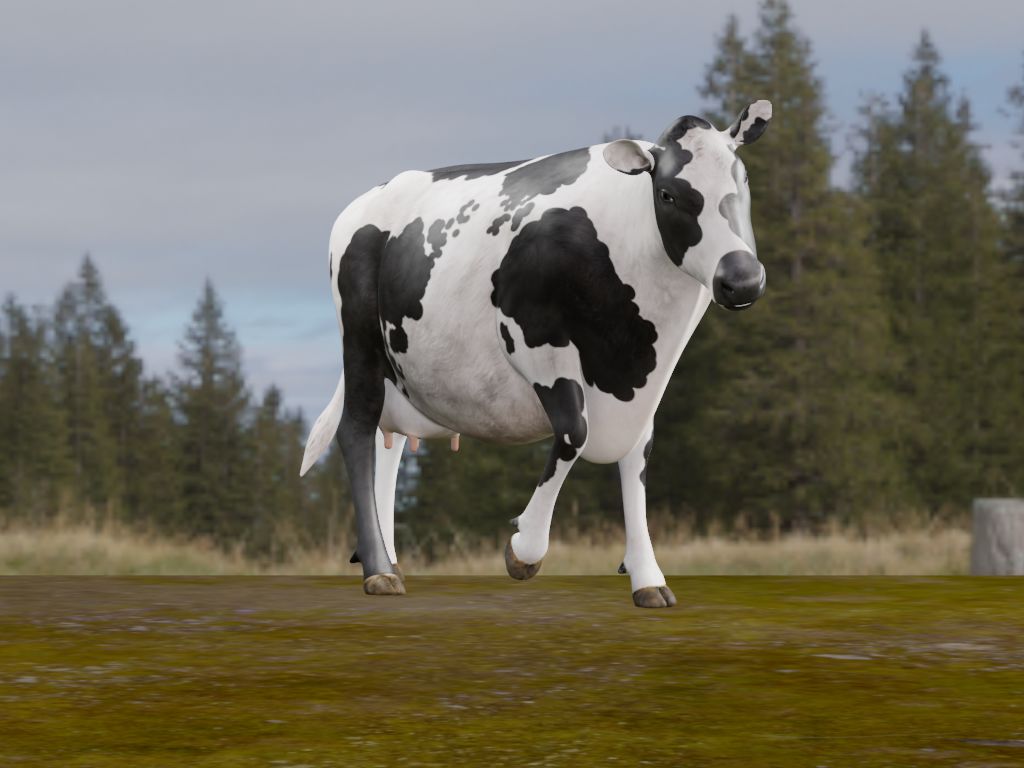}
        \includegraphics[width=0.24\textwidth,trim={2cm 4cm 5cm 2cm},clip]{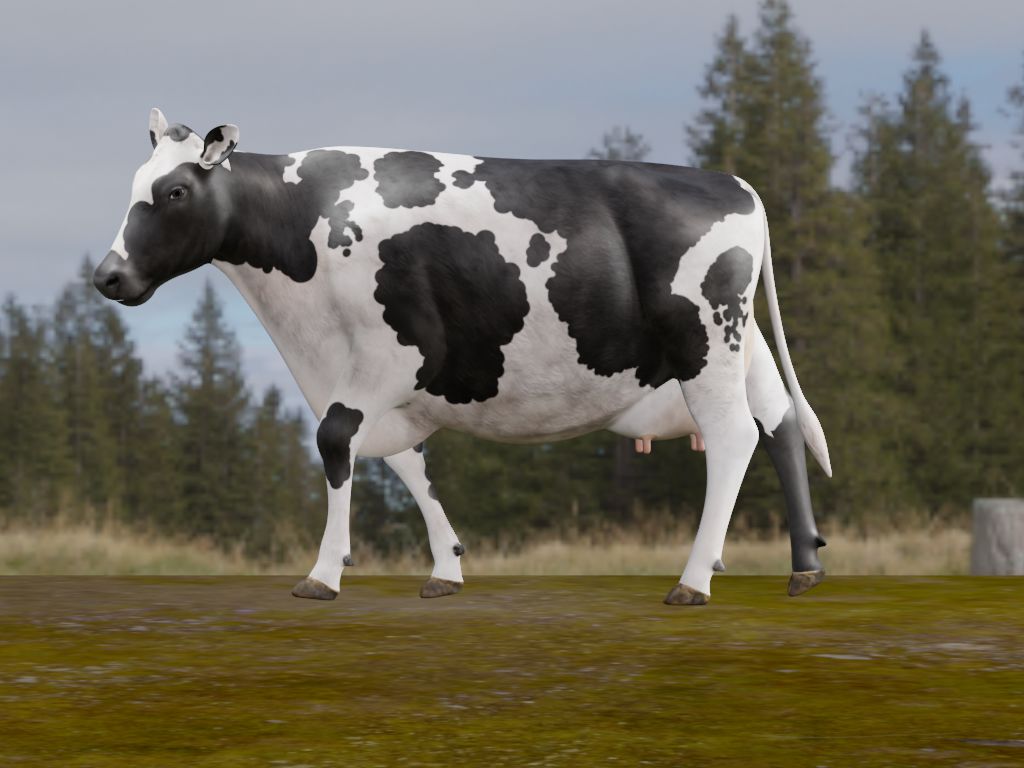}
        \includegraphics[width=0.24\textwidth,trim={2cm 4cm 5cm 2cm},clip]{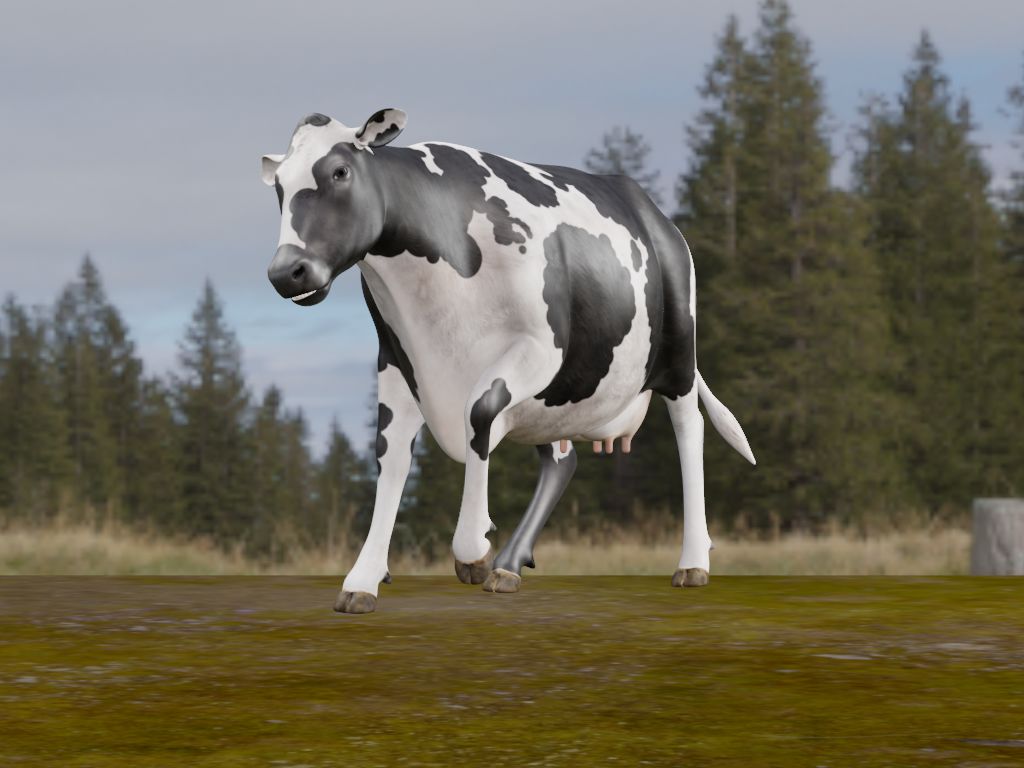}
        \includegraphics[width=0.24\textwidth,trim={2cm 4cm 5cm 2cm},clip]{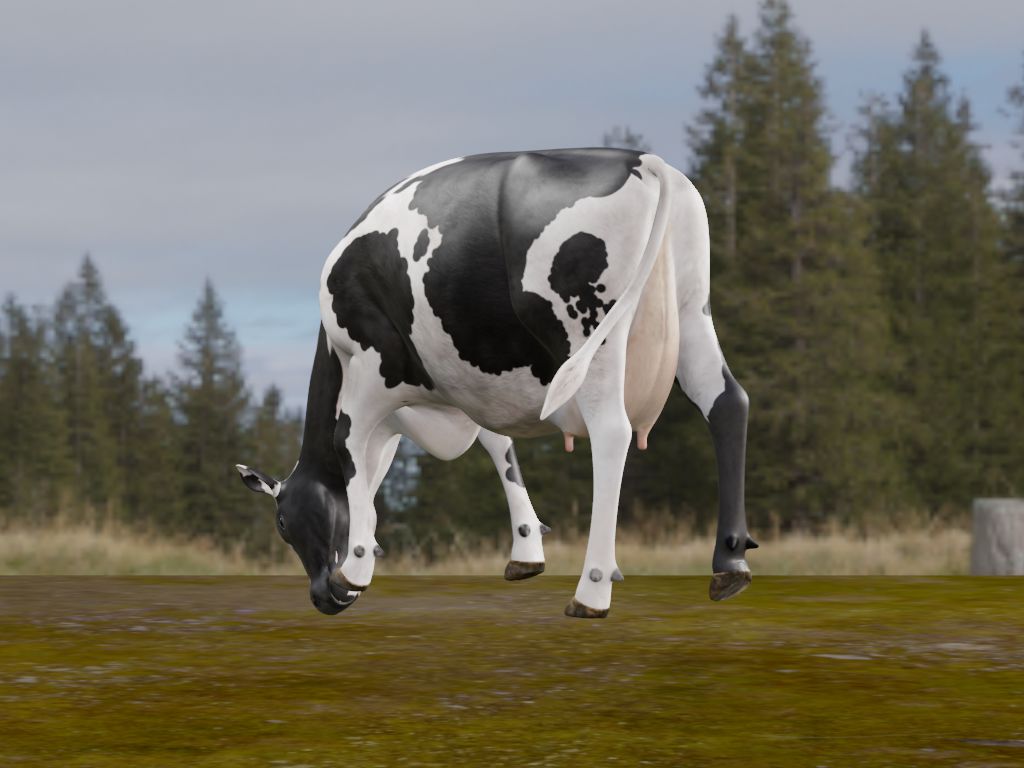}
        \caption{Texture $7$}
    \end{subfigure}
    \hfill
        \begin{subfigure}[b]{\linewidth}
        \centering
        \includegraphics[width=0.24\textwidth,trim={2cm 4cm 5cm 2cm},clip]{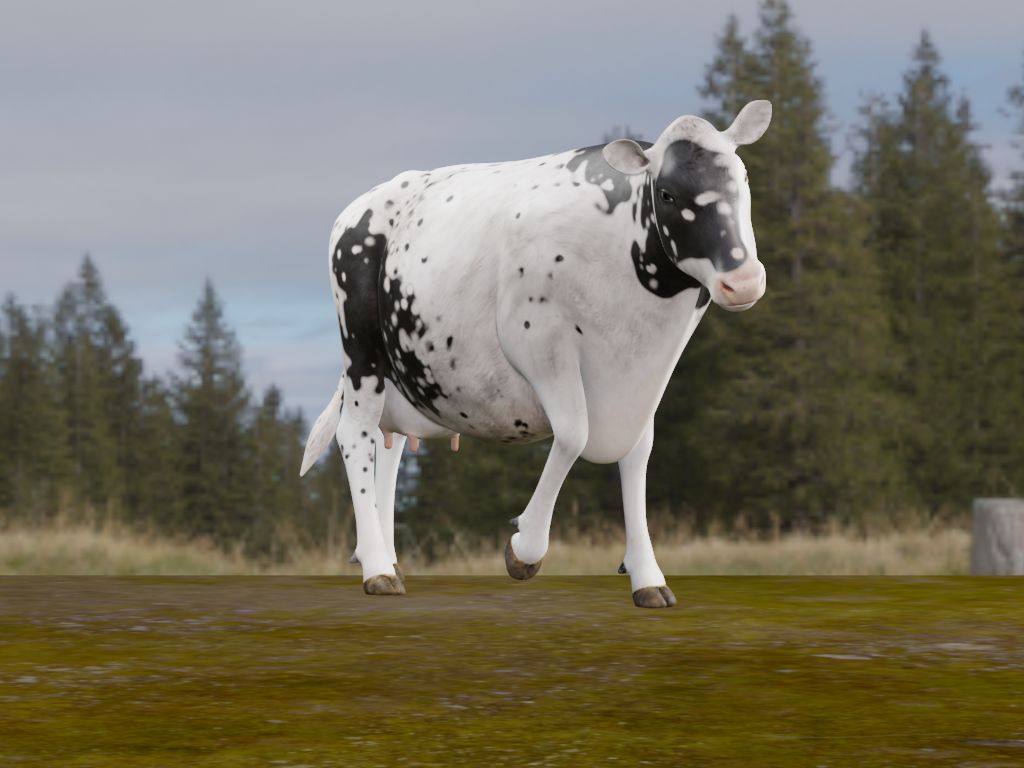}
        \includegraphics[width=0.24\textwidth,trim={2cm 4cm 5cm 2cm},clip]{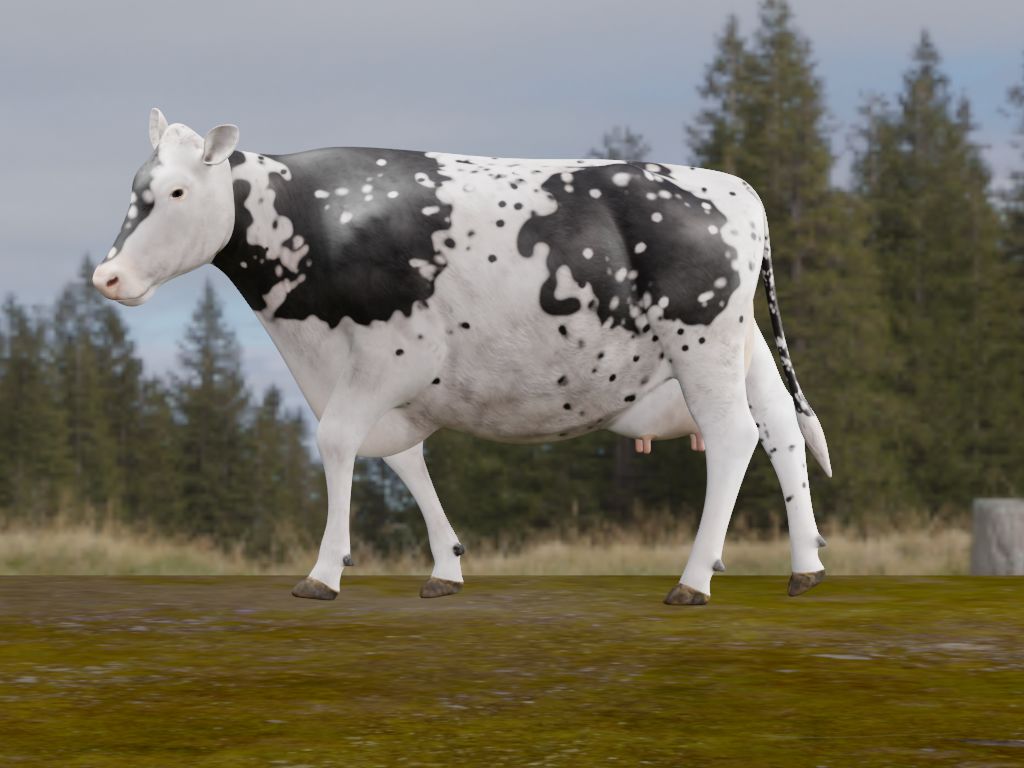}
        \includegraphics[width=0.24\textwidth,trim={2cm 4cm 5cm 2cm},clip]{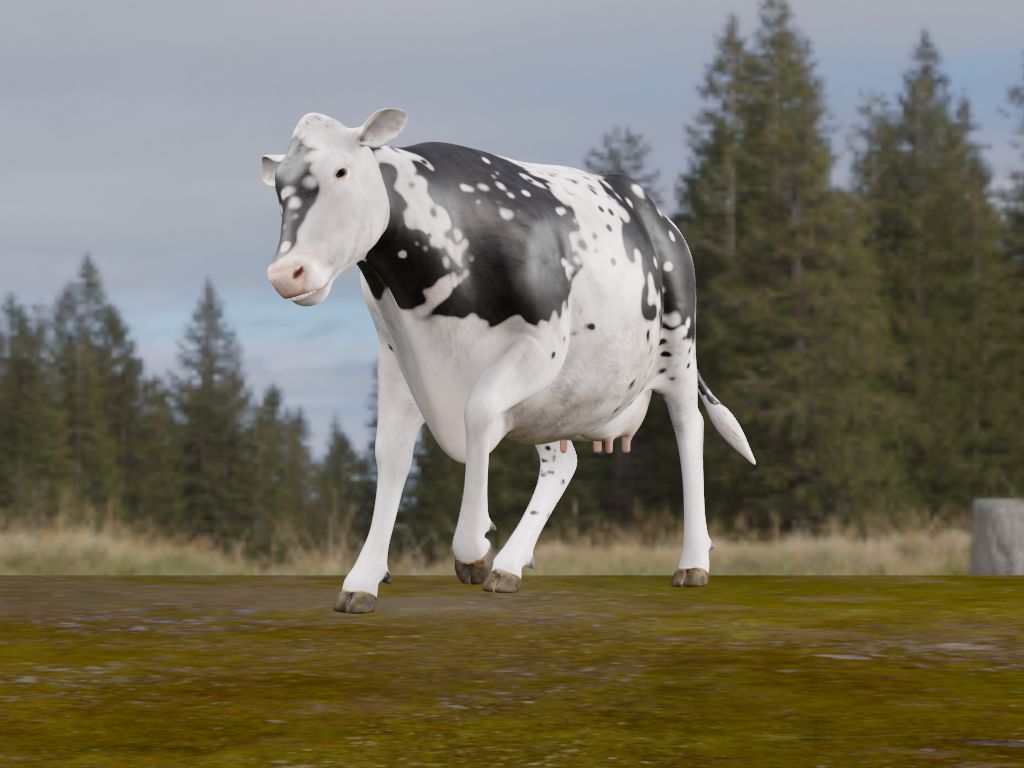}
        \includegraphics[width=0.24\textwidth,trim={2cm 4cm 5cm 2cm},clip]{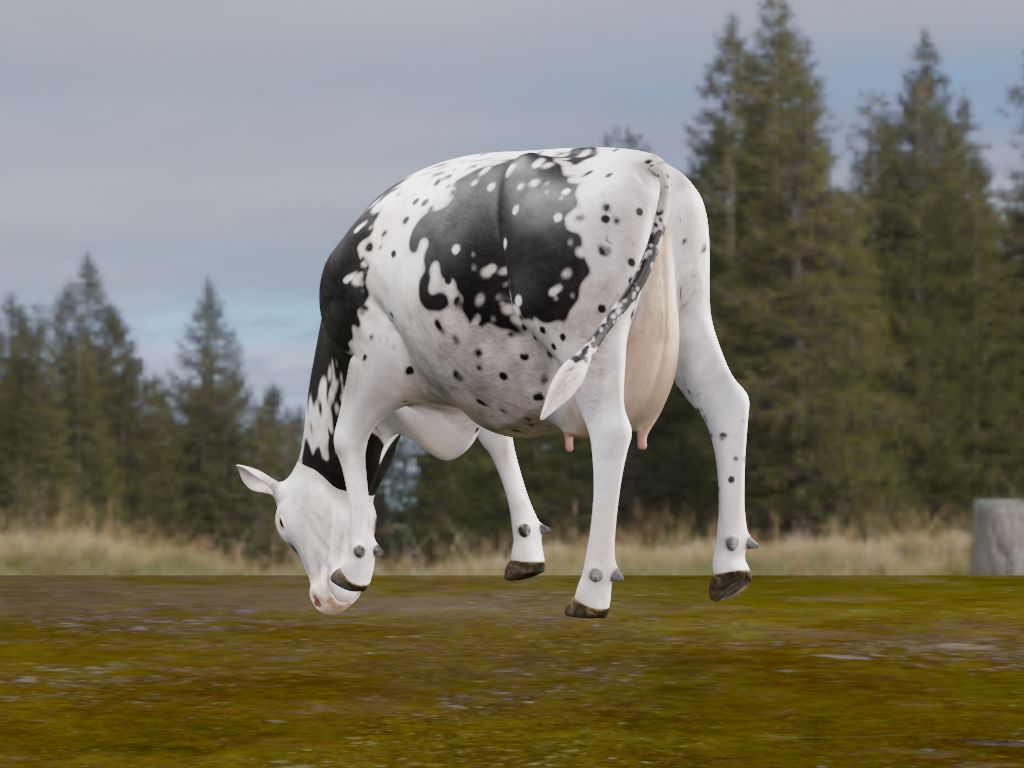}
        \caption{Texture $8$}
    \end{subfigure}
    \hfill
        \begin{subfigure}[b]{\linewidth}
        \centering
        \includegraphics[width=0.24\textwidth,trim={2cm 4cm 5cm 2cm},clip]{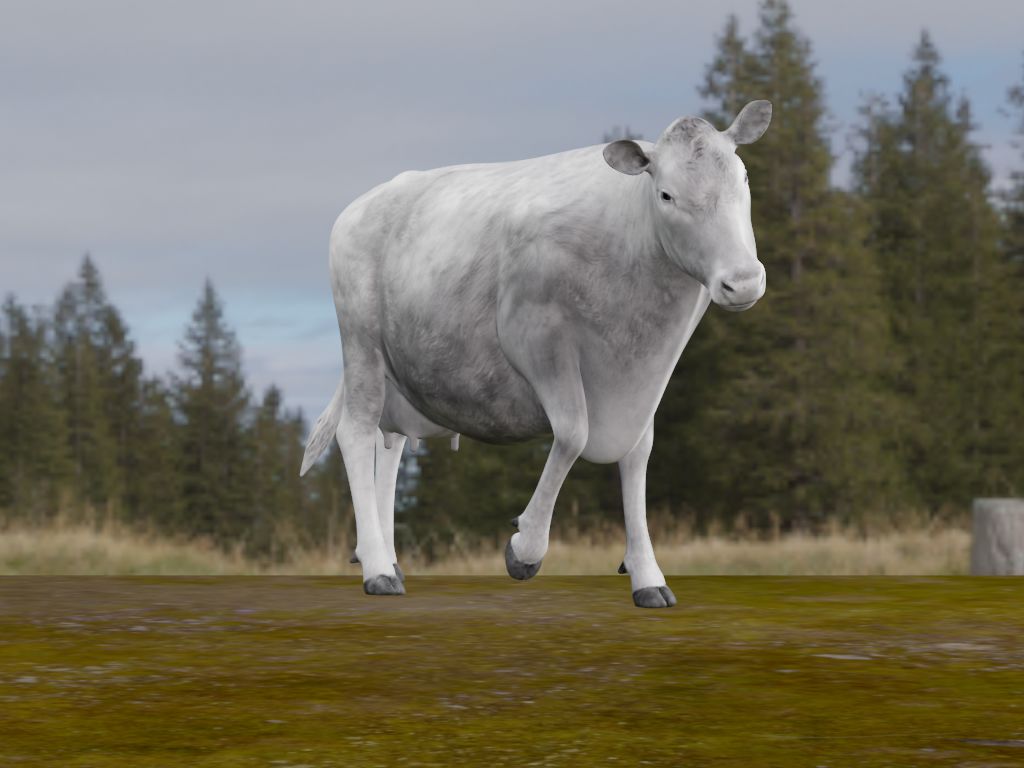}
        \includegraphics[width=0.24\textwidth,trim={2cm 4cm 5cm 2cm},clip]{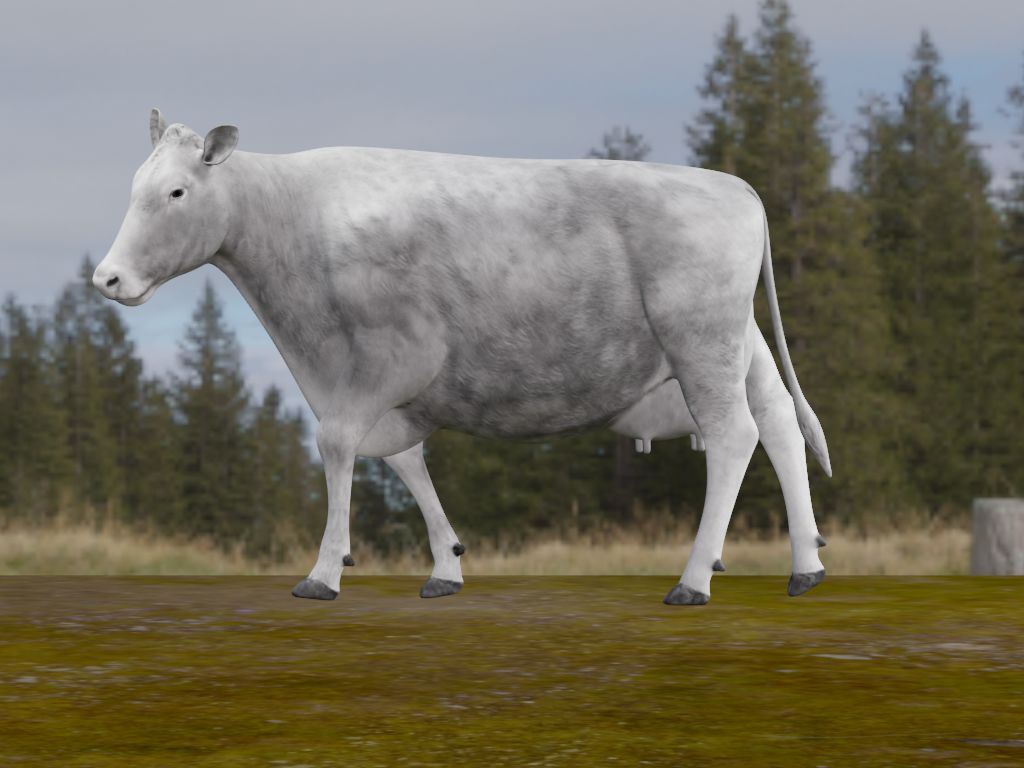}
        \includegraphics[width=0.24\textwidth,trim={2cm 4cm 5cm 2cm},clip]{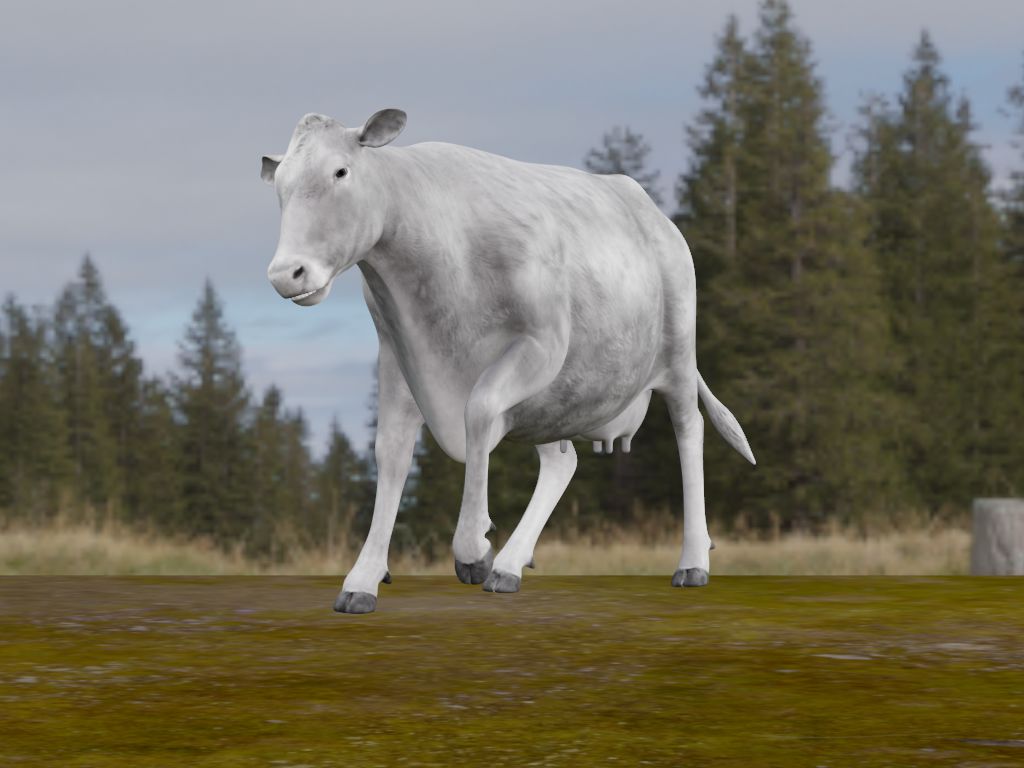}
        \includegraphics[width=0.24\textwidth,trim={2cm 4cm 5cm 2cm},clip]{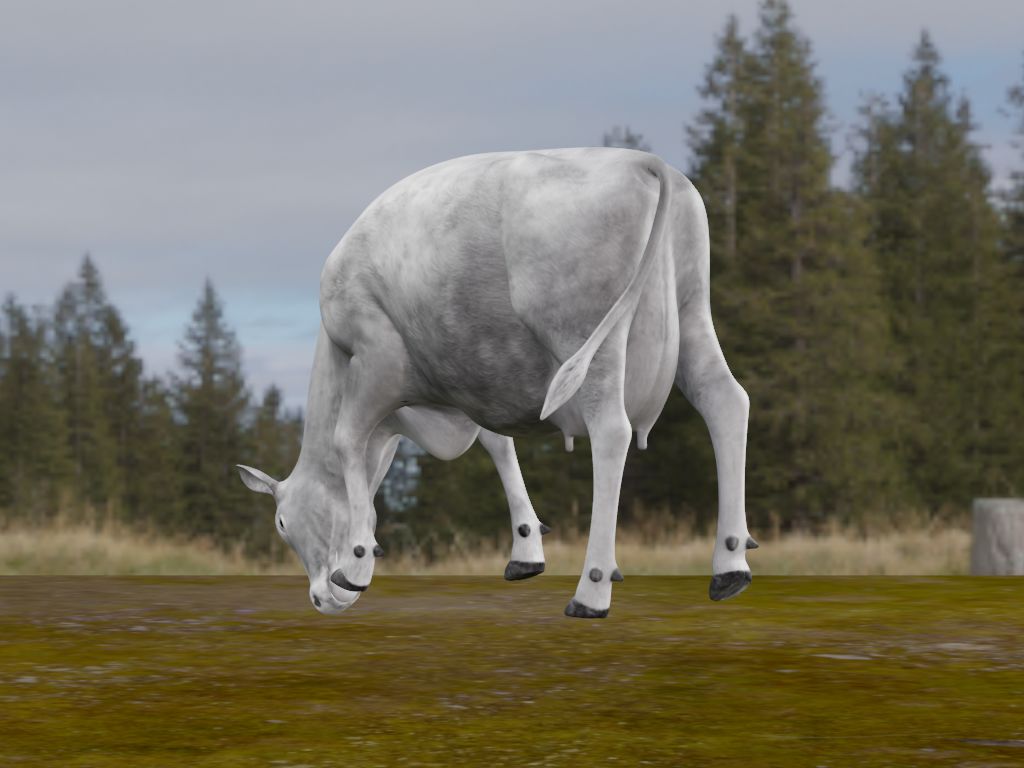}
        \caption{Texture $9$}
        \label{fig:supp:dataset:9}
    \end{subfigure}
    \hfill
        \begin{subfigure}[b]{\linewidth}
        \centering
        \includegraphics[width=0.24\textwidth,trim={2cm 4cm 5cm 2cm},clip]{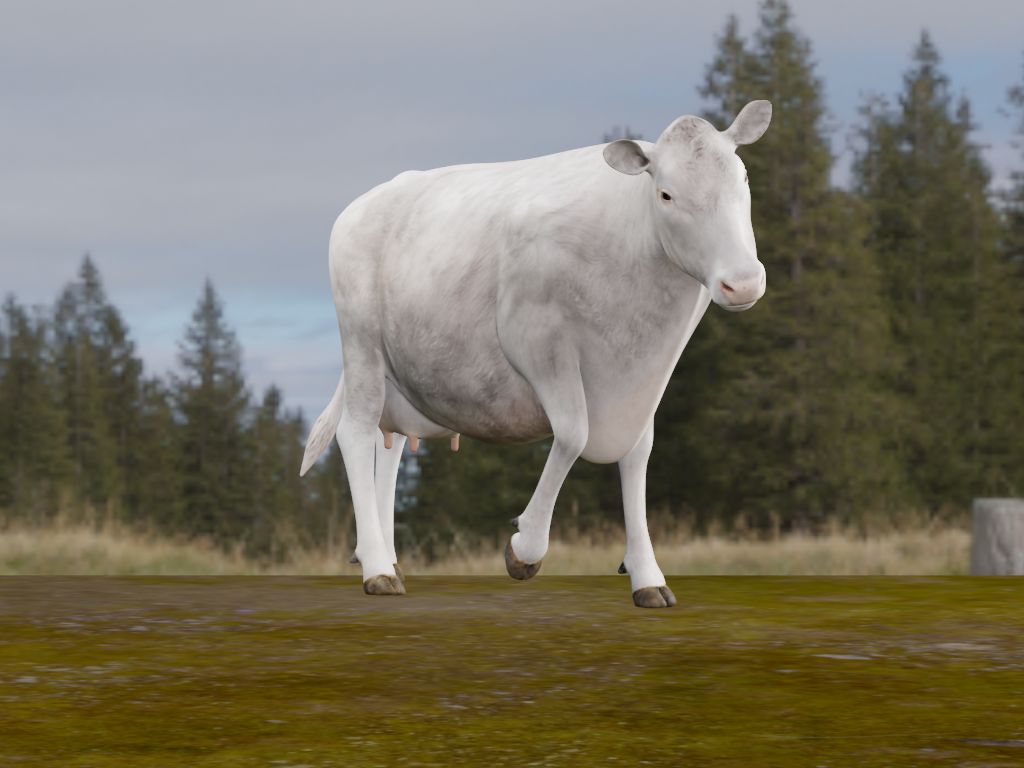}
        \includegraphics[width=0.24\textwidth,trim={2cm 4cm 5cm 2cm},clip]{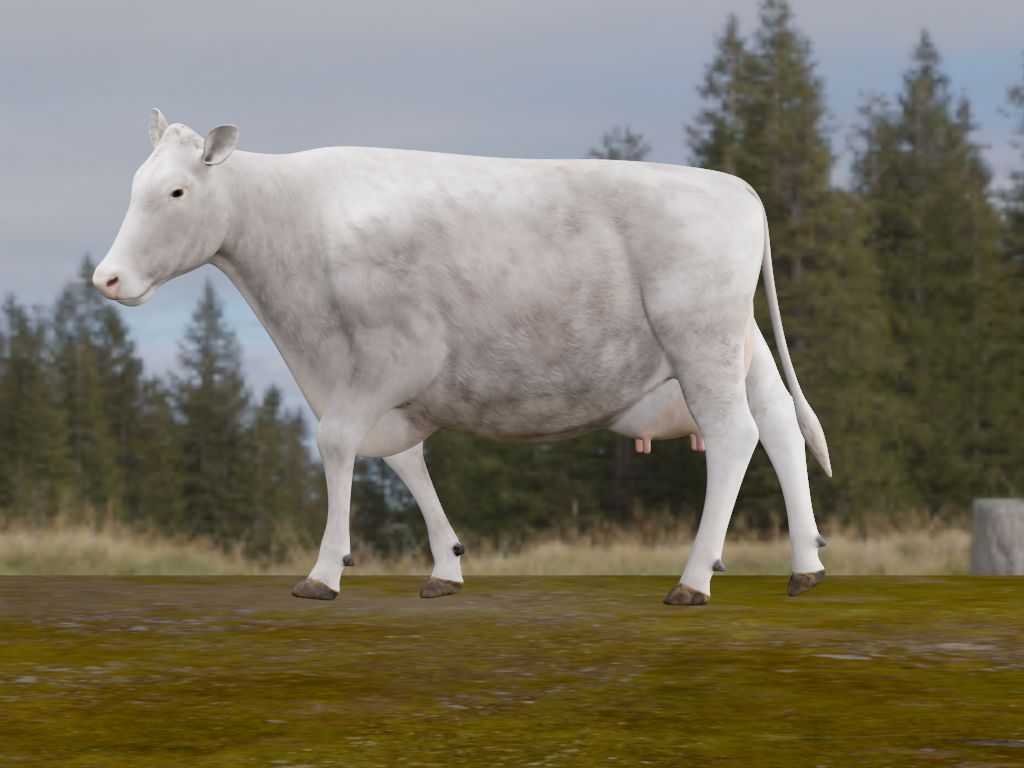}
        \includegraphics[width=0.24\textwidth,trim={2cm 4cm 5cm 2cm},clip]{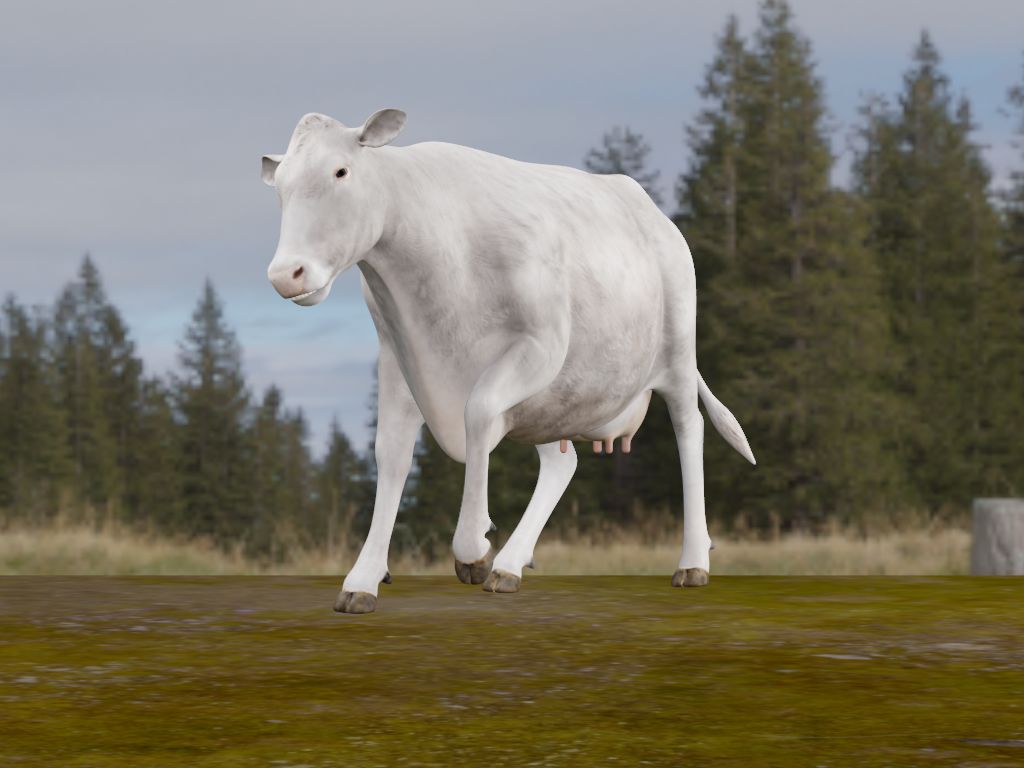}
        \includegraphics[width=0.24\textwidth,trim={2cm 4cm 5cm 2cm},clip]{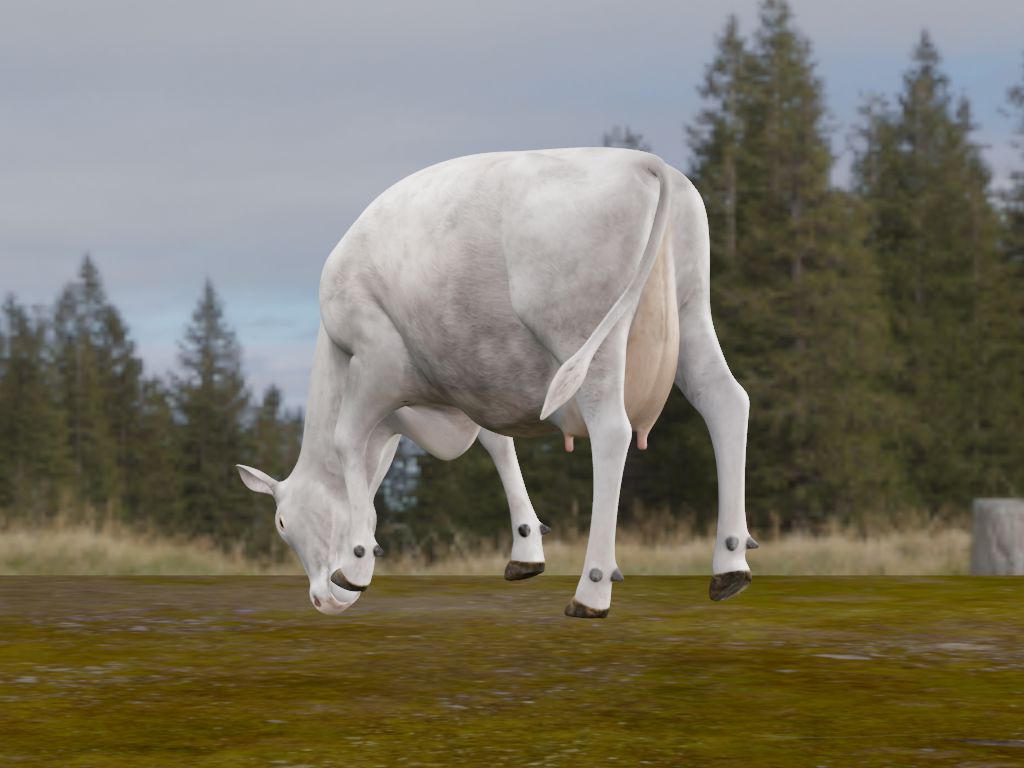}
        \caption{Texture $10$}
        \label{fig:supp:dataset:10}
    \end{subfigure}
    \hfill
        \begin{subfigure}[b]{\linewidth}
        \centering
        \includegraphics[width=0.24\textwidth,trim={2cm 4cm 5cm 2cm},clip]{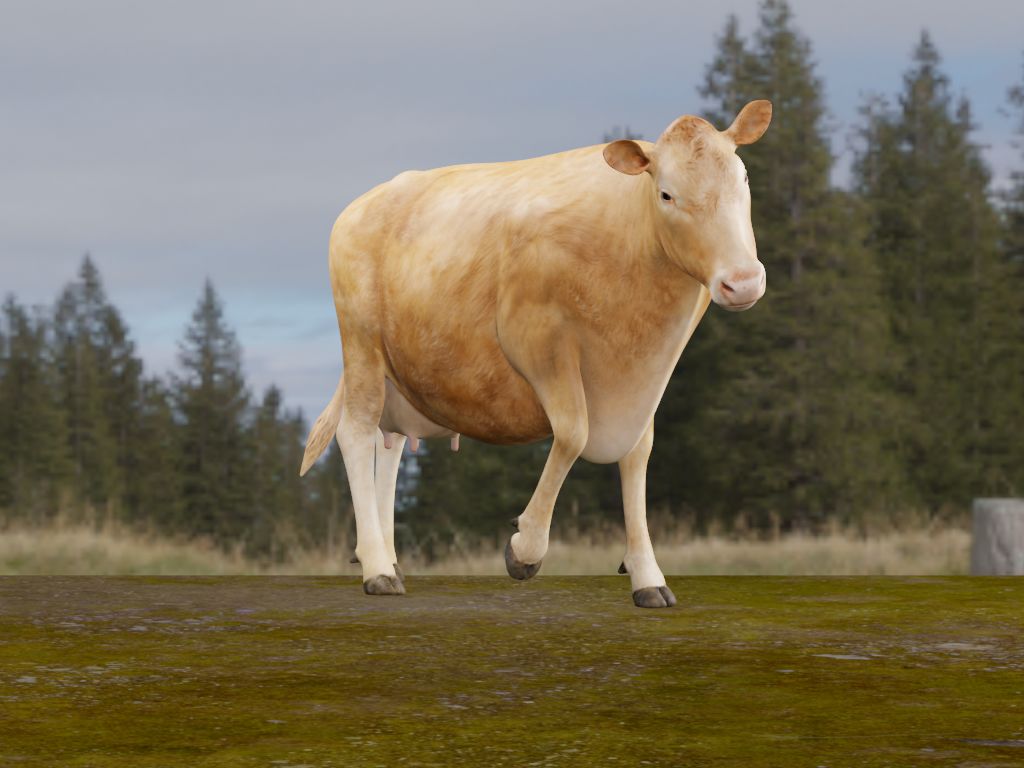}
        \includegraphics[width=0.24\textwidth,trim={2cm 4cm 5cm 2cm},clip]{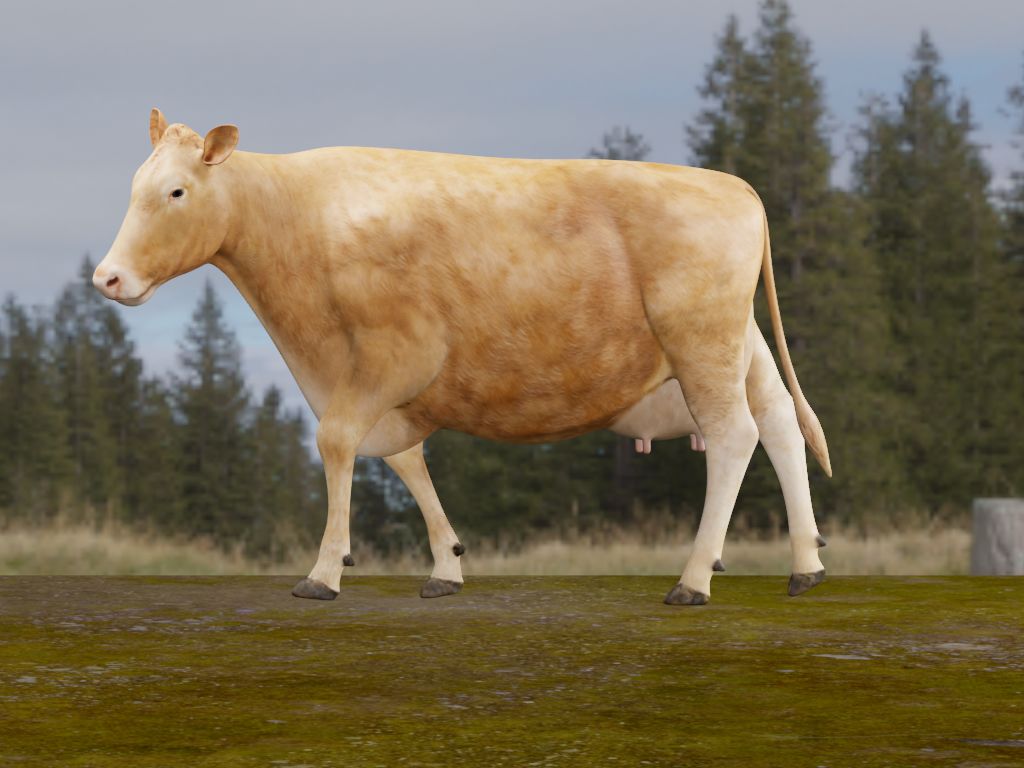}
        \includegraphics[width=0.24\textwidth,trim={2cm 4cm 5cm 2cm},clip]{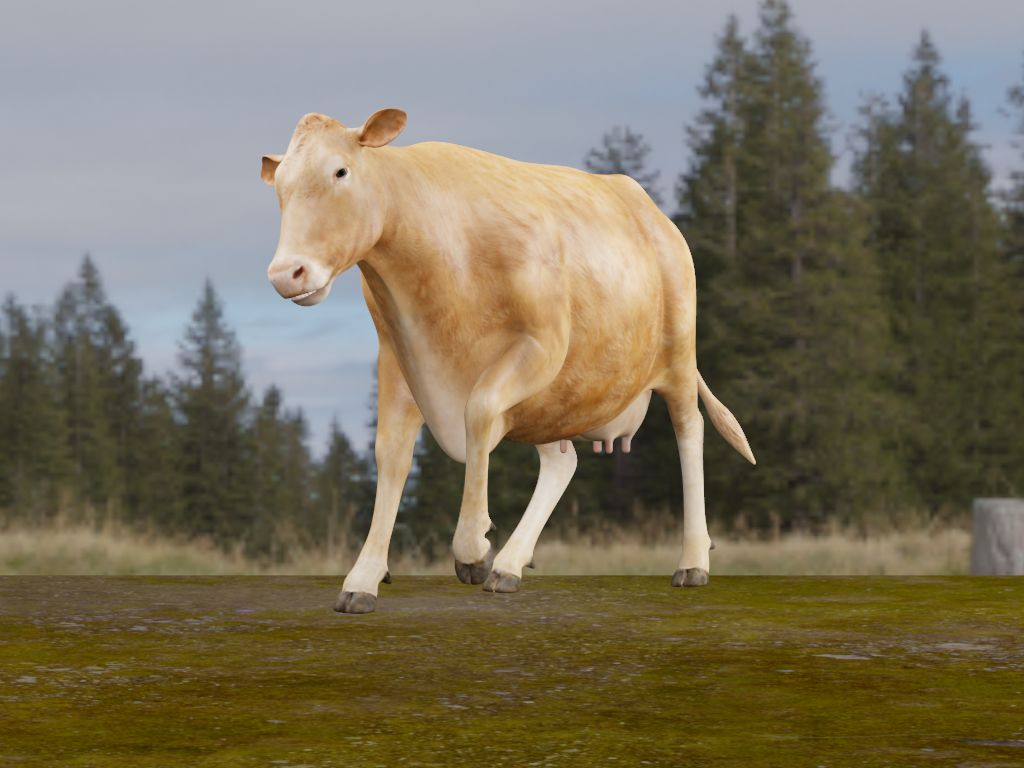}
        \includegraphics[width=0.24\textwidth,trim={2cm 4cm 5cm 2cm},clip]{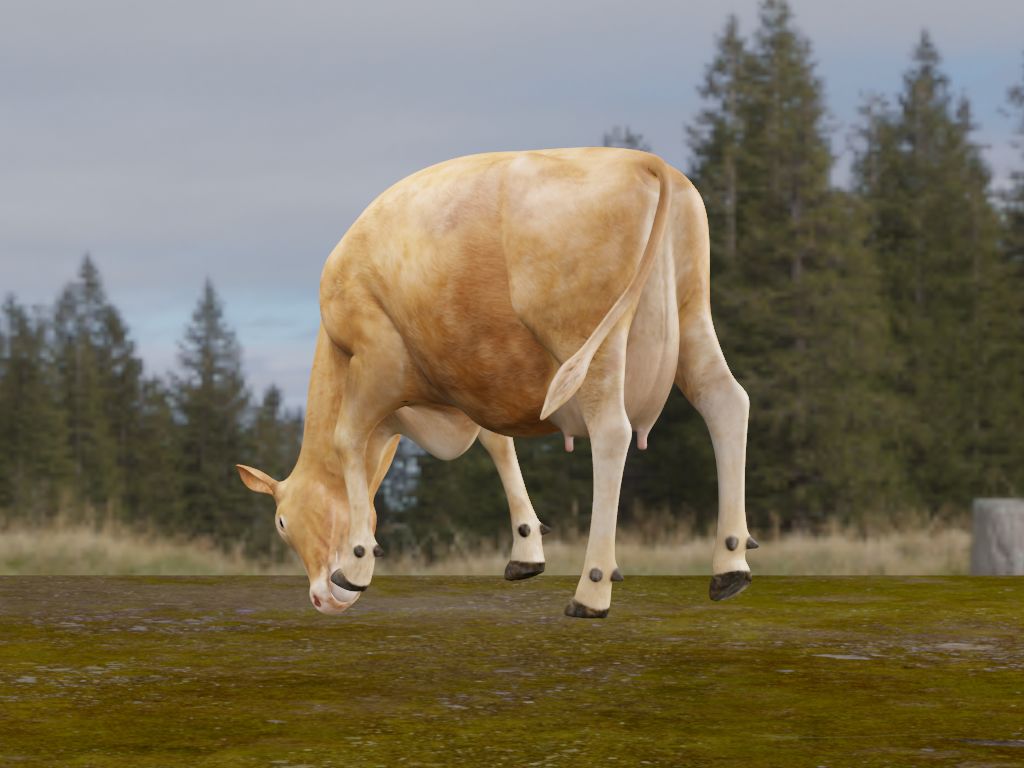}
        \caption{Texture $11$}
        \label{fig:supp:dataset:11}
    \end{subfigure}
    \caption{
        \textbf{Dataset (continued).}
        We show four examples for each distinct texture in our novel~\dataabr dataset.
    }
    \label{fig:supp:dataset}
\end{figure*}

\end{document}